\documentclass{article}

\usepackage{arxiv}

\usepackage[utf8]{inputenc} 
\usepackage[T1]{fontenc}    
\usepackage{hyperref}       
\usepackage{url}            
\usepackage{booktabs}       
\usepackage{amsfonts}       
\usepackage{nicefrac}       
\usepackage{microtype}      
\usepackage{lipsum}
\usepackage{graphicx}
\graphicspath{ {./images/} }
\usepackage[numbers]{natbib}
\usepackage{subfig}
\usepackage{mathtools}
\usepackage{color}
\usepackage{multirow}
\usepackage{colortbl}
\usepackage{xcolor}

\title{Deep-Learning-Based Single-Image Height Reconstruction from Very-High-Resolution SAR Intensity Data}

\author{
 Michael Recla \\
  Institute of Space Technology \& Space Applications\\
  Bundeswehr University Munich\\
  Neubiberg, 85577 (Germany)\\
  \texttt{michael.recla@unibw.de} \\
   \And
 Michael Schmitt \\
  Institute of Space Technology \& Space Applications\\
  Bundeswehr University Munich\\
  Neubiberg, 85577 (Germany) \\
  \texttt{michael.schmitt@unibw.de} \\
}

\begin{document}
\maketitle
\begin{abstract}
Originally developed in fields such as robotics and autonomous driving with image-based navigation in mind, deep learning-based single-image depth estimation (SIDE) has found great interest in the wider image analysis community. Remote sensing is no exception, as the possibility to estimate height maps from single aerial or satellite imagery bears great potential in the context of topographic reconstruction. A few pioneering investigations have demonstrated the general feasibility of single image height prediction from optical remote sensing images and motivate further studies in that direction. With this paper, we present the first-ever demonstration of deep learning-based single image height prediction for the other important sensor modality in remote sensing: synthetic aperture radar (SAR) data. Besides the adaptation of a convolutional neural network (CNN) architecture for SAR intensity images, we present a workflow for the generation of training data, and extensive experimental results for different SAR imaging modes and test sites. Since we put a particular emphasis on transferability, we are able to confirm that deep learning-based single-image height estimation is not only possible, but also transfers quite well to unseen data, even if acquired by different imaging modes and imaging parameters.  
\end{abstract}

\keywords{Deep learning \and Synthetic aperture radar (SAR) \and 3D Reconstruction \and Radargrammetry \and Single image depth estimation}

\section{Introduction}
Besides a semantic mapping of our environment, the reconstruction of the Earth's topography is one of the core tasks of remote sensing. Established standard procedures, such as photogrammetry, radagrammetry, or synthetic aperture radar (SAR) interferometry are based on the stereo principle, i.e. two or more images acquired from slightly different viewpoints are required to reconstruct three-dimensional point coordinates by means of trigonometric intersection. 

In their desire to make scene understanding more economic by reducing the necessary input data, researchers from fields such as robotics and autonomous driving have started to work on the task of single image depth estimation (SIDE), which aims at reconstructing the distance between the camera and the individual target pixels of optical images in a dense manner \citep{Mertan2021}. Once, the pose of the camera is known, these depths can be converted into a 3D representation of the environment. Although SIDE results tend to be a bit more coarse than results based on conventional stereo \citep{Koch2018}, single image 3D reconstruction is a useful tool for scenarios that require only a rough idea of the surroundings. 

In remote sensing, once can imagine many situations, in which a coarse reconstruction of the topographic elevation from a single input would be helpful. Examples include an initial orthorectification of satellite images to ease the co-registration to existing geodata or the chance to provide starting values for a more detailed stereo 3D reconstruction. Thus, researchers have started to transfer the close-range SIDE approaches to the top-view imaging situation prevalent in remote sensing. Among the first experiments into that direction were the \texttt{IM2HEIGHT} approach by \citet{Mou2018} and the \texttt{IMG2DSM} approach by \citet{Ghamisi2018}, which both aimed at height map prediction from airborne optical imagery. While \texttt{IM2HEIGHT} consists of a convolutional and deconvolutional subnetwork, analogous to an encoder-decoder system, \texttt{IMG2DSM} is based on the well-known conditional generative adversarial network architecture \texttt{pix2pix} \citep{Isola2017}, i.e. a generator network creates a ``fake'' height map, which is shown to a discriminator network, which decides whether it is real or self-generated. During the adversarial training, the generator and discriminator networks improve each other in an iterative manner, so that eventually quite realistic height maps are produced.  
In a similar manner, \citet{Amirkolaee2019} also use encoder-decoder CNNs for single image height prediction from aerial images. Their approach is centered on a feature extractor consisting of a complex network of residual convolutional blocks and pooling layers. After the deepest point in the network a more lightweight upsampling network based on unpooling, various convolutional and interpolation layers follows. 

Often times in remote sensing, the availability of suitable training data is an obstacle to the successful application of deep learning \citep{Reichstein2019}, and single image height estimation is no exception. To tackle this challenge,  \citet{Pellegrin2020} modified the KITTI dataset \citep{Geiger2013}: In a two-step process, first the training images are cropped to include only those areas that are farthest from the sensor (i.e. the area around the horizon) and then put into a set of different CNNs and
directed acyclic graph (DAG) networks for height estimation. The resulting models were tested on aerial imagery from their own drone flights. The quality of the results does not come to the ones trained with specially created datasets, but still outperforms the available models from computer vision applied to aerial imagery. 

Finally, \citet{Mahmud2020} add a multi-task learning component to the single image height prediction, as the height component is considered useful additional information for other tasks. Even though their final goal is to reconstruct 3D building block models from single overhead images, the creation of a height map is a ``byproduct'' of their approach. 

Their experiments carried out on both satellite and aerial imagery show increased performance in building outline detection compared to state-of-the-art models designed purely for this task.

\begin{figure}
    \begin{center}
    	\includegraphics[width=.6\linewidth]{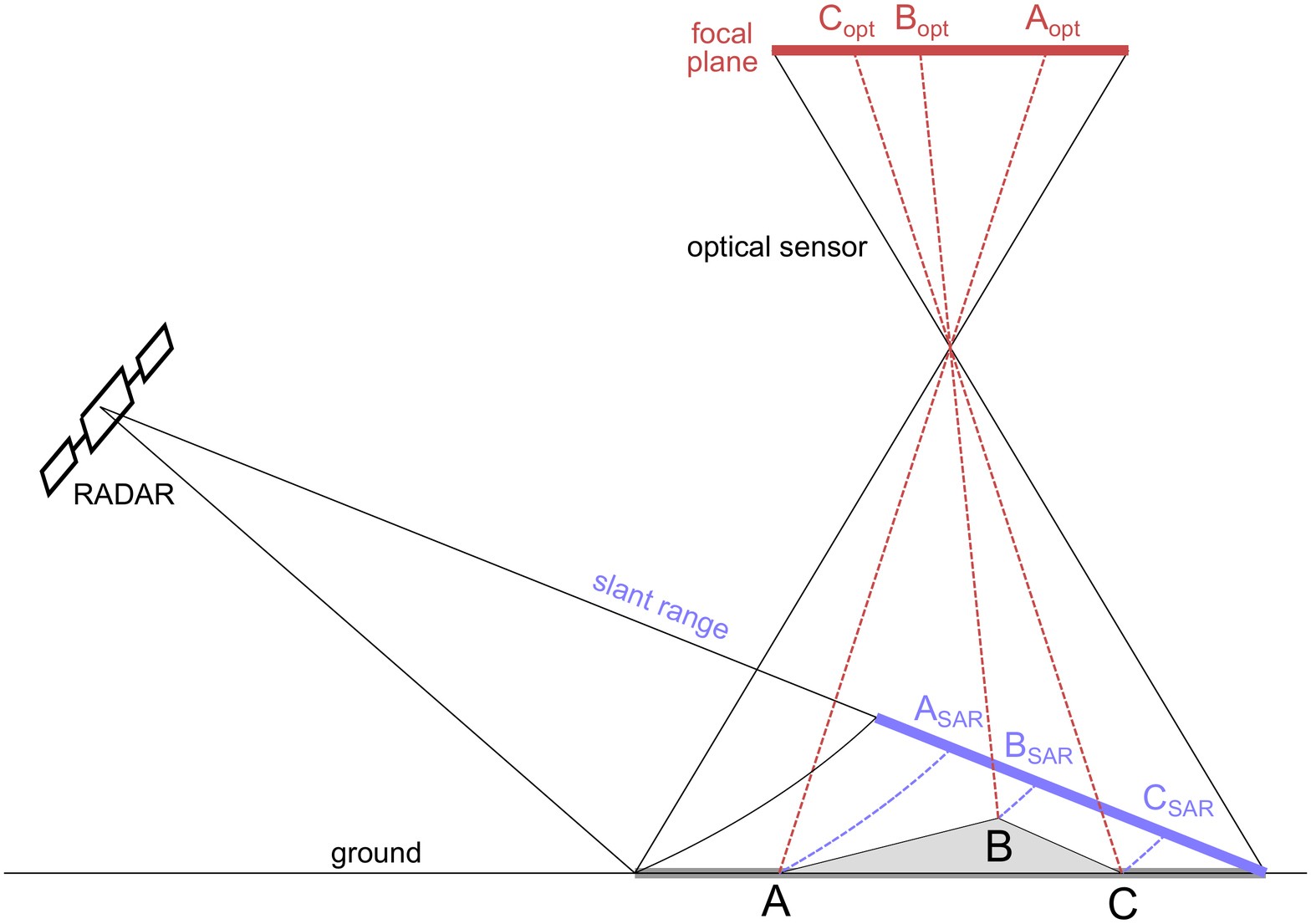}
    	\caption{Comparison between the different imaging geometries of a radar and an optical sensor. While an optical system measures angles with respect to a projection center, a radar records signal travel times and thus distances to the sensor.}
        \label{fig:sar_imaging_geometry}
    \end{center}
\end{figure}
The remainder of this paper is structured as follows: Section~\ref{sec:Method} contains the description of the methodology, which consists of both a small adaptation of an existing single image height estimation neural network architecture as well as an approach to annotate slant-range SAR intensity imagery with heights. In Section~\ref{sec:Data}, we describe the data used in our study, including all relevant peculiarities. Extensive experimental results are summarized in Section~\ref{sec:Experiments}. Finally, the results are discussed in Section~\ref{sec:Discussion}, before the findings are summarized in Section~\ref{sec:Summary}.

So far, all methods for single image height prediction from remote sensing data have only been based on optical images, while the other important sensor modality, SAR, has not been addressed in this context, yet. However, SAR-based single image height reconstruction is also a promising opportunity to explore. On the one hand, SAR offers some fundamental advantages over optical imaging. It is an active sensor, which means that it does not rely on an external source of illumination such as the sun. In addition, wavelengths longer than one centimeter used in SAR imaging penetrate the atmosphere and even small water droplets almost unhindered, so that SAR sensors can be operated regardless of the time of day or weather conditions. Thus, being able to generate at least a coarse representation of the topography from a single SAR image would bear great potential in the context of time-critical, weather-independent Earth observation scenarios. In addition, many established SAR techniques, such as SAR interferometry or SAR tomography would benefit from an approximate knowledge of the observed terrain, e.g. in the context of phase unwrapping or the conversion of relative into absolute heights. Besides, also the cross-modal matching and registration of SAR and optical imagery would benefit from prior knowledge about the 3D structure of the scene \citep{Hughes2020}. 

However, due to the unconventional imaging principle and the resulting imaging geometry, SAR images are difficult to handle for image analysis techniques developed for optical data. Even though the (focused) radar image resembles a conventional optical image at first glance, it carries completely different information. While the image of an optical sensor reflects the chemical characteristics of the observed surface as well as the external scene illumination, in SAR imaging, physical properties such as roughness and dielectric constant of observed targets determine the pixel intensities. In addition, the geometry of a SAR image does not correspond to a central projection as one is used to from optical sensors (and the human visual system). Refer to Figure~\ref{fig:sar_imaging_geometry} for a comparison. Instead of angles, signal travel times are measured. Rows and columns of SAR images relate to the distance of the object to the sensor (range) and the position of the sensor in the direction of flight (azimuth).
These different properties and peculiarities of SAR images make it impossible to apply models developed and trained on optical imagery with SAR data. On the other hand, the side-looking nature of SAR may even promise, at least in theory, an advantage for the estimation of absolute heights above ground compared to classic aerial photographs taken from a nadir point of view.

With this paper, we present the first-ever investigation of deep learning-based single-image height estimation from very-high-resolution SAR intensity imagery.

\section{Deep Learning-based Single Image Height Reconstruction for SAR Imagery}\label{sec:Method}

As a backbone for the investigations in this paper, we use a slightly adapted variant of the \texttt{IM2HEIGHT} architecture proposed by \citet{Mou2018}, which is described in the following. Our approach for the connection of height maps and SAR images in their native slant-range geometry, which is crucial for the training of the SAR-specific \texttt{IM2HEIGHT} network, is described as well. 

\subsection{The Adapted IM2HEIGHT Architecture}

\begin{figure}[t!]
    \begin{center}
    	\includegraphics[width=\textwidth]{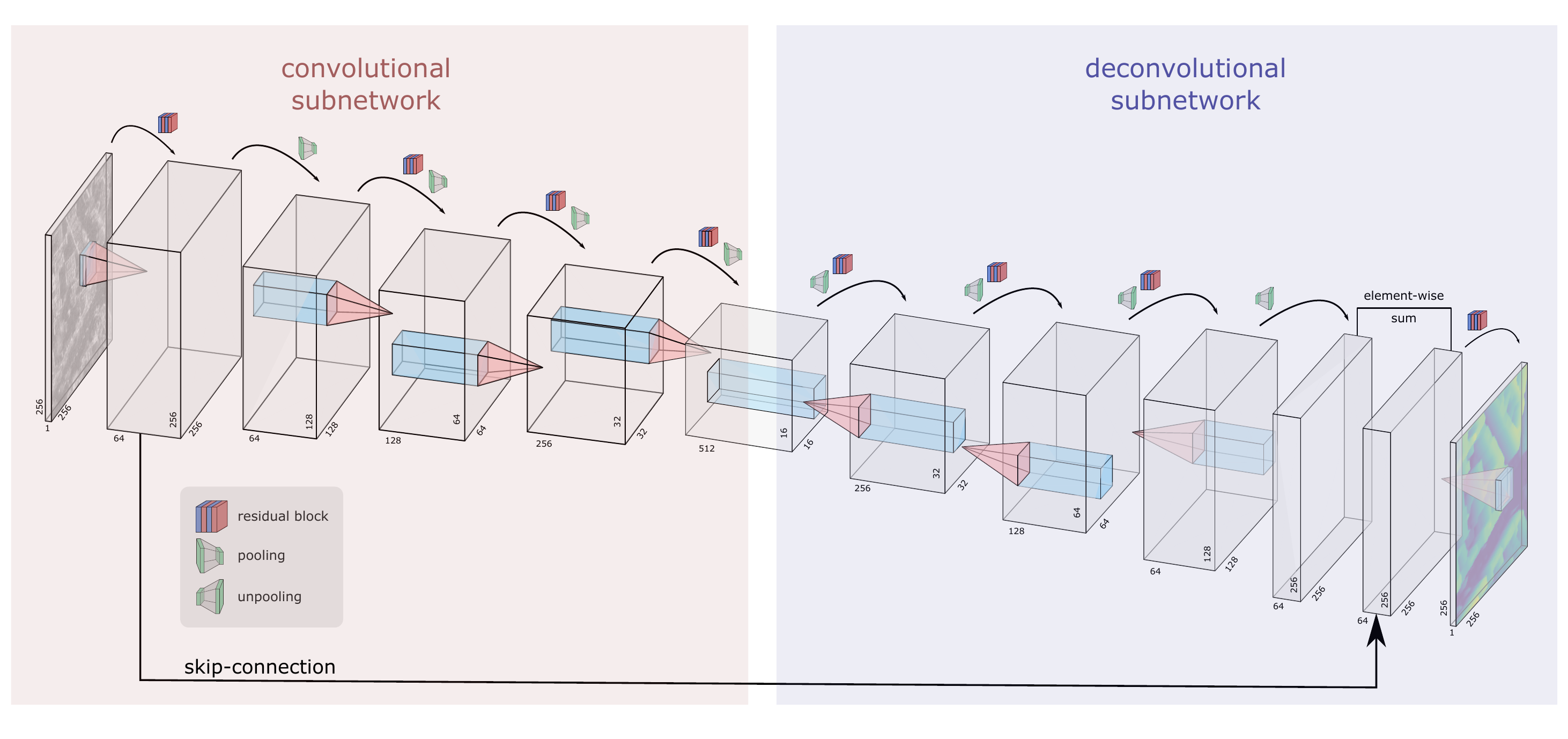}
    	\caption{Overview of the network architecture. The color shaded areas represent the convolutional (encoding) and the deconvolutional (decoding) part of the network. The input (a SAR intensity image) and output (the corresponding height map) are of the same dimensions and consist of one channel, marked by the small numbers on the edges of the layers. Between the layers, the residual blocks, pooling, and unpooling operations are applied, as this is illustrated by the icons.}
        \label{fig:network_architecture}
    \end{center}
\end{figure}

The original \texttt{IM2HEIGHT} network was designed to predict height maps from color aerial imagery at very high resolution. To make it fit for the task with SAR intensity images as input, several modifications were required:
\begin{itemize}
    \item The ResNet blocks, which comprise the core of the architecture, use a different layout.
    \item The optimizer and the loss function were changed for a better fit to the SAR data.
    \item Several minor adaptations were made, including a change of the first layer to accept single-channel SAR intensity imagery instead of RGB photos, and an element-wise summation of the cross-bottleneck skip connection instead of concatenation .
\end{itemize}
Details of the basic network architecture and our adaptations are described in the following, while the network is illustrated in Fig. \ref{fig:network_architecture}. Following the principle of a fully convolutional network, it contains only convolutional and pooling layers, no fully connected layers. This has the advantage that it does not depend on the dimension of the input and that the dimension of the output matches that of the input. The network itself consists of two parts. A convolutional sub-network and a deconvolutional subnetwork. The convolutional part can be seen as a kind of encoder. It performs an abstraction of the global and local features in the input SAR image and thus a condensation of important information. The deconvolutional part turns this process around and deep feature maps need to become high resolution images again, or in this case, height maps. For this purpose, the encoder is simply mirrored and becomes a decoder. So the convolutions map deep structures to shallower ones and instead of pooling, unpooling is carried out. In its core, the network consists of residual blocks as introduced with the ResNet architecture \citep{He2016}. The blocks of traditional neural networks attempt to learn within one layer the mapping $\mathcal{H}$ from an input $x$ to a desired output $y$, i.e. $y = \mathcal{H}(x)$. In contrast, a residual block follows a different approach. It tries to learn only the residual function $\mathcal{F}$: $y = \mathcal{F}(x) + x $. Practically, this is achieved by feeding the input back into the result bypassing each block across the entire network. As a result, gradients retain numerically much more stable values even in deeper layers, largely eliminating the problem of vanishing gradients. In addition, the network is motivated (but not required) to generate an output that is similar to the input in structure, which is a reasonable choice in image-to-image translation tasks \citep{Meraner2020}.

\begin{figure}[]
    \begin{center}
    	\includegraphics[width=.35\linewidth]{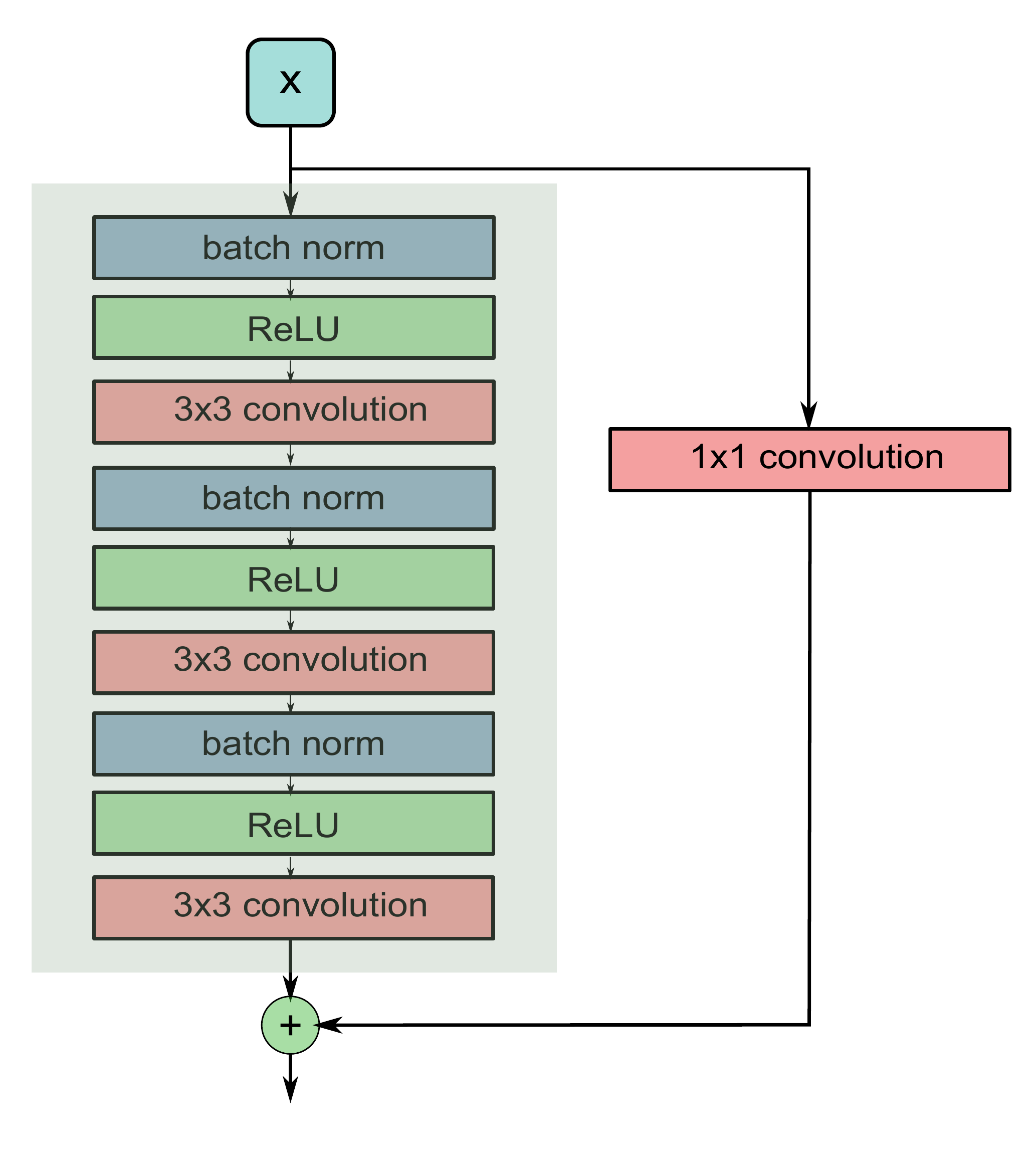}
    	\caption{Fully pre-activated residual block. As opposed to a ``conventional'' residual block (like it was used by \cite{Mou2018}), the order of the individual layers is upside down and it keeps the shortcut path clean of activation or batch norm.}
        \label{fig:res_block}
    \end{center}
\end{figure}

Figure~\ref{fig:res_block} shows the structure of such a residual block. The layout of these blocks has been arranged differently compared to those in the original \texttt{IM2HEIGHT} network. The shortcut path is free of activation functions or batch normalization, and also the order of these individual layers within the block has been reversed. From our experience, this leads to faster learning and slightly improved results in our validation runs. The individual convolutional layers use only very small filters with a size of $3 \times 3$. The convolutional stride is fixed to 1 pixel, as is the padding. As a result, the width and height of the input image are maintained. These dimensions are only changed in the maxpooling layers where they get exactly halved. Masks with a size of $2 \times 2$ and a stride of 2 pixels are used. Since the depth of the feature maps is changed within the block, the identity shortcut must also be adapted to the output. This is done with another convolutional layer, but in this case with a $1 \times 1$ filter. It is worth mentioning that the change in the number of channels happens only in the first convolution operator of the block, which is why they are not all identical. Furthermore, for the convolutional layers which are followed by a batch normalization, no biases need to be stored or trained, since they would be lost anyway by the subsequent batch norm operations.

Since the nature of an encoder-decoder system creates a kind of bottleneck, spatial information is lost, especially through pooling operations. To counteract this somewhat, the indices of the maximum values (that are transferred during maxpooling) are stored for each pooling layer. These indices are then used in the corresponding unpooling process to map the according value back to the original position. The usefulness of this implementation is, of course, related to whether there is some spatial correlation between the input image and the desired result. This is certainly the case in single-image height prediction: Edges in the image usually also represent edges in the height map, for example.

Following these considerations, the introduction of a skip connection across the bottleneck is equally promising. A skip connection transfers features from an early layer (with more intact spatial information) across the whole network to the last layers, where they get conflated with the output. This was implemented here by element-wise summation, very much in the ResNet manner (as opposed to the \texttt{IM2HEIGHT} design, where the features are concatenated). This should result in edges not being soft washed in the deconvolutional process.

Finally, since SAR intensity images are used here instead of RGB images, the input image in the network of course consists of only one single channel, leading to an input size of $h \times w \times c = 256 \times 256 \times 1$. As mentioned above, the architecture does not depend on the height and width of the image and works with any size. However, tests showed that the originally used square input images of size $256$ are well suited. With a size of $256$, the network reaches a feature map size of $16 \times 16$ at a depth of $512$ after 4 pooling operations at its bottleneck.

\subsection{Model Optimization}

To optimize the model, an ADAM optimizer with the default learning rate of 0.001 was used. The mean square error (MSE), also referred to as $\mathcal{L}_2$ loss, was chosen as loss function. With $y$ the ground truth and $\hat{y}$ the estimated value for $N$ estimates it can be expressed as:
\begin{equation}
	\mathcal{L}(y, \hat{y}) = \frac{1}{N} \sum_{i=0}^{N}(y - \hat{y}_i)^2
\end{equation}
Just as with \texttt{IM2HEIGHT}, only a single image gets passed as a batch. As our own experiments showed, the quality of the results suffers from a larger batch size, even if this would shorten the training time. Because of the use of ReLUs as activation functions, all weights are initialized with a Kaiming uniform distribution. To avoid overfitting, 15\% of the training data is randomly declared as a validation set for an early stopping mechanism.

To make the network more robust to different viewing angles, a kind of data augmentation was implemented. Instead of actually increasing the number of training images, a random flipping is included in the data loader. Thus, for each draw from the data pool, the image is randomly mirrored at one of the main axes before flowing into the network. Doing so, the orientation of the images changes from epoch to epoch and the network learns to handle images from different viewing angles, where the buildings are tilted towards the left image border instead of the right, for example.

\subsection{Linking SAR Images and Height Maps}\label{sec:HeightProjection}

One of the key components of using supervised learning to estimate heights from a SAR image in an image-to-image manner is the possibility to provide the model with training data consisting of pixel-wisely coregistered SAR intensity images as observations and height maps as annotation targets. Although it would theoretically be possible to produce such height maps directly from pairs or stacks of complex SAR data using SAR interferometry or SAR tomography, such maps are usually comparably noisy and incomplete due to inherent noise in the interferometric phase, ambiguous heights due to the layover effect as well as phase unwrapping residuals, and due to gaps caused by radar shadow. We thus decided to use the best possible height reference available for urban areas: governmental 3D data derived from either airborne laser scanning or dense photogrammetry and available either in the form of point clouds or digital surface models. However, when using such an external height source, an intelligent 
projection of given height data into the geometry of the radar system, which also takes visibility into account, becomes crucial.

\begin{figure}[]
    \begin{center}
    	\begingroup
		\subfloat[\label{fig:height_projection_principle_3d}]{\includegraphics[width=.35\textwidth]{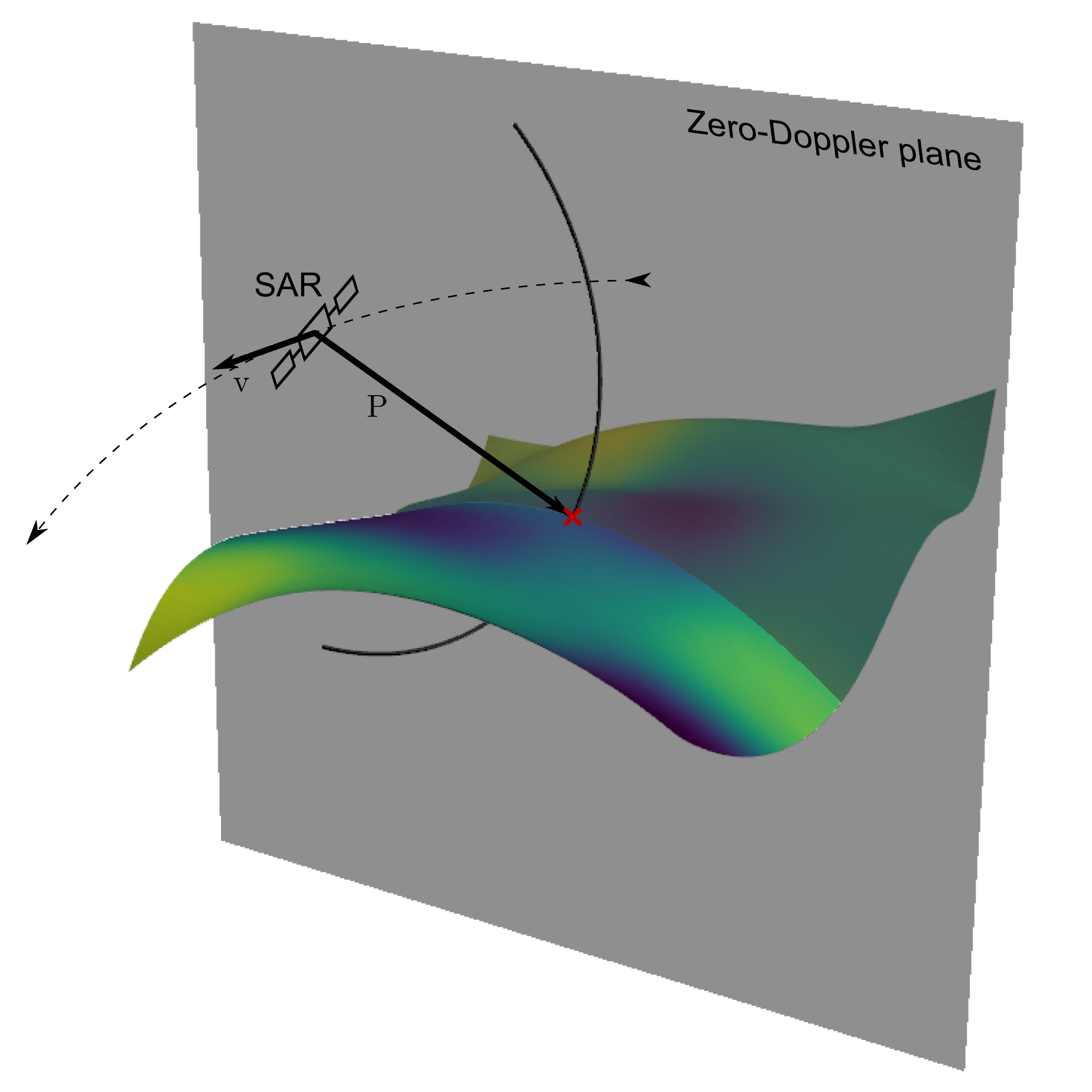}}
		\hspace{1mm}
		\subfloat[\label{fig:height_projection_principle}]{\includegraphics[width=.5\textwidth]{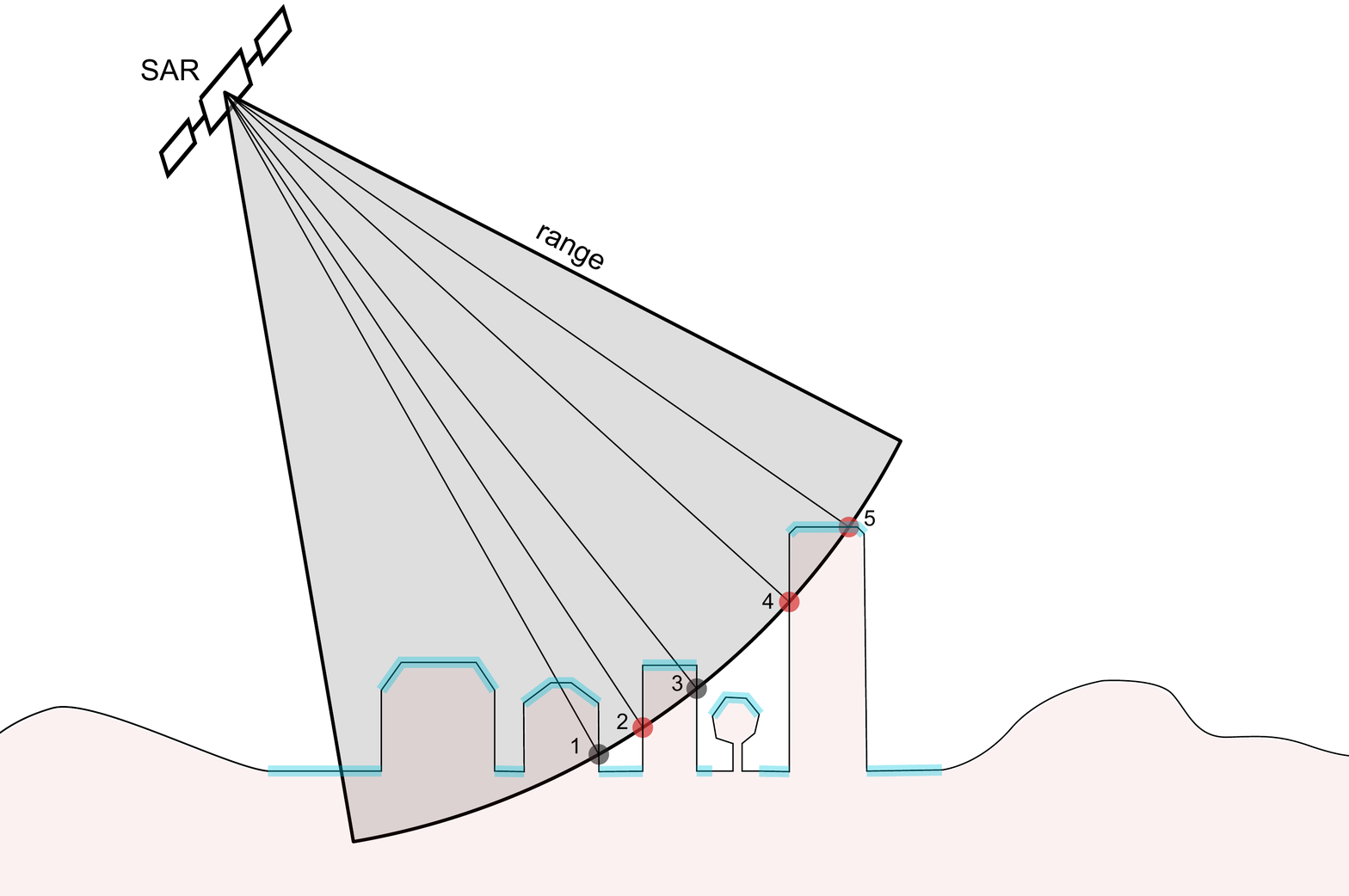}}
        \endgroup
    	\caption{(a) Geometrical relationships of intersecting the terrain model with the Zero-Doppler plane (grey plane). The plane is defined by the position of the sensor $\mathbf{x}_s$ (origin) and its velocity vector $\mathbf{v}_s$ (plane normal). The plane specifies one row in the SAR image as azimuth, the column is then given by the radius of the circle (range) with the sensor as origin. The intersection point of the circle on the plane with the terrain gives the height value for this pixel. \newline (b) Principle of the projection geometry. Shown is a slice through the terrain on the Zero-Doppler plane. The red dots are visible for the sensor, the black dots lie in the shadow. The blue shaded areas represent the height data provided by the digital surface model as a raster file.}
    \end{center}
\end{figure}

The basis for this projection is the Zero-Doppler geometry of the SAR system. Within this geometry, it is possible to locate each pixel of the image on a surface model. The basic idea of Zero-Doppler geometry is that the measurement direction is exactly perpendicular to the trajectory of the sensor. If a backscattering object is exactly in the measuring direction, the resulting Doppler frequency changes its sign and becomes zero for this moment. The geometric relationship between the known position vector of the sensor, the known velocity vector and the measurement direction can now be
made use of. The goal is to assign a height value from the ground reference data to each pixel in the SAR image. Figure \ref{fig:height_projection_principle} shows the principle behind the approach. The Zero-Doppler position defines a slice of the terrain, where the backscattering object lies. However, known as the layover effect, several points can have the same distance to the sensor and are therefore mapped together to one resolution cell. In addition to that, some points are behind blocking obstacles and as a result not seen from the sensor. 

The established way of linking a SAR image with ground heights is called backward geocoding or back projection \citep{Raggam1988}. In essence, it connects every ground pixel of a 2.5D height map with its corresponding image pixel by simply solving the range-Doppler equations in a straight-forward manner.
The major drawback of back projection-based geocoding, however, is that there is usually not a discrete point in the height data for every combination of range and azimuth, so there are inevitably holes in the resulting elevation image, i.e. pixels to which no height value has been assigned. In the case of orthometric images, these holes could simply be closed by interpolation. Since we are working in radar slant-range geometry, however, this is not possible: Due to the layover effect, multiple height assignments per pixel could occur. Besides, if the ground height reference is not provided in the form of a 3D point cloud but as a 2.5D height map, vertical elements (e.g. facades) are not represented. In slant-range SAR imagery, however, facades make up a significant share for urban scenes. Finally, detecting points lying in the radar shadow is also challenging (cf. points 1 and 3 in Figure~\ref{fig:height_projection_principle}). These heights must not be included in the result because they were never seen by the sensor.

\begin{figure}[]
    \begin{center}
    	\includegraphics[width=.4\textwidth]{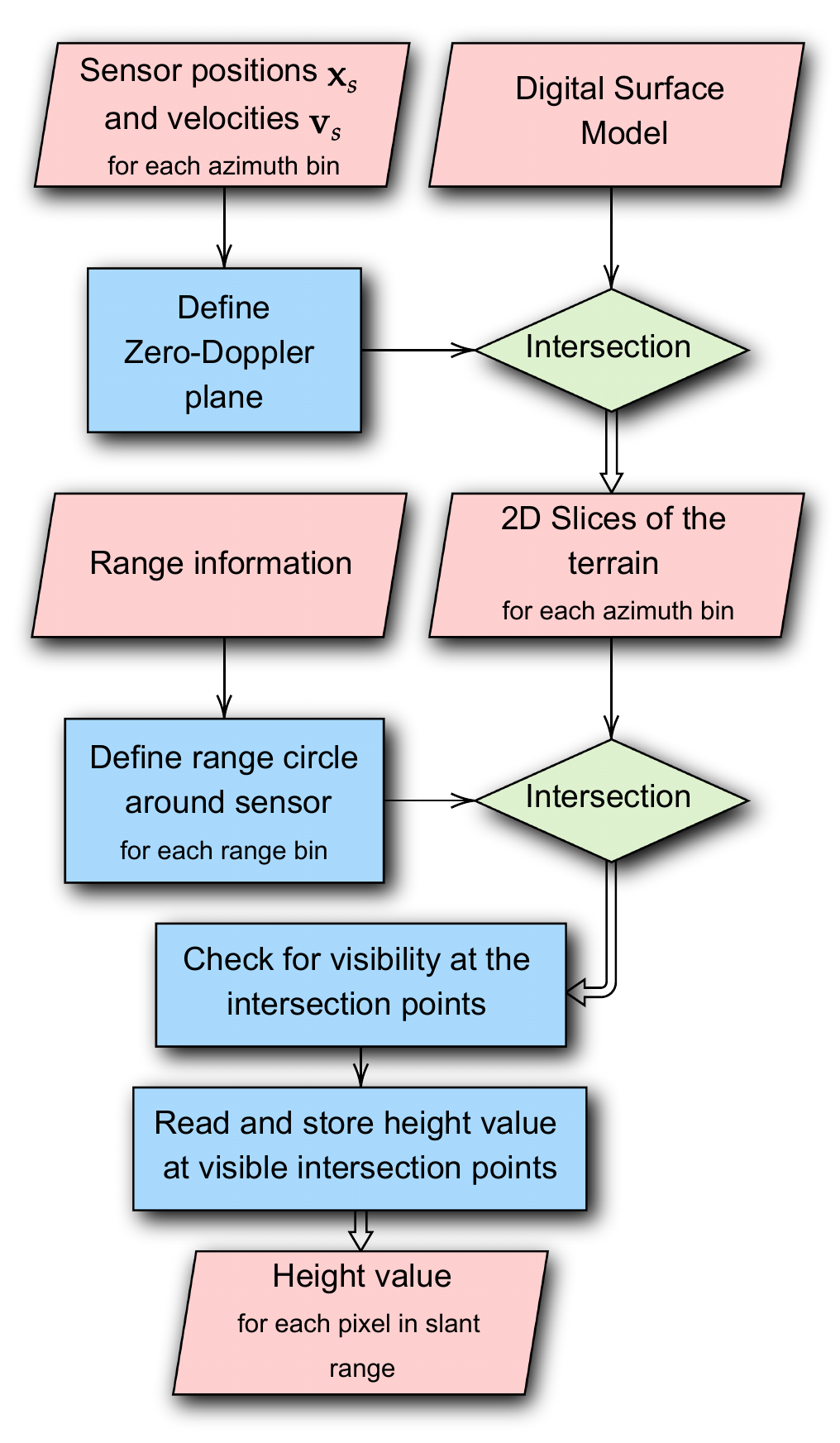}
    	\caption{Workflow of the height projection process from a digital surface model to the pixels in slant range geometry.}
        \label{fig:projection_flowchart}
    \end{center}
\end{figure}

For all those reasons, we propose an enhanced forward projection-based procedure. As usually, instead of searching for each discrete point in the elevation model its corresponding position in the radar image, in this case for each pixel a position in the height model is determined, whose height is then stored. The detailed procedure is illustrated in Figure~\ref{fig:projection_flowchart} and described in the following: 

The first step is to determine the position and the velocity of the sensor from the metadata of the SAR image. The velocity vector $\mathbf{v}_s = \dot{\mathbf{x}}_s$ then represents the normal and the position vector $\mathbf{x}_s$ represents the origin of the Zero-Doppler plane (visualized as gray plane in Fig.~\ref{fig:height_projection_principle_3d}), defined by 
\begin{equation}
\mathbf{v}_s \cdot \mathbf{p} = \frac{\dot{\mathbf{x}}_s}{|\dot{\mathbf{x}}_s|} \cdot \frac{\mathbf{x}_s - \mathbf{x}_t }{|\mathbf{x}_s - \mathbf{x}_t|},    
\end{equation}
where $\mathbf{p}$ is the vector pointing from the satellite position $\mathbf{x}_s$ to the target position $\mathbf{x}_t$. This plane can then be used to create an intersection with the height model (represented by the colored surface in Figure~\ref{fig:height_projection_principle_3d}). This can be done by simply intersecting lines with a plane. The lines in this case are given by the grid of the surface model. For example, one row of the height raster can be considered as a polyline with the single elements of this polyline being the lines which are going to get intersected. Mathematically, the intersection point $\mathbf{p}_{\text{inter}}$ can be found by calculating
\begin{equation}
    \mathbf{p}_{\text{inter}} = \mathbf{p}_0 + \mathbf{u} \cdot t,
\end{equation}
with
\begin{equation}
    \mathbf{u} = \mathbf{p}_1 - \mathbf{p}_0 
\end{equation}
being the orientation of the line going through its start and end points $\mathbf{p}_0$ and $\mathbf{p}_1$ and 
\begin{equation}
    t = -\frac{\mathbf{v}_s\cdot\mathbf{w}}{\mathbf{v}_s\cdot\mathbf{u}}
\end{equation}
being a parameter that determines whether the intersection point is between ${p}_0$ and ${p}_1$. Here, $\mathbf{w} = \mathbf{p}_0 - \mathbf{x}_s$ is the vector pointing from the sensor to the start point of the corresponding line. If $t>1$ or $t<0$, the intersection point is not on the relevant part of the line, but part of its extension before or its start or end, respectively.

\begin{figure}
    \begin{center}
		\begingroup
		\subfloat[SAR image]{\includegraphics[width=.32\textwidth]{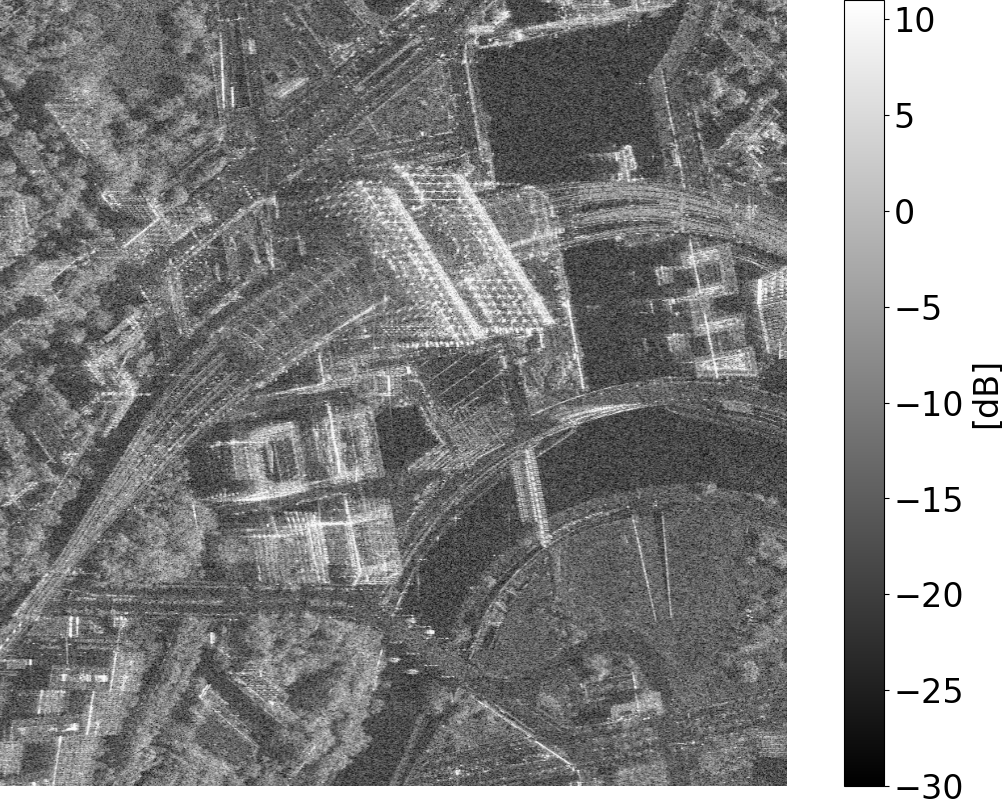}}
		 \hspace{1mm}
		\subfloat[Digital surface model \label{fig:proj_a}]{\includegraphics[width=.32\textwidth]{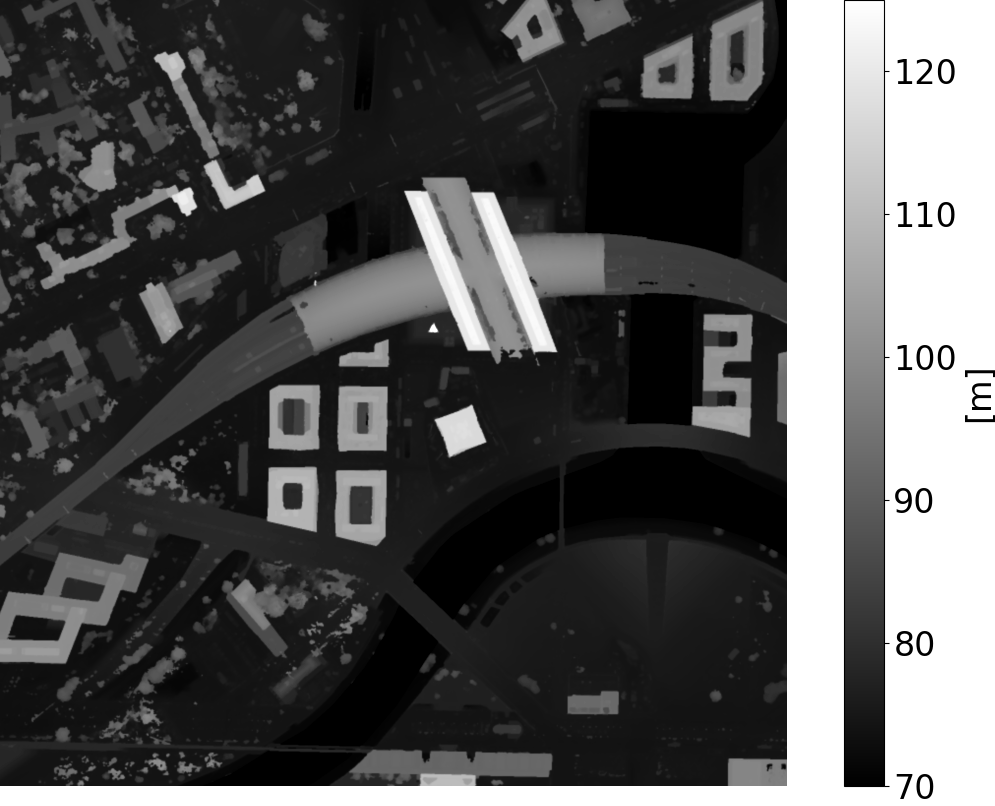}}
		 \hspace{1mm}
		\subfloat[Projected heights \label{fig:proj_c}]{\includegraphics[width=.32\textwidth]{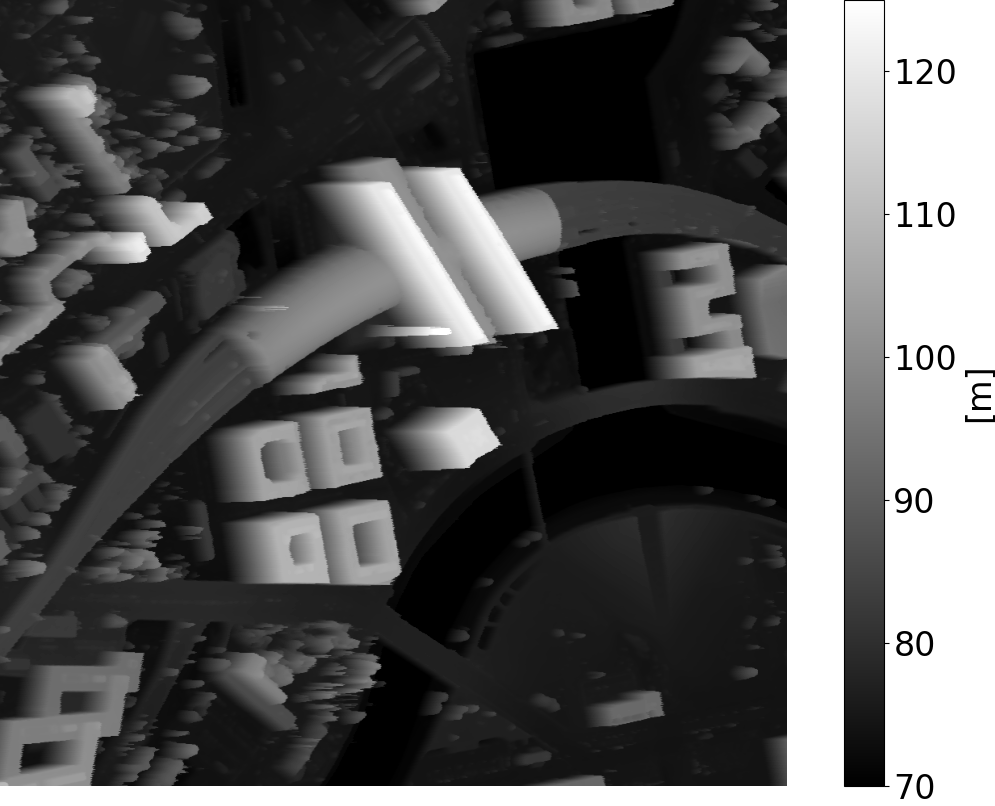}}
		\endgroup
		\caption{Example of the height projection process for a SAR image showing the area around the Central Station in Berlin, Germany. (a) SAR intensity image in slant range geometry (pixels approximately squared), (b) digital surface model (2.5D height map), (c) heights projected into the geometry of (a). All heights are geometrically defined above the reference ellipsoid.}
        \label{fig:proj_heights}
    \end{center}
\end{figure}

\begin{figure}
	\begin{center}
        \includegraphics[width=.8\textwidth]{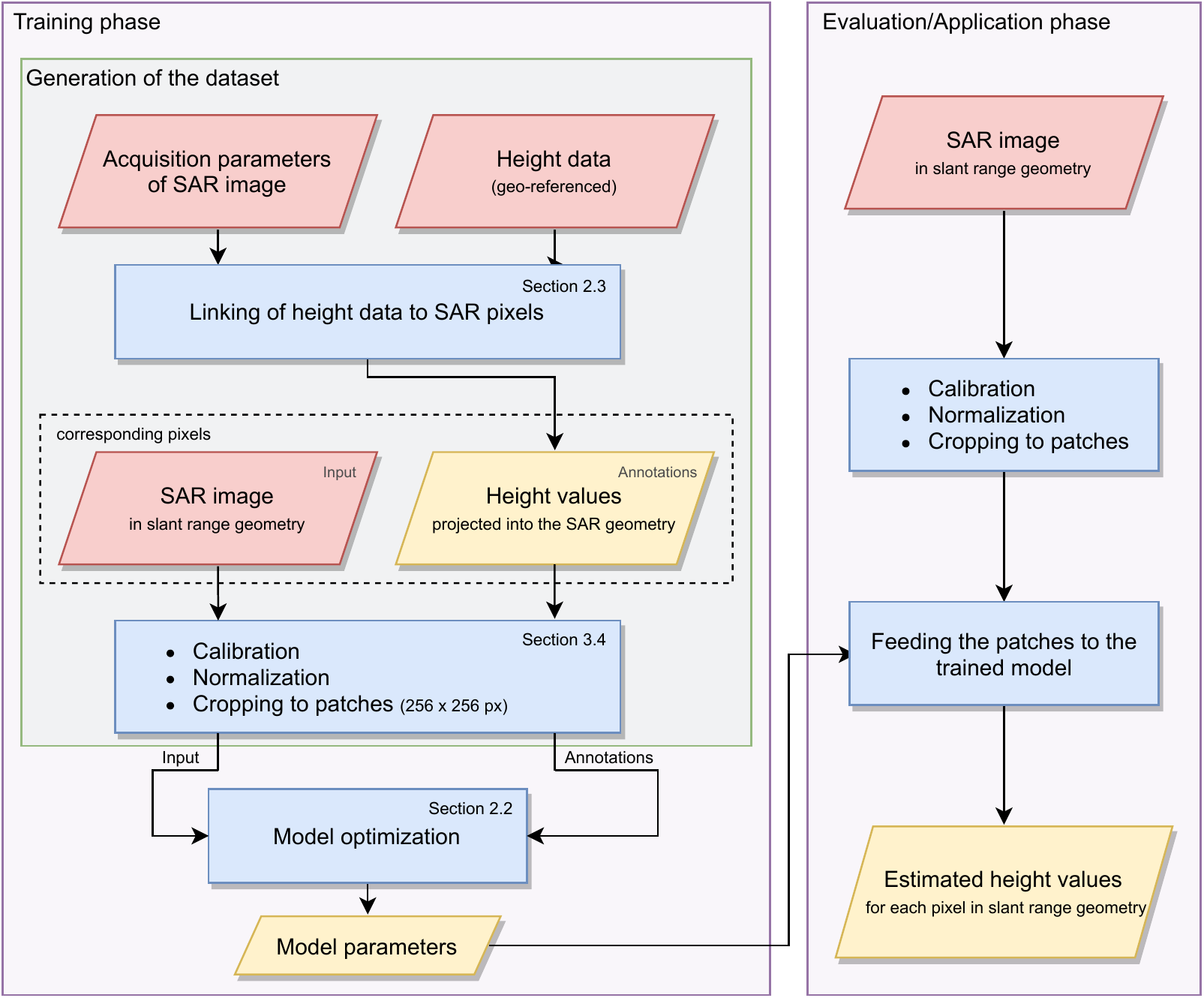}		
		\caption{Flowchart of the work described in this paper. The left box describes the necessary steps to train a convolutional neural network for single-image height estimation, while the right side schematically illustrates the procedure to use the trained network for actual height estimation.}
		\label{fig:global_flowchart}
	\end{center}
\end{figure}

By doing this for every polyline of the height raster, the intersection points generate a slice of the terrain on the Zero-Doppler plane. From then on, the problem is reduced to a two-dimensional one, since all of the intersection points lie on the plane. Now, with the sensor as the center, a circle can be intersected with the terrain slice for each range value. Each slice is again defined as a polyline which can be split up to single lines, defined by a start and an end point $\mathbf{p}_0$ and $\mathbf{p}_1$, respectively. The range circle is defined by its radius $r$ and its center co-located with the sensor position $\mathbf{x}_s$. With 
\begin{equation}
    \mathbf{p}'_i = \mathbf{p}_i - \mathbf{x}_s,
\end{equation}
we can calculate
\begin{equation}
    \mathbf{d} = \left(\begin{matrix}d_x\\d_y\end{matrix}\right) = \mathbf{p}'_1 - \mathbf{p}'_0
\end{equation}
and derive 
\begin{equation}
    d_r = \sqrt{d_x^2 + d_y^2}
\end{equation}
and 
\begin{equation}
    D = x'_0y'_1 - x'_1y'_0
\end{equation}
therefrom. 

Finally, the coordinates of the intersection points can be determined by
\begin{equation}\label{eq:circleInter}
\begin{split}
    x_{\text{inter}} &= \frac{Dd_y \pm \text{sgn}\left(d_y\right)d_x\sqrt{\Delta}}{d_r^2}\\
    y_{\text{inter}} &= \frac{Dd_x \pm |d_y|\sqrt{\Delta}}{d_r^2}
\end{split}
\end{equation}

In equation (\ref{eq:circleInter}), $\Delta$ serves as discriminant, which defines the type of intersection:
\begin{equation}
    \Delta = r^2d_r^2 - D^2 \begin{cases}<0~~~\text{no intersection}\\=0~~~\text{tangent}\\>0~~~\text{intersection}\end{cases}
\end{equation}

The resulting intersection points $\mathbf{p}_\text{inter} = \left(\begin{matrix}x_\text{inter}&y_\text{inter}\end{matrix}\right)^T$ are typically between two points defined by the elevation model. These heights can be easily interpolated linearly, which solves, e.g., the problem of facade heights not present in 2.5D height maps. Also, invisible points can be excluded by checking if the connecting line between sensor and intersection point intersects the height slice for a second time, resembling the act of ray-tracing. However, the problem of multiple height possibilities for a SAR image pixel remains due to the side-looking nature of the SAR imaging geometry. To deal with this situation, we decided to always store only the largest height value for every pixel. 

By repeating the whole process for every pixel in the SAR image, automatically all pixels in the radar raster are occupied, which also solves the problem of the holes as observed in standard backward geocoding. Figure~\ref{fig:proj_heights} shows an example result of the proposed height preparation procedure. 

Figure~\ref{fig:global_flowchart} shows the overall methodological process of the here presented approach in a high-level manner, linking the individual subtopics of section~\ref{sec:Method} and \ref{sec:Data}.

\section{Data}\label{sec:Data}

For the investigations documented in this paper, we have made use of very-high-resolution SAR imagery acquired by the German TerraSAR-X satellite. In this section, we will first discuss the peculiarities of SAR intensity data as provided by such a modern SAR mission. Then, we will describe the preparation of the actual experimental data used in the deep learning experiments.

\begin{figure}[b!]
	\begin{center}
		\begingroup
		\subfloat[stripmap mode\label{fig:stripmap}]{\includegraphics[width=.3\textwidth]{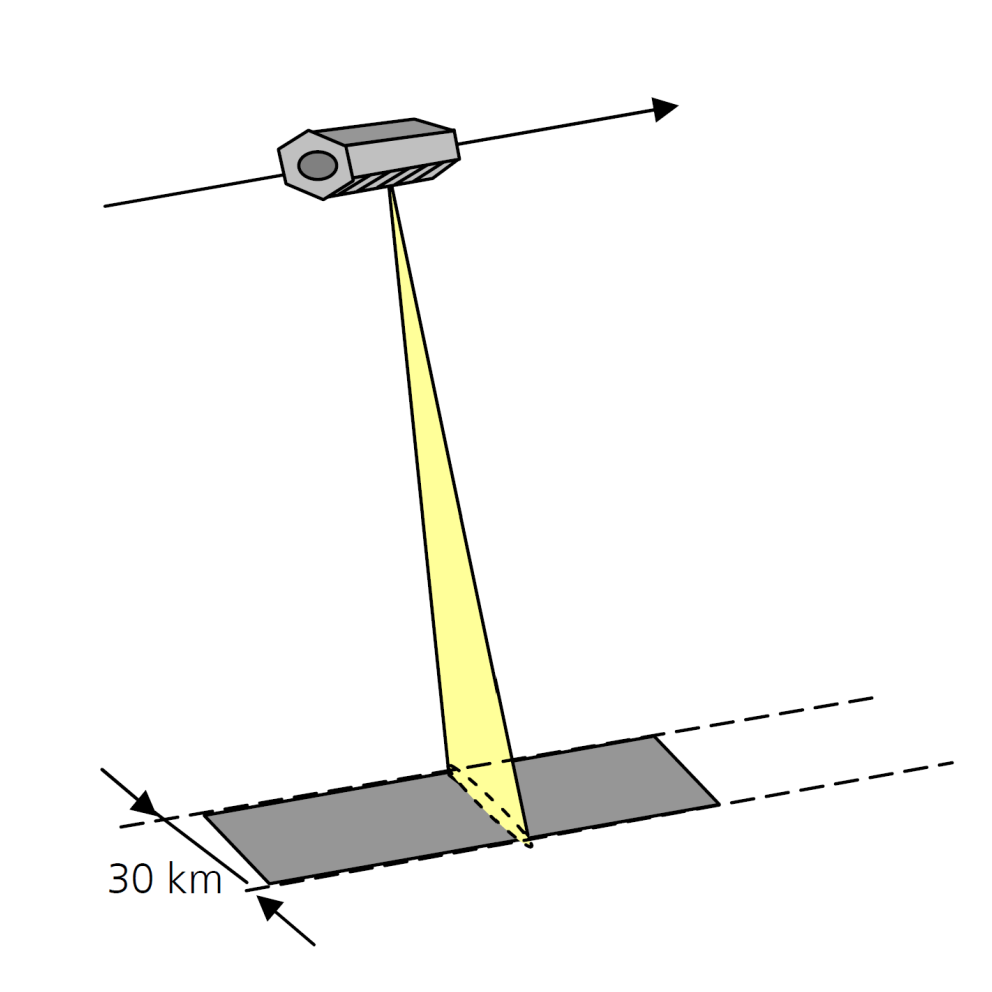}}
		\hspace{3mm}
		\subfloat[four beam ScanSAR mode\label{fig:scansar}]{\hspace*{2mm}\includegraphics[width=.3\textwidth]{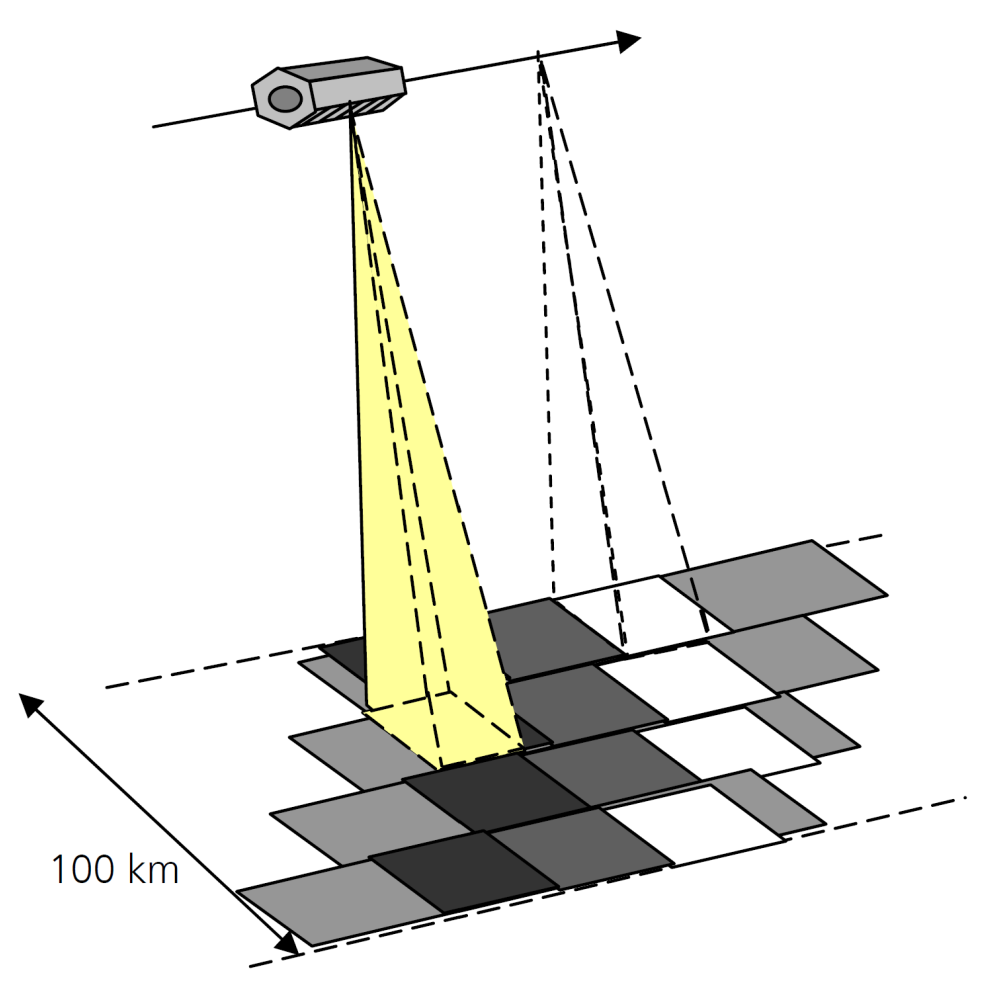}}
		\hspace{3mm}
		\subfloat[spotlight mode\label{fig:spotlight}]{\includegraphics[width=.3\textwidth]{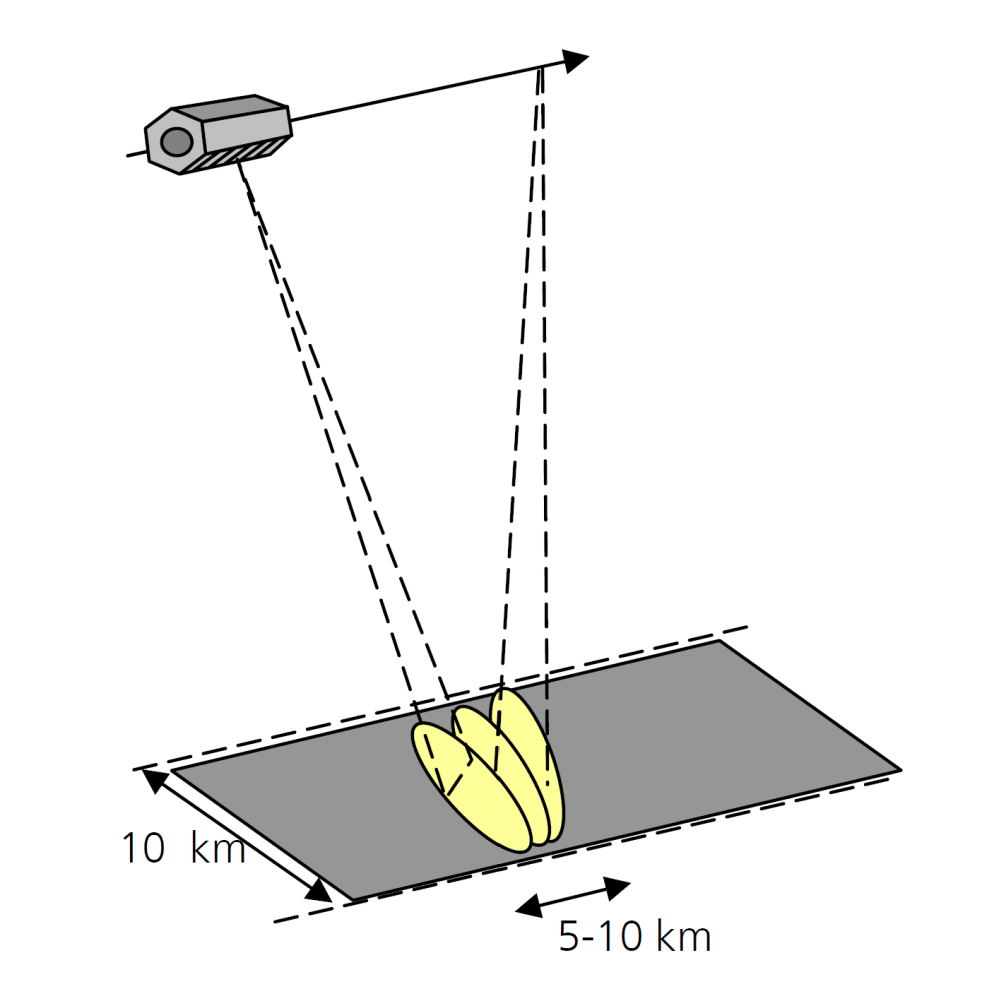}}
		\endgroup
		\caption{Imaging geometries of TerraSAR-X (from \cite{Eineder2008})}
		\label{fig:terrasar_modes}
	\end{center}
\end{figure}

\subsection{TerraSAR-X Image Products}

TerraSAR-X is a German high resolution imaging radar Earth observation satellite operating in X-band. In general, imagery are acquired in three different imaging modes (cf. Figure~\ref{fig:terrasar_modes}): The Stripmap (SM) mode is the standard mode of SAR imaging. The ground swath is illuminated with a continuous sequence of pulses while the antenna beam is pointed to a fixed angle in elevation and azimuth. This results in an image strip with constant image quality in azimuth \citep{Eineder2008}. Figure~\ref{fig:stripmap} shows the principle. The Spotlight (SL) imaging modes use phased array beam steering in azimuth direction to increase the illumination time which results in a larger synthetic aperture. The larger aperture leads to a higher azimuth resolution at the cost of azimuth scene size. The 300 MHz HighRes Spotlight (HS300) mode uses a higher bandwidth for an improved range resolution. In the extreme case of Staring Spotlight (ST) the antenna footprint would rest on the scene and the scene length corresponds to the length of the antenna footprint \citep{Eineder2008}. Of course, this mode does not produce a continuous image of the Earth’s surface, but is used to provide images of highest resolution for selected areas of interest. In general images acquired in one of the spotlight modes are of higher resolution, but also much more expensive than stripmap images, since the antenna has to be traced specifically (cf. Figure~\ref{fig:spotlight}).
Finally, the ScanSAR mode (which is illustrated in Fig. \ref{fig:scansar}) increases the overall swath size on the ground by electronic antenna elevation steering to switch between SM swaths with different incidence angles. The increased imaging area leads to a faster coverage at the cost of a lower azimuth resolution due to a reduced azimuth bandwidth (i.e. a smaller synthetic aperture). Since this paper focuses on (very)-high-resolution imagery, ScanSAR data is not considered in our experiments. Table~\ref{tab:terrasar_products} summarizes the characteristics of the stripmap and spotlight modes used in this work. Generally, TerraSAR-X images are delivered in the form of different products, which differ with respect to the amount of geometric pre-processing that was applied to them. They range from complex-valued images in the original radar slant-range geometry (SSC: single look slant range complex) to variants that have been multi-looked and geocoded. Since we are interested in performing single-image height prediction for original SAR images, only SSC images are used in our study.

\begin{table}[b!]
	\caption{Properties of the different imaging modes of TerraSAR-X as used in this work (taken from \cite{Airbus2014})}
	\begin{center}
		\begin{tabular}{m{0.2\textwidth} m{0.1\textwidth}  m{0.1\textwidth}  m{0.12\textwidth}  m{0.1\textwidth}  m{0.1\textwidth}  m{0.1\textwidth} }
			\toprule
			Imaging Mode &   Standard Scene Size \small{[km]} &  Maximum Acquisition Length \small{[km]} & Slant Range Resolution \small{[m]} & Azimuth Resolution \small{[m]} & Polarisation  &  Full Performance Range \small{[°]} \\ \cmidrule(r){1-1} \cmidrule(lr){2-2} \cmidrule(lr){3-3} \cmidrule(lr){4-4} \cmidrule(lr){5-5} \cmidrule(lr){6-6} \cmidrule(l){7-7}
			Stripmap (SM) &  \parbox{0.1\textwidth}{30 $\times$ 50 \\ 15 $\times$ 50} & 1,650 & \parbox{0.12\textwidth}{1.2 \\ 1.2} & \parbox{0.1\textwidth}{3.3 \\ 6.6} & \parbox{0.1\textwidth}{Single \\ Dual} & 20° to 45° \\\addlinespace
			Spotlight (SL) &  10 $\times$ 10 & 10 & \parbox{0.12\textwidth}{1.2 \\ 1.2} & \parbox{0.1\textwidth}{1.7 \\ 3.4} & \parbox{0.1\textwidth}{Single  \\  Dual} & 20° to 55° \\\addlinespace
			HighRes Spotlight \newline 300~MHz (HS300) &  10 $\times$5 & 5  & 0.6 & 1.1 & Single  & 20° to 55° \\\addlinespace
			HighRes Spotlight (HS) & 10 $\times$ 5 & 5  & \parbox{0.12\textwidth}{1.2 \\ 1.2} & \parbox{0.1\textwidth}{1.1 \\ 2.2} & \parbox{0.1\textwidth}{Single \\  Dual} & 20° to 55°  \\\addlinespace 
			Staring Spotlight (ST) & 4 $\times$ 3.7  &  3.7 & 0.6 & 0.24  &  Single &  20° to 45°   \\ 
			\bottomrule
		\end{tabular}
		\label{tab:terrasar_products}
	\end{center}
\end{table}

\subsection{SAR Image Data}\label{sec:SARimageData}
\begin{figure}[]
	\begin{center}
		\begingroup
		\subfloat[\label{fig:sar_comp_a}Munich: HighRes SpotLight (300 MHz)]{\includegraphics[width=.24\textwidth]{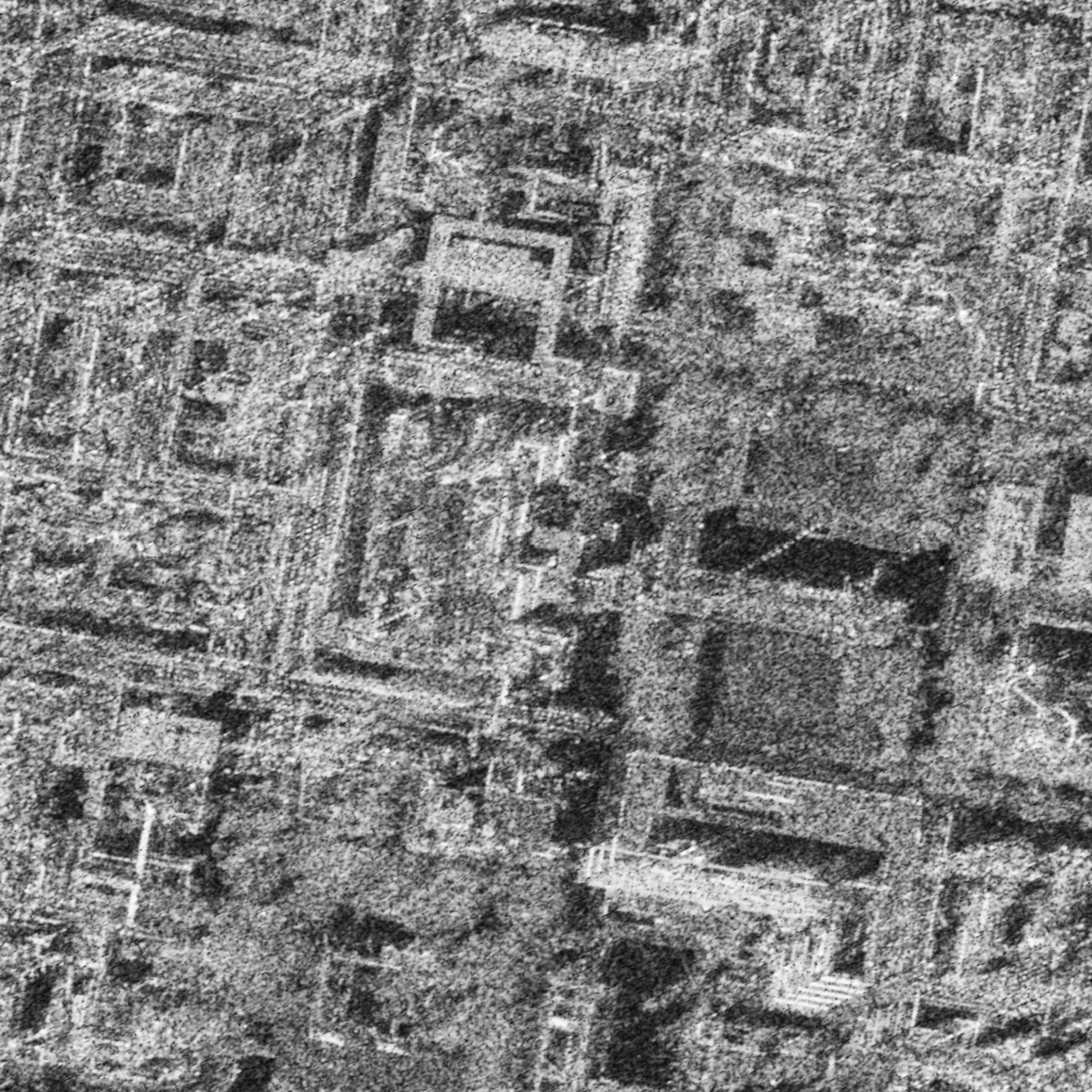}} \hspace{.5mm}
		\subfloat[\label{fig:sar_comp_b}Munich: StripMap]{\includegraphics[width=.24\textwidth]{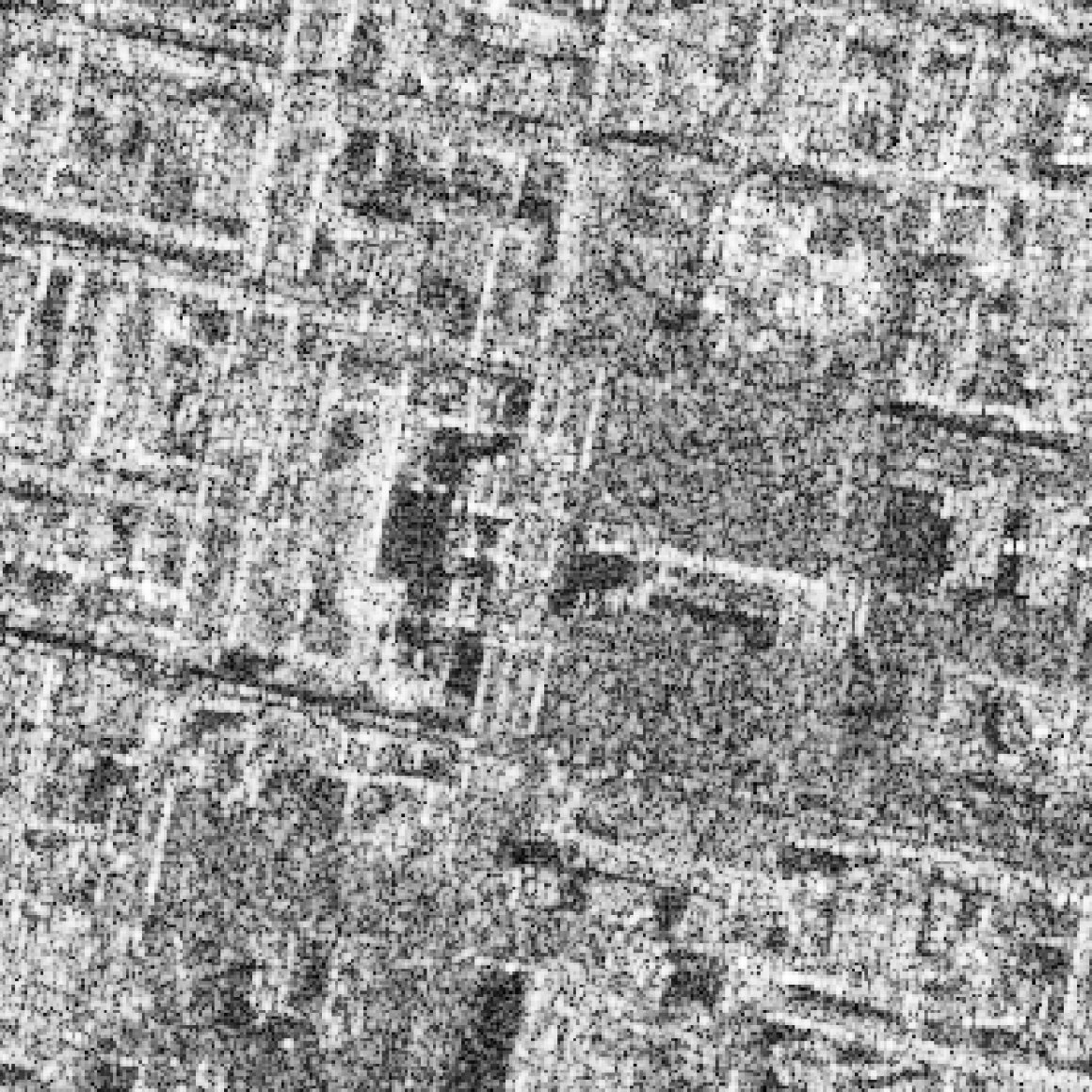}} \hspace{2.5mm}
		\subfloat[\label{fig:sar_comp_c}Berlin: Staring SpotLight]{\includegraphics[width=.24\textwidth]{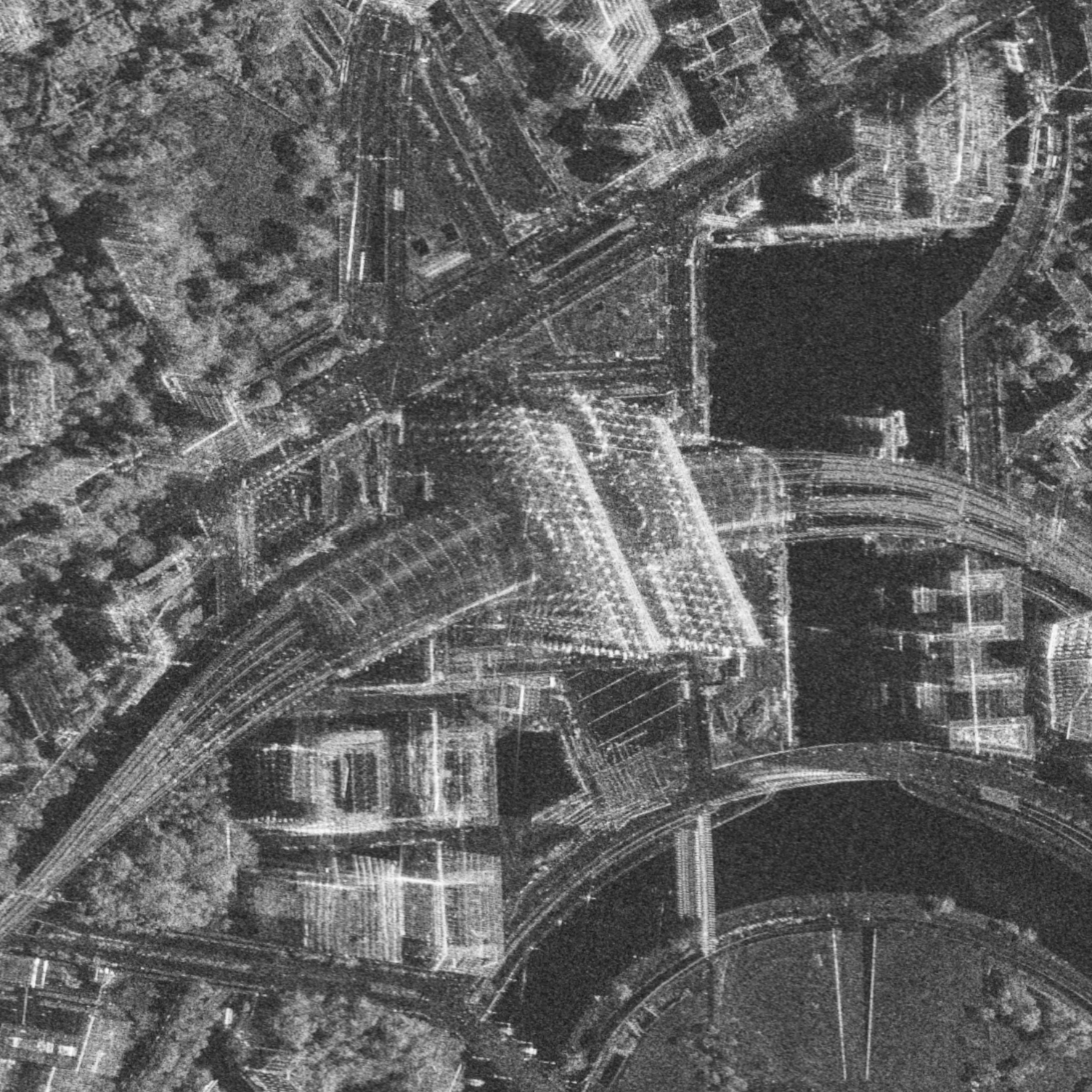}} \hspace{.5mm}
		\subfloat[\label{fig:sar_comp_d}Berlin: StripMap]{\includegraphics[width=.24\textwidth]{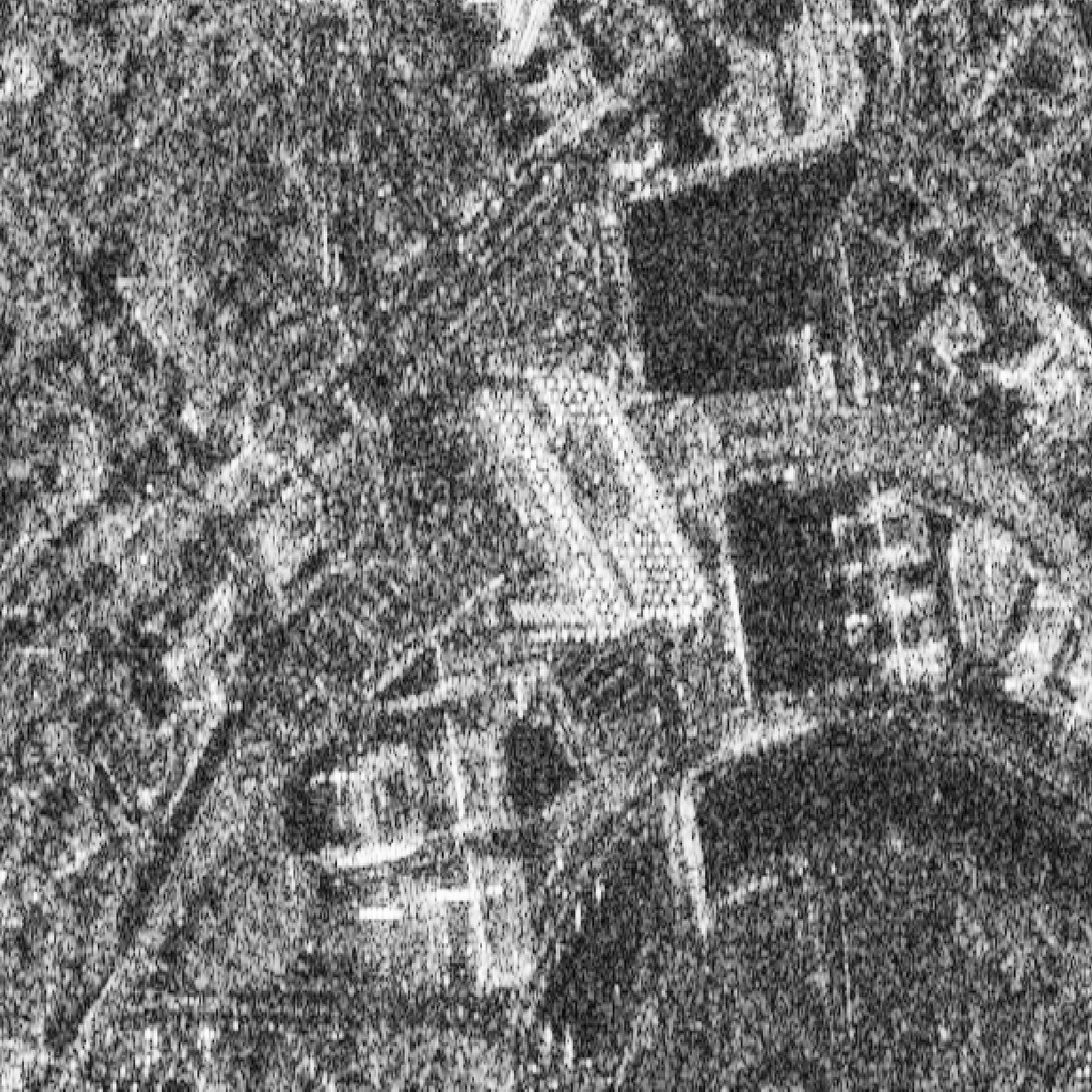}}
		\endgroup
		\caption{Comparison between the different SAR image data types. (a) and (b): area around the Technical University in Munich; (c) and (d): Central Station in Berlin. The higher resolution and the smaller amount of speckle of the HS300 and ST data is clearly visible.}
		\label{fig:sar_comp}
	\end{center}
\end{figure}

\begin{figure}[]
	\begin{center}
		\begingroup
		\subfloat[Area around Munich]{\includegraphics[width=.3\linewidth]{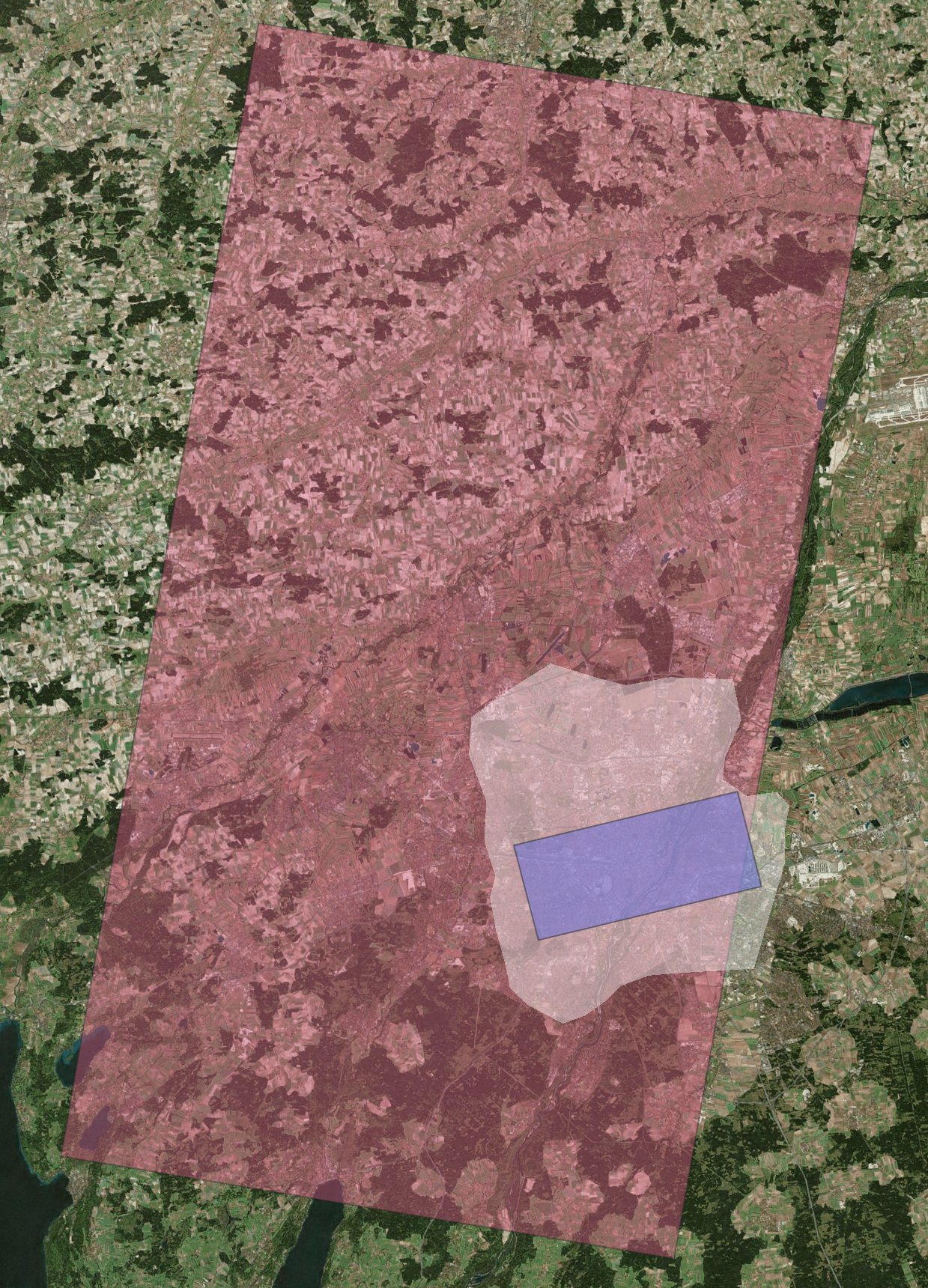}} \hspace{3mm}
		\subfloat[Area around Berlin]{\includegraphics[width=.3\linewidth]{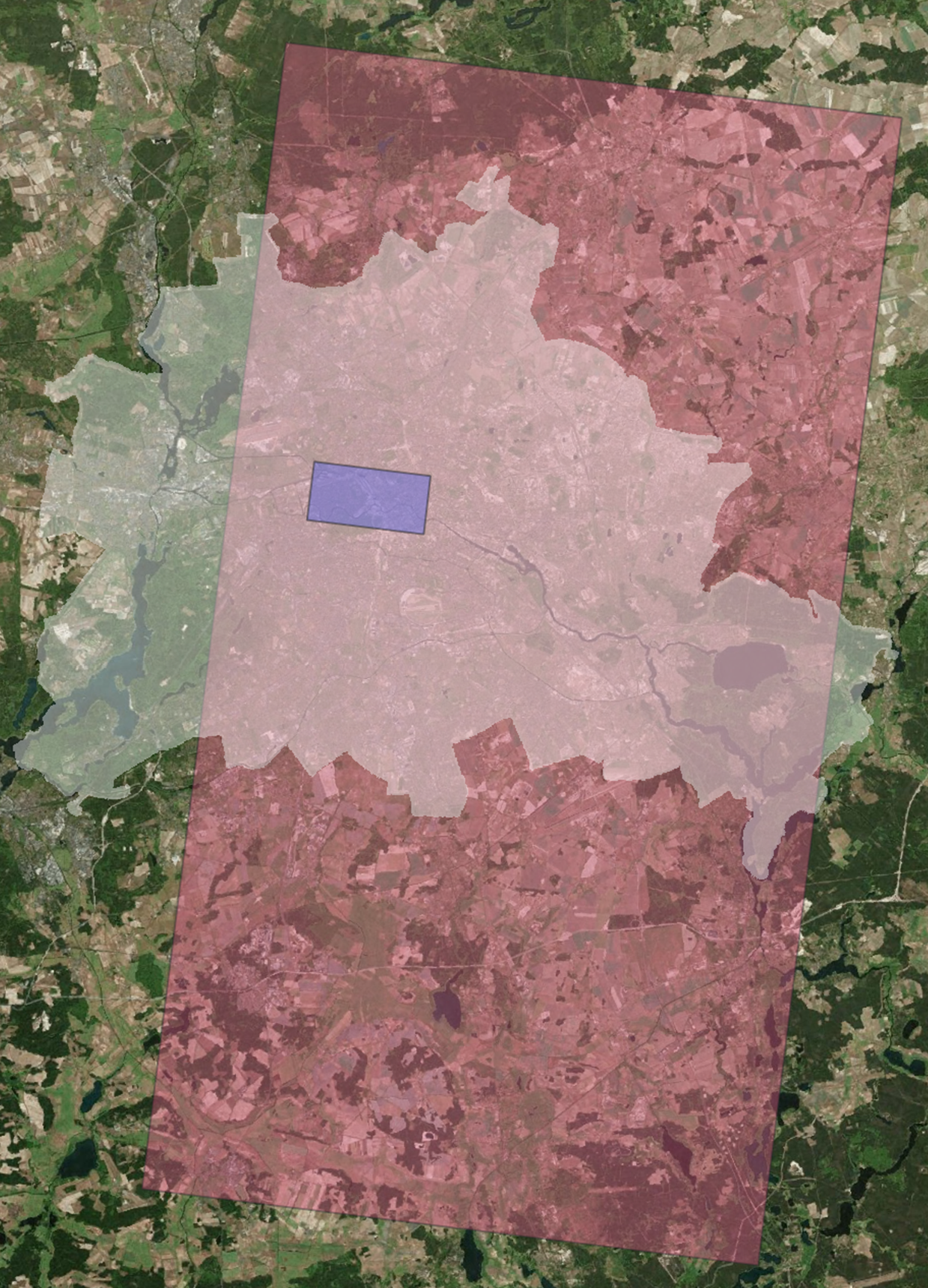}}
		\endgroup
		\caption{Size comparison of the available SAR images. The red shaded areas are covered by the Stripmap data, the blue ones by the Spotlight data. The white zones show the regions for which elevation data are available. [Background satellite images from \textit{Bing Maps}.]}
		\label{fig:size_comparison}
	\end{center}
\end{figure}

For the experiments documented in this paper, we have made use of four TerraSAR-X image products. Those images cover the cities of Munich and Berlin in Germany. For the Munich test area, both an SM and a HS300 image are available. For the Berlin test area, besides an SM image also an ST image was used. Table~\ref{tab:sar_images} gives an overview of the data. As can be seen in Figure~\ref{fig:size_comparison}, the images differ greatly in their extent. The spotlight data only represent the downtown areas, while the stripmap data cover also the city outskirts. Since this work focuses on the height estimation in urban areas, the stripmap data are not used to their full extent. Figure~\ref{fig:sar_comp} shows a comparison of downtown subsets of the four images to provide a feeling for the qualitative differences of the individual imaging modes. 

\begin{table}
	\caption{Properties of the SAR images used in this work.}
	\begin{center}
		\begin{tabular}{ m{.2\linewidth} m{.15\linewidth} m{.17\linewidth} m{.15\linewidth} m{.15\linewidth} }
			\toprule
			 \textbf{Property}& \multicolumn{2}{c}{\textbf{Munich}} & \multicolumn{2}{c}{\textbf{Berlin}} \\ 
			 \cmidrule(r){1-1}\cmidrule(lr){2-3} \cmidrule(l){4-5}
			 Imaging Mode & Stripmap (SM) & Spotlight (HS300) & Stripmap (SM) & Spotlight (ST) \\ 
			 Covered Area & 32 x 60 km & 12 x 5 km & 32 x 60 km & 6x3 km \\ 
			 Azimuth Res. & 3.3 m & 1,1 m & 3.3 m & 0.24 m \\ 
			 Slant Range Res. & 1.2 m & 0.6 m & 1.2 m & 0.6 m  \\ 
			 Ground Range Res. & 2.5 -- 2.9 m & 1.5 -- 1.6 m & 2.0 -- 2.1 m & 1.0 m  \\ 
			 Incidence Angle & 24.7 -- 28.0 deg & 22.4 -- 23.6 deg & 34.0 -- 36.7 deg & 35.9 -- 36.4 deg  \\ 
			 Pixels in Azimuth & 30326 & 5976 & 31660 & 18298 \\ 
			 Pixels in Range & 15918 & 10332 & 20460 & 7918 \\ 
			 Orbit direction & descending & ascending & descending & descending \\
            \bottomrule
		\end{tabular}
	    \label{tab:sar_images}
	\end{center}
\end{table}

\subsection{Height Data}\label{sec:HeightData}
For Munich, the height reference data in this study is available in the form of a 3D point cloud derived from airborne laser scanning. The point cloud comes with a point density of more than 1 point per $\text{m}^2$. In order to make the data fit for the height projection approach described in Section~\ref{sec:HeightProjection}, the 3D point cloud first has to be reduced to a 2.5D height map. For that, the point cloud is sampled via ground cells with a cell size of 0.5 m. For every cell, the maximum occurring value within the cell is stored as height value. This ensures that roofs and canopies are pertained, even if there are data points underneath. On the other hand, this sampling bears the consequence that also street lamps, signs and any other small elevated objects show up as pillars in the result. To counteract this effect somewhat, the grid is finally cleaned with a median filter. The point clouds are originally provided in the UTM coordinate system with normal heights. Since later all calculations are made in a geocentric system, the coordinates have to be transformed, leading to geometrically defined ellipsoidal heights. For the transition to ellipsoidal heights, a constant value for the geoid height in Munich was used, since the variations of the geoid are very small in that area. 

For the entire area of the Federal State of Berlin, a digital surface model with a ground resolution of 1 m is freely available and provided by the Senate Department for Urban Development and Housing Berlin. It is a photogrammetry-based model calculated from the imagery acquired by an aerial survey. The quality of the data is comparable to those from the laser scan in Munich -- maybe even slightly better, as it is cleared of distracting objects like construction cranes. Again, the normal heights must be converted to ellipsoidal heights. For this purpose, the geoid undulation was determined with a transformation tool\footnote{Trans3win 2016, a freely available transformation tool provided by the Senate Department for Urban Development and Housing Berlin} for some points and averaged over the whole area.

\subsection{ML-Ready Dataset}
After combining the SAR imagery described in Section~\ref{sec:SARimageData} with the height data described in Section~\ref{sec:HeightData} using the projection approach proposed in Section~\ref{sec:HeightProjection}, still a machine learning-ready dataset has to be produced. This includes a calibration of the image data, the removal of areas not covered by both height and imagery, the tiling of the original images into smaller patches digestible by most CNN architectures, and the splitting of the dataset into training, validation and test subsets. 

\subsubsection{Image Calibration}
In general, the intensity values of a SAR image follow an exponential distribution. This is unfavorable for a model as input variable. To approximate a normal distribution, intensity data is commonly converted to a logarithmic scale in SAR image processing. The so-called radar brightness $\beta^0$ is given in dB and represents the reflectivity per unit area in slant range. To counteract various influences like different incidence angles, ascending-descending geometries or opposite look directions, a calibration factor $k_s$ is applied. Thus, the radar brightness in dB of a complex SAR pixel $u$ is given by:
\begin{equation}
    \beta^0 = 10\cdot\log_{10}\left(k_s\cdot|u|^2\right)
\end{equation}
The calibration factor is provided as part of the metadata of the TerraSAR-X products. After calibration, the backscatter values are clipped to the interval $[-30;+10]$~dB and subsequently normalized to the interval $[0;1]$.

\subsubsection{Creation of the Data Splits}
\begin{figure}
	\begin{center}
		\begingroup
		\subfloat[]{\includegraphics[width=.45\linewidth]{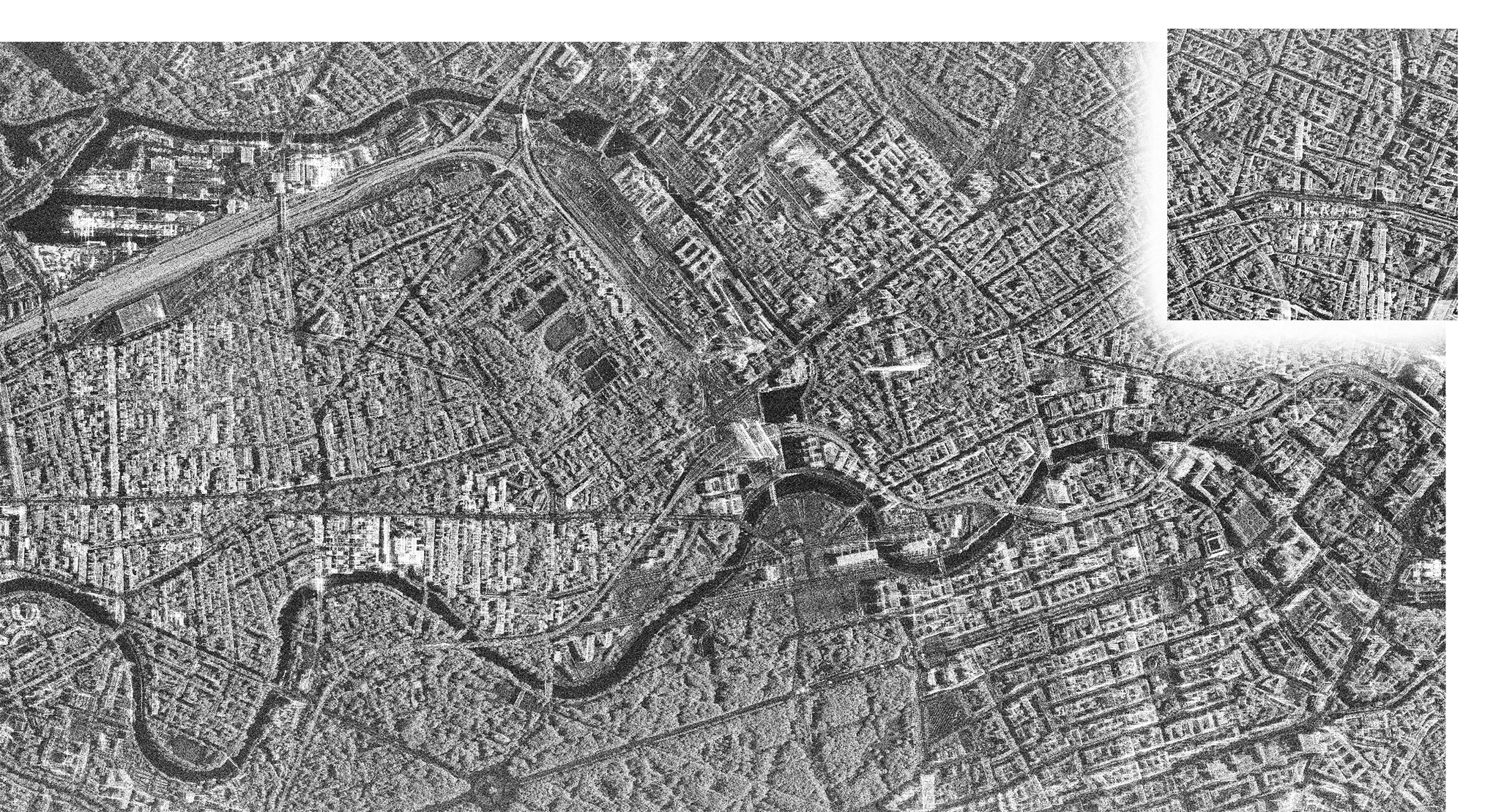}} \hspace{2mm}
		\subfloat[]{\includegraphics[width=.45\linewidth]{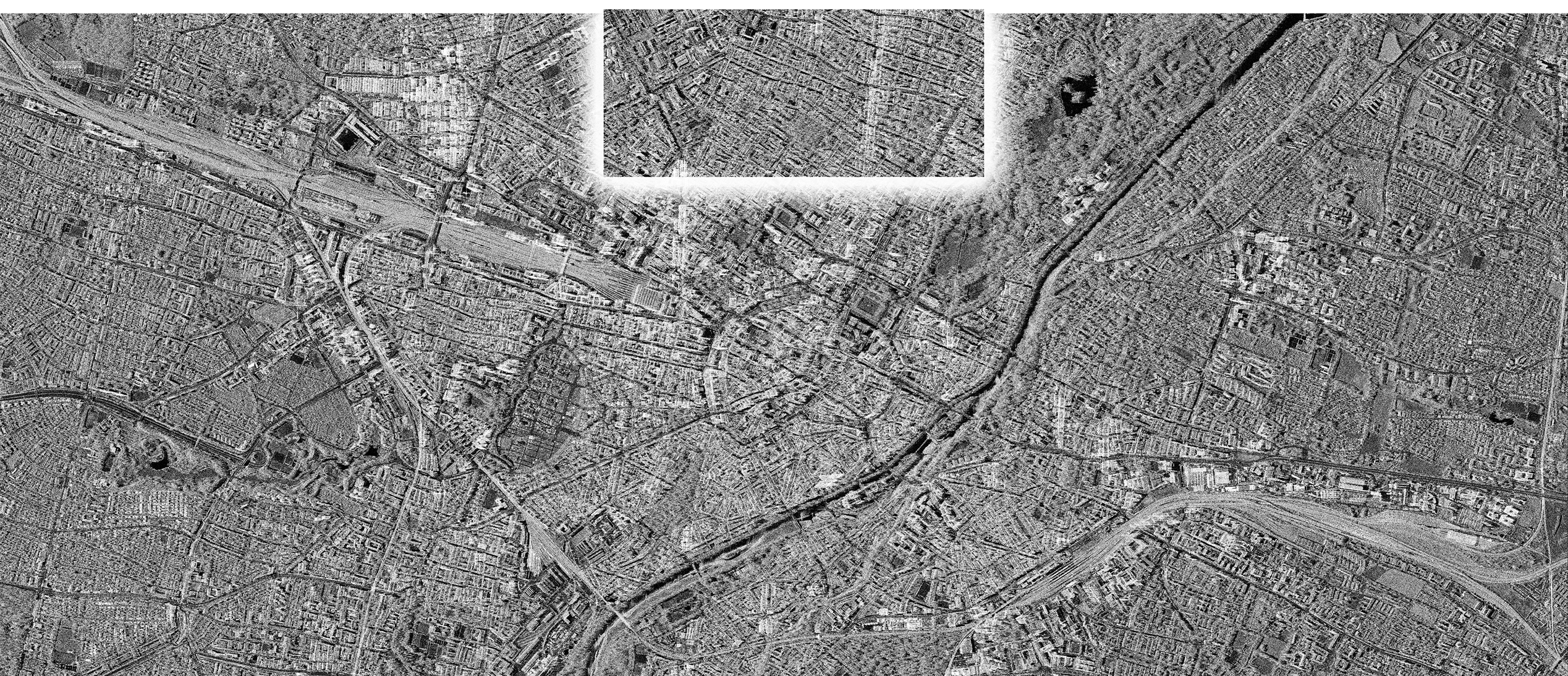}} \\
		\subfloat[]{\includegraphics[width=.39\linewidth]{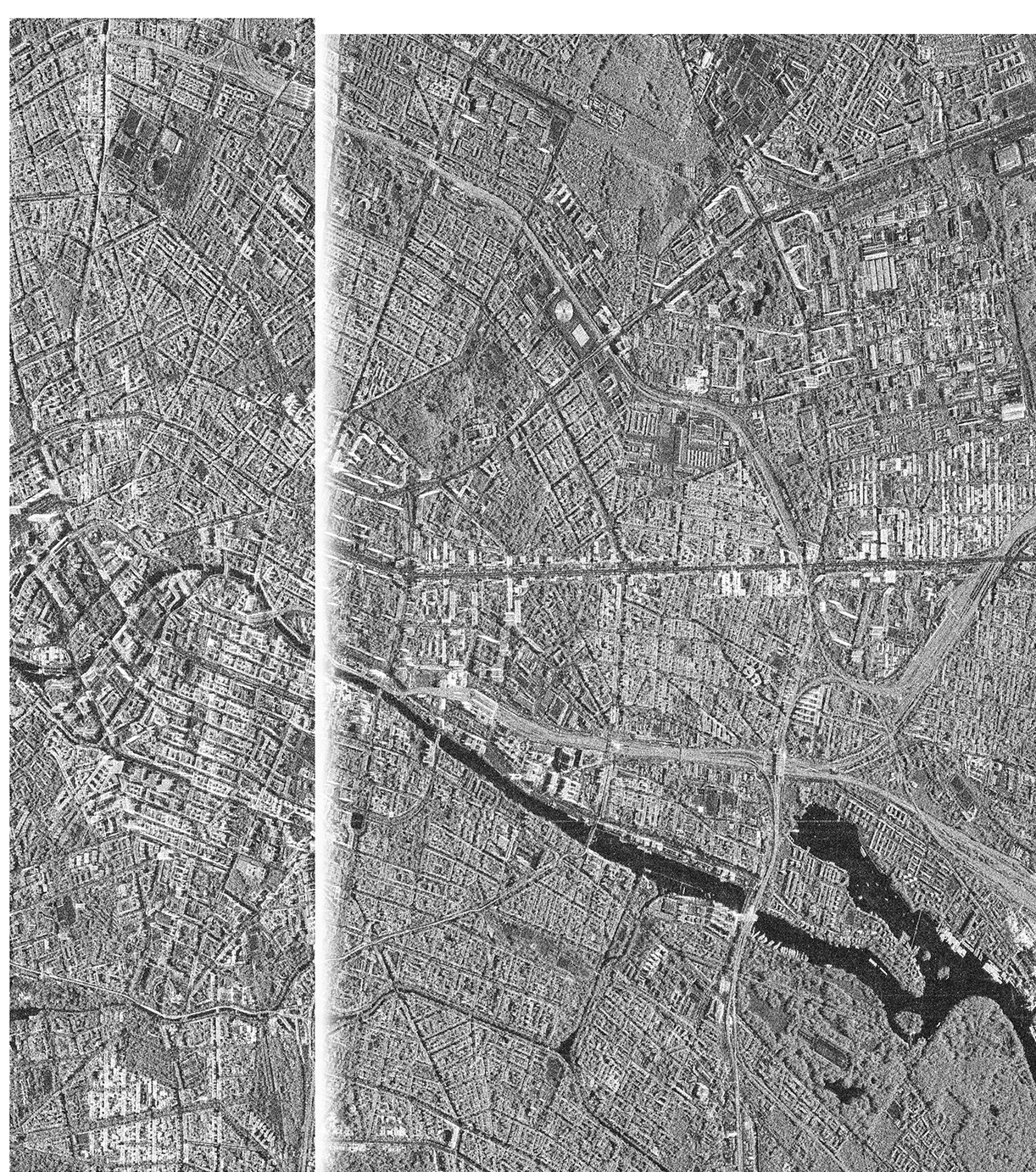}} \hspace{10mm}
		\subfloat[]{\includegraphics[width=.40\linewidth]{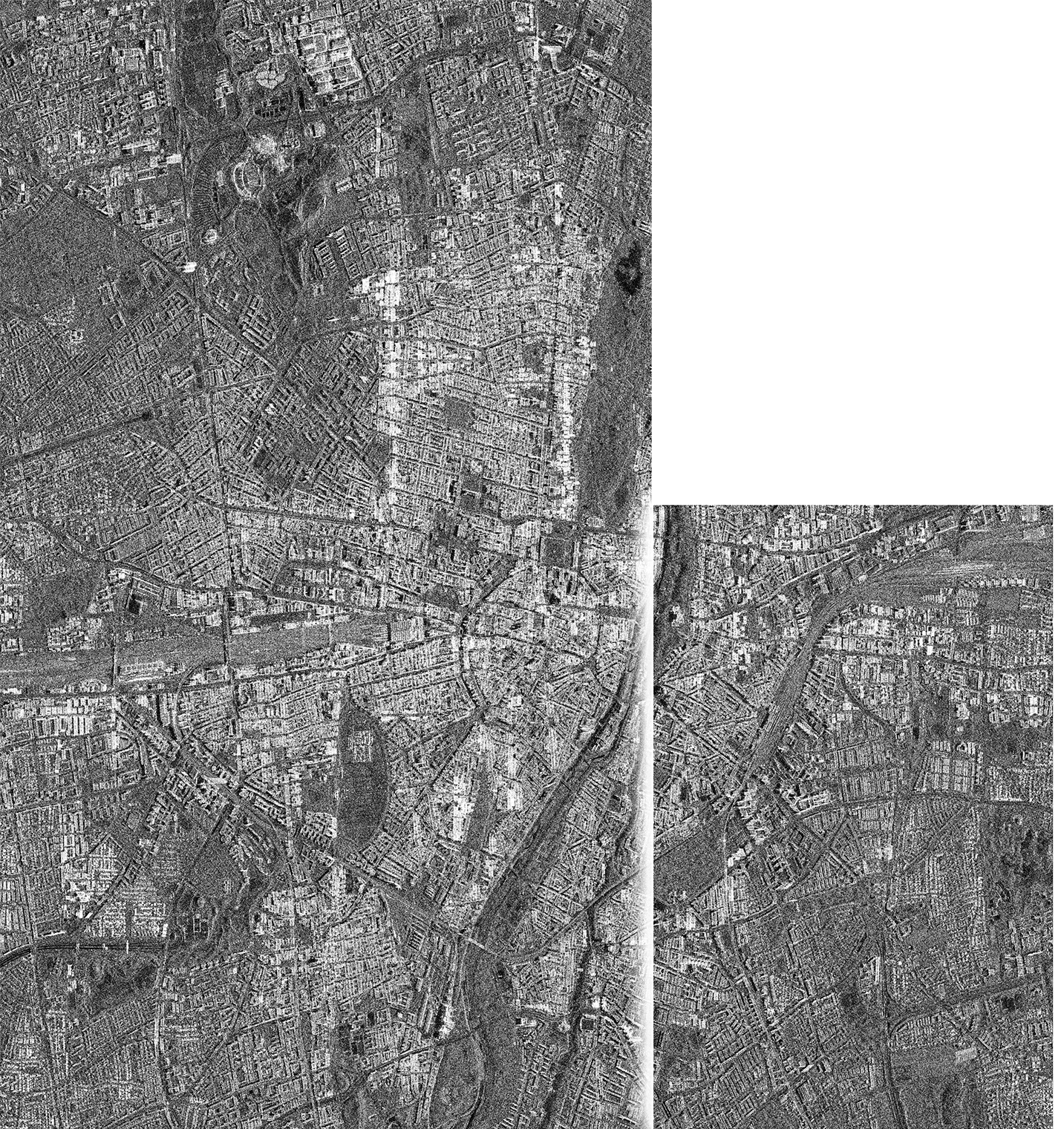}}
		\endgroup
		\caption{Illustration of the spatial split between intra-scene training and test data. (a) Berlin ST, (b) Munich HS300, (c) Berlin SM, (d) Munich SM. The large parts constitute the areas used for training, the smaller rectangles those for testing in the intra-scene experiments. For display purposes, the images are not shown to scale.}
		\label{fig:splits}
	\end{center}
\end{figure}

As a first step towards the creation of ML-ready data splits, the SAR intensity images and their corresponding height maps are cut into patches of $256 \times 256$ pixels each, with an overlap of 50\% between neighboring patches. Then, a manually selected subset of each image, covering roughly 10\% of the overall area is taken to define the hold-out test set. Here, it was important to us to focus on urban areas and -- mostly in the case of the larger SM imagery -- not on agricultural areas or wastelands. The hold-out subsets for the four test scenes are depicted in Figure~\ref{fig:splits}. The eventually available data splits are summarized in Table~\ref{tab:trainTestSplits}.

\begin{table}[h]
    \caption{Number of image patches per data split.}
    \vspace*{.5mm}
    \centering
    \begin{tabular}{lll}
    \toprule
        \textbf{Dataset} & \multicolumn{2}{l}{\textbf{Number of patches}}  \\
          &  Train Set & Test Set\\
          \cmidrule(r){1-1} \cmidrule(lr){2-2} \cmidrule(l){3-3}
          Berlin ST & 1671 & 100\\
          Berlin SM & 5394 & 2373\\
          Munich HS300 & 5441 & 260\\
          Munich SM & 6153 & 2183\\
          \bottomrule
    \end{tabular}
    \label{tab:trainTestSplits}
\end{table}

\section{Experiments and Results}\label{sec:Experiments}

\subsection{Structure of the Experiments}
In order to get a thorough understanding about the potentials and limits of SAR-based single-image height prediction with a particular focus on transferability across imaging modes and scenes, we have carried out a plethora of experiments. Those are divided as follows: First of all, individual networks were each trained on data from one area (e.g. Munich) using one imaging mode (e.g. HS300). Then, a first test was carried out on hold-out data of the same area and mode (see Figure~\ref{fig:splits} for an illustration of the hold-out situation). This is the simplest case for the model, since the prevailing conditions of the test set are very similar to the training data, both in terms of imagery and surface structure. 

Then, to test the transferability of the models  data from other imaging modes and/or areas, the models are also cross-tested against each other. In order to ensure basic compability, the images must be scaled to a uniform ground sampling distance of 1 m.

\subsection{Quality Metrics}
To evaluate the accuracy of the proposed method we make use of the established metrics of the SIDE literature \citep{Eigen2014, Amirkolaee2019}. In all cases, the height maps projected into the radar slant range geometry, as also used as annotation in the training process, are used as reference. The standard root mean square error is defined as
\begin{equation}
	\text{RMSE} = \sqrt{\frac{1}{N} \sum_{i}(y_i - \hat{y}_i)^2},
\end{equation}
with the estimated height values $\hat{y}_i$ over the reference values $y_i$ and $N$ being the total number of pixels. In addition, we use the log-version of the RMSE, which has become a common metric for regression problems due to its robustness to outliers and its relative nature, leaving the scale of the error irrelevant:
\begin{equation}
	\text{RMSE}_{log} = \sqrt{\frac{1}{N} \sum_{i}\left\vert\log_{10}(y_i + 1) - \log_{10}(\hat{y}_i + 1)\right\vert^2}.
	\label{eq:rmselog}
\end{equation}
Another common metric is the relative error
\begin{equation}
	Rel = \frac{1}{N} \sum_{i} \frac{|y_i - \hat{y}_i|}{|y_i| + 1}
	\label{eq:rel}
\end{equation}
and its logarithmic variant 
\begin{equation}
	Rel_{log} = \frac{1}{N} \sum_{i} \left\vert\log_{10}(y_i + 1) - \log_{10}(\hat{y}_i + 1)\right\vert,
	\label{eq:rellog}
\end{equation}
which both are relative to the size of the item being measured. The occurring ``$+1$''s in  equations (\ref{eq:rmselog}) -- (\ref{eq:rellog}) are a way to avoid divisions by zero. They do not distort the resulting value or only in an equal manner for all of the experiments.

Finally, the delta measure calculates a ratio for each pair of pixels and then counts how many of the pixels are below a certain threshold. The information is given as a percentage value:
\begin{equation}
	\delta_i = \max \left(\frac{y}{\hat{y}}, \frac{\hat{y}}{y}\right) < 1.25^i, i \in \{1, 2, 3\}
	\label{eq:delta}
\end{equation}

While all those metrics are designed to measure the quantitative reconstruction error, it is also interesting to investigate the structural reconstruction quality. For this, we use the structural similarity index measure (SSIM), which is defined as
\begin{equation}
	\text{SSIM} = \frac{(2\mu_y \mu_{\hat{y}} + C_1)(2\sigma_{y\hat{y}} + C_2)}{(\mu_y^2 + \mu_{\hat{y}}^2 + C_1)(\sigma_y^2 + \sigma_{y\hat{y}}^2 + C_2)}.
\end{equation}
The SSIM is a metric to evaluate the qualitative appearance of the images and uses statistical indicators to give a value for the structural similarity of two images  \citep{Wang2004}. Here, $\mu_y$, $\mu_{\hat{y}}$, $\sigma_y$, $\sigma_{\hat{y}}$ and $\sigma_{y\hat{y}}$ are the means, the standard deviations and the covariances of the images. $C_1$ and $C_2$ are small constants (like $10^{-6}$).

\subsection{Results}

\begin{figure}[h!]
	\begin{center}
		\captionsetup[subfigure]{labelformat=empty}
		\begingroup
		\subfloat{\includegraphics[width=.3\textwidth, clip, trim=3cm 1cm 2cm 1cm]{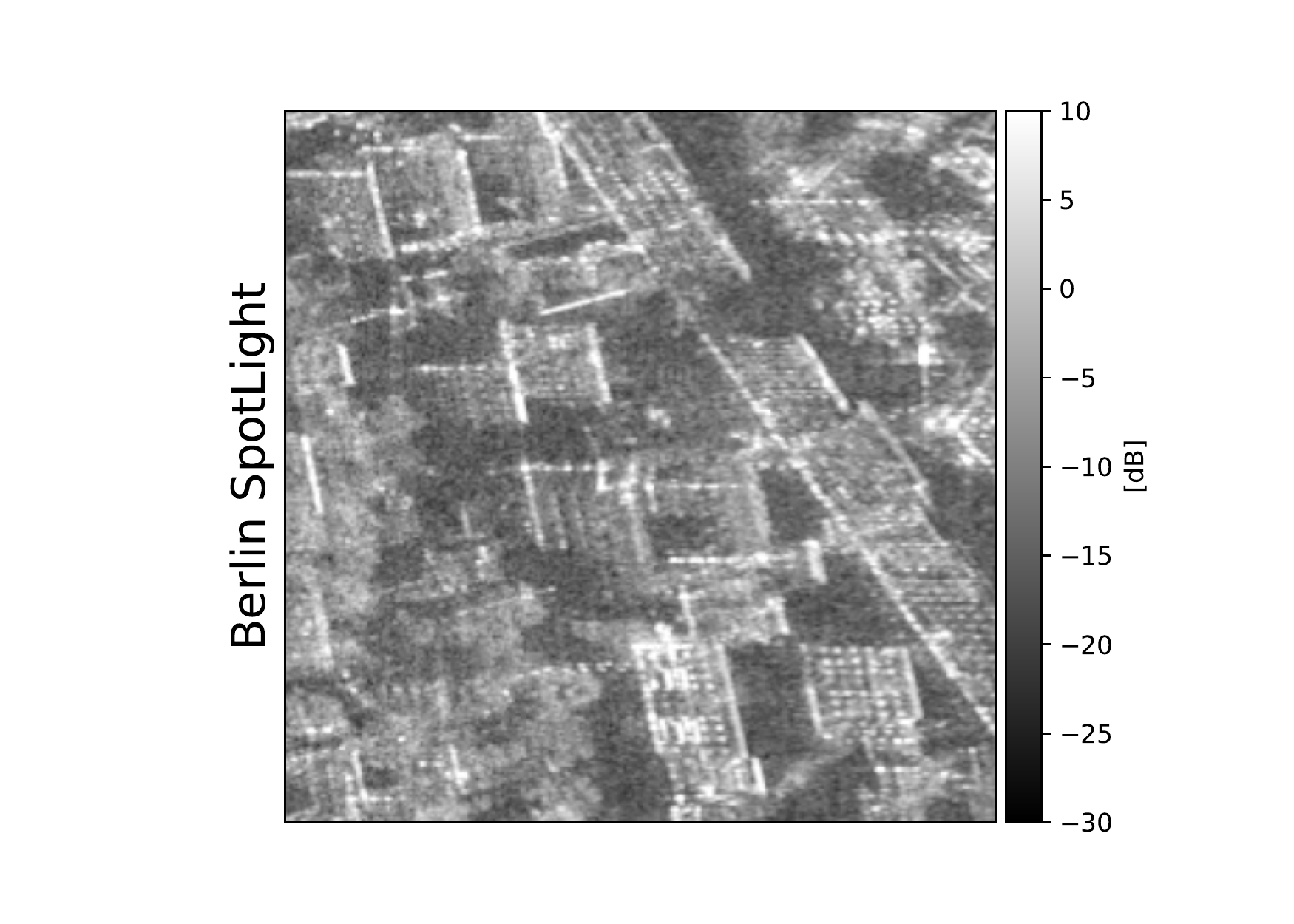}} \hspace{1mm}
		\subfloat{\includegraphics[width=.3\textwidth, clip, trim=3cm 1cm 2cm 1cm]{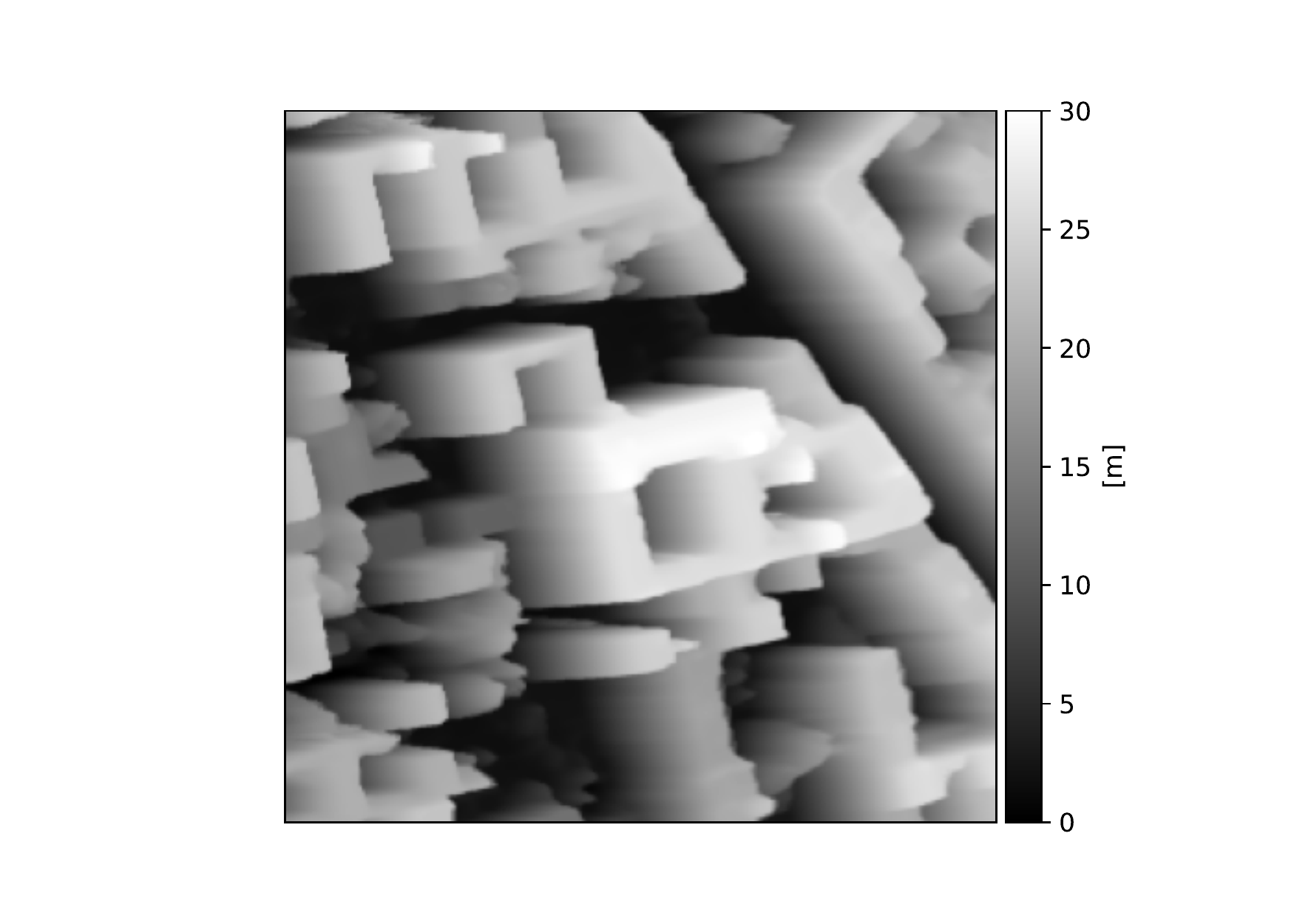}} \hspace{1mm}
		\subfloat{\includegraphics[width=.3\textwidth, clip, trim=3cm 1cm 2cm 1cm]{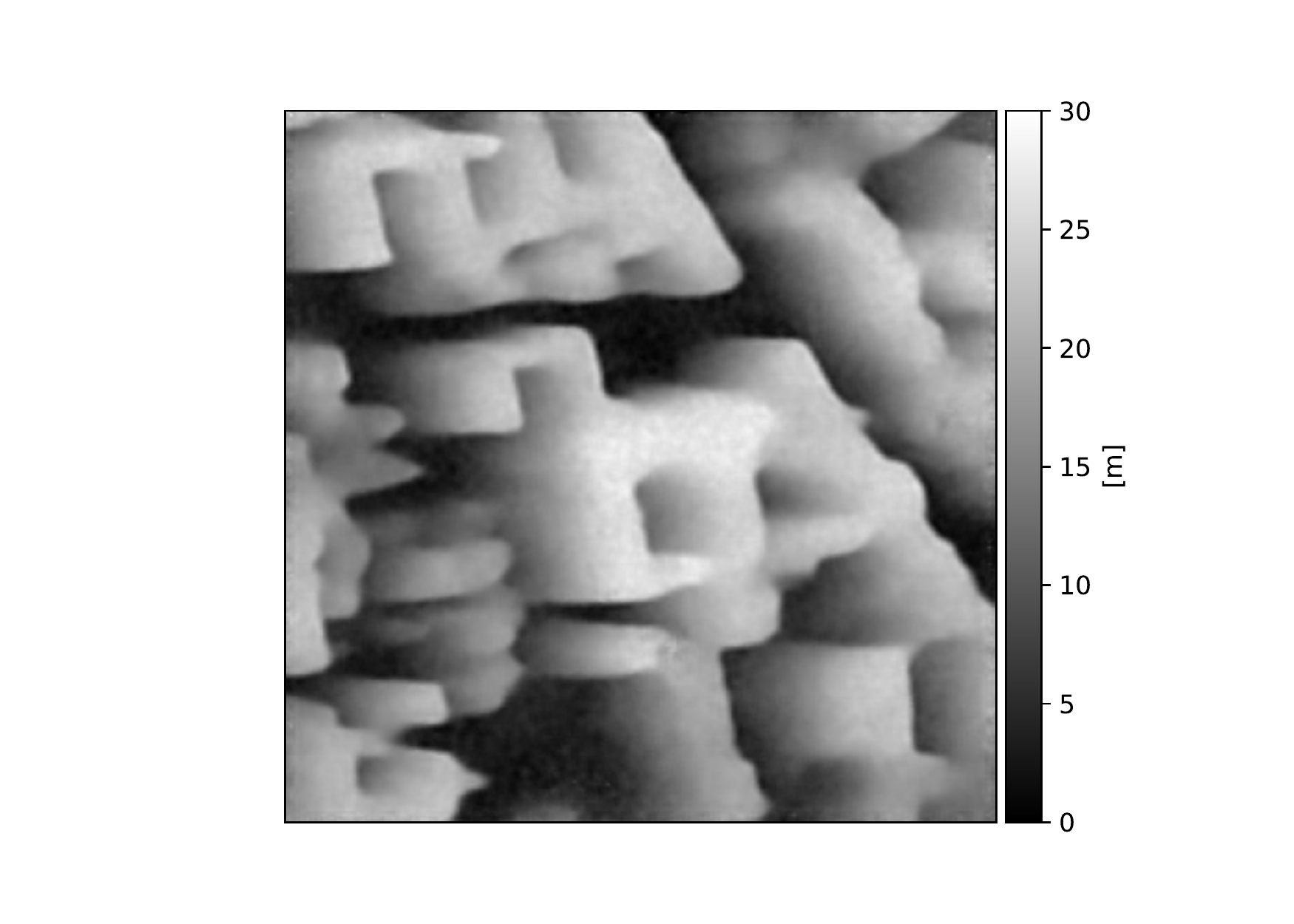}} \\
		\subfloat{\includegraphics[width=.3\textwidth, clip, trim=3cm 1cm 2cm 1cm]{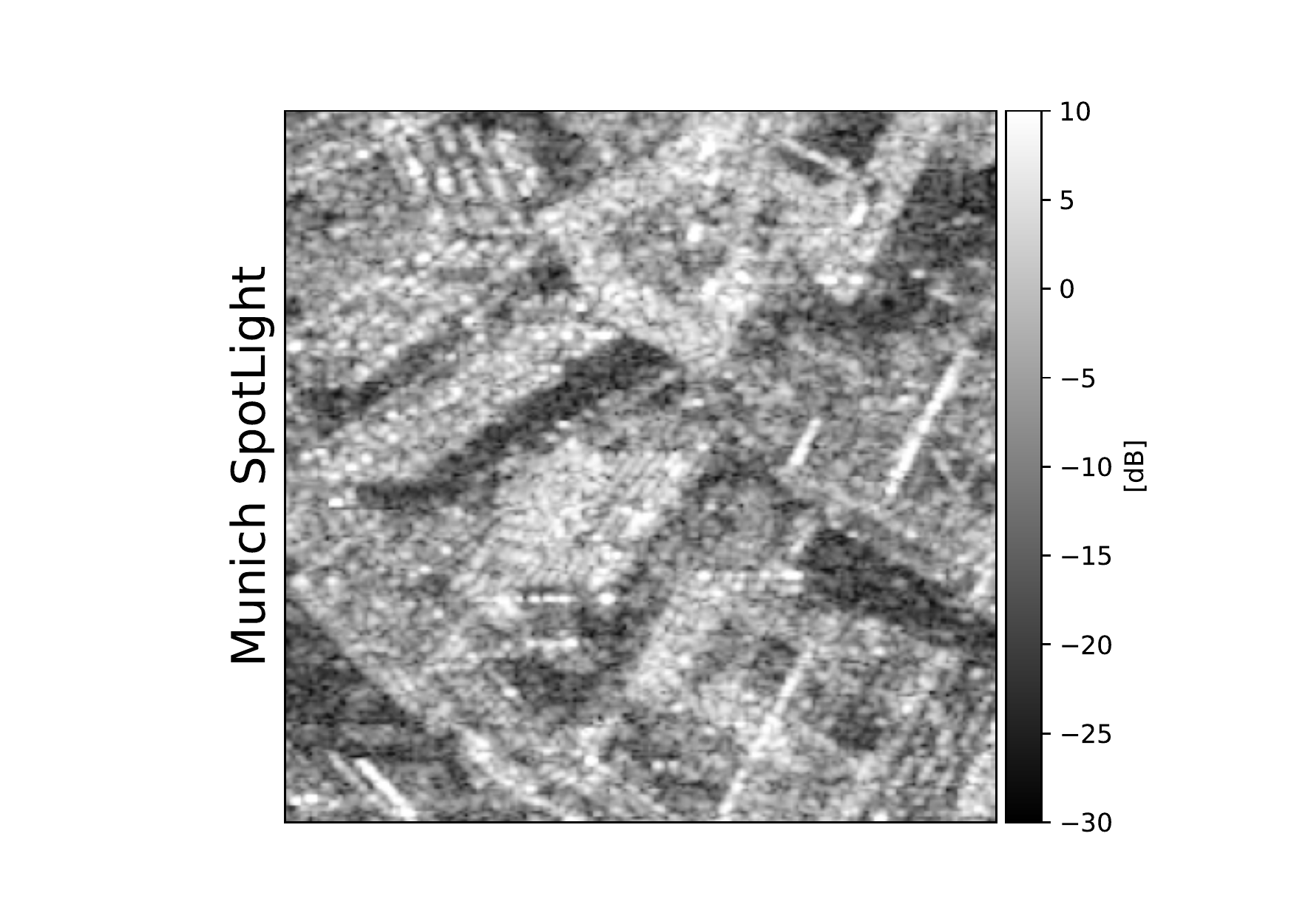}} \hspace{1mm}
		\subfloat{\includegraphics[width=.3\textwidth, clip, trim=3cm 1cm 2cm 1cm]{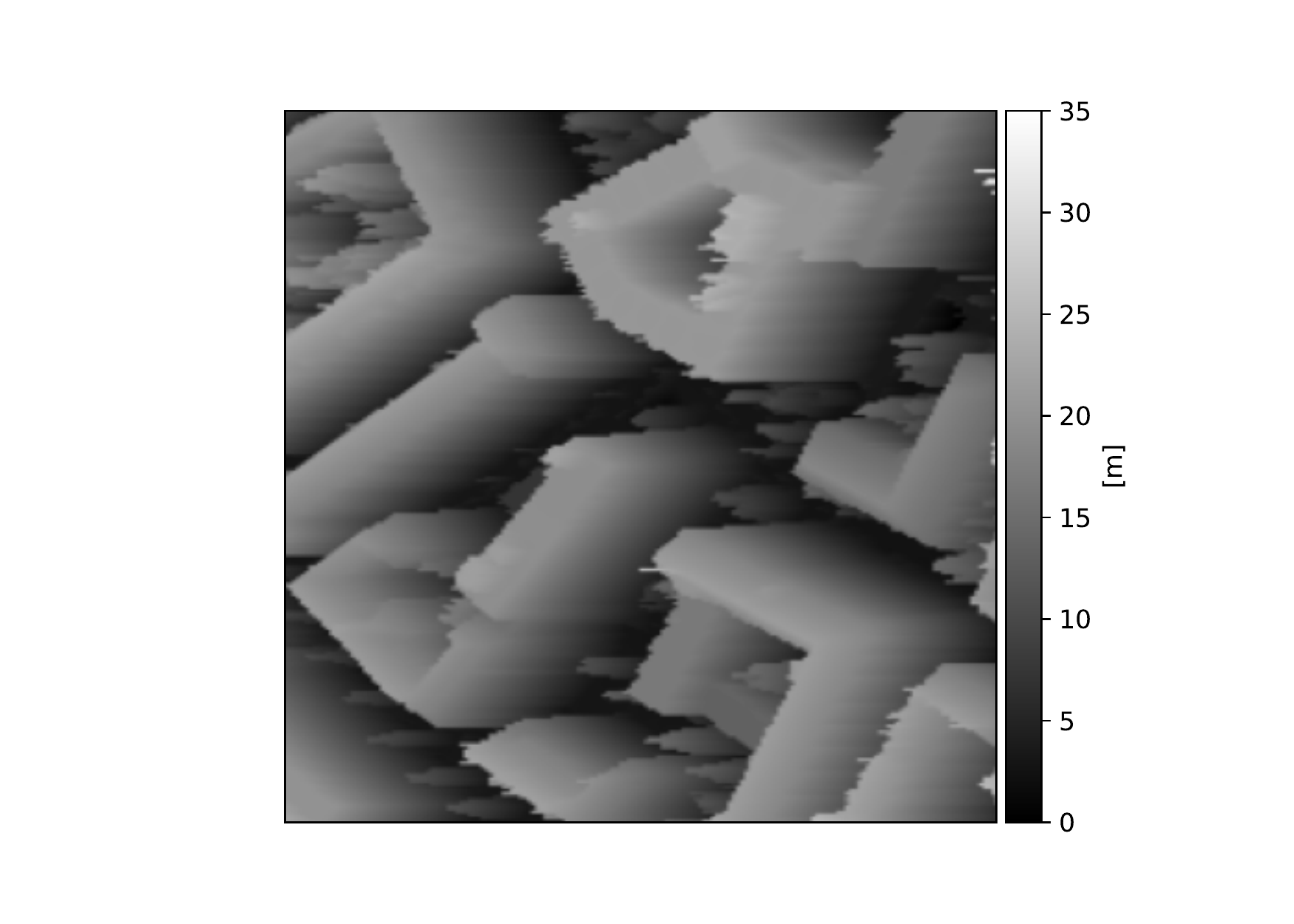}} \hspace{1mm}
		\subfloat{\includegraphics[width=.3\textwidth, clip, trim=3cm 1cm 2cm 1cm]{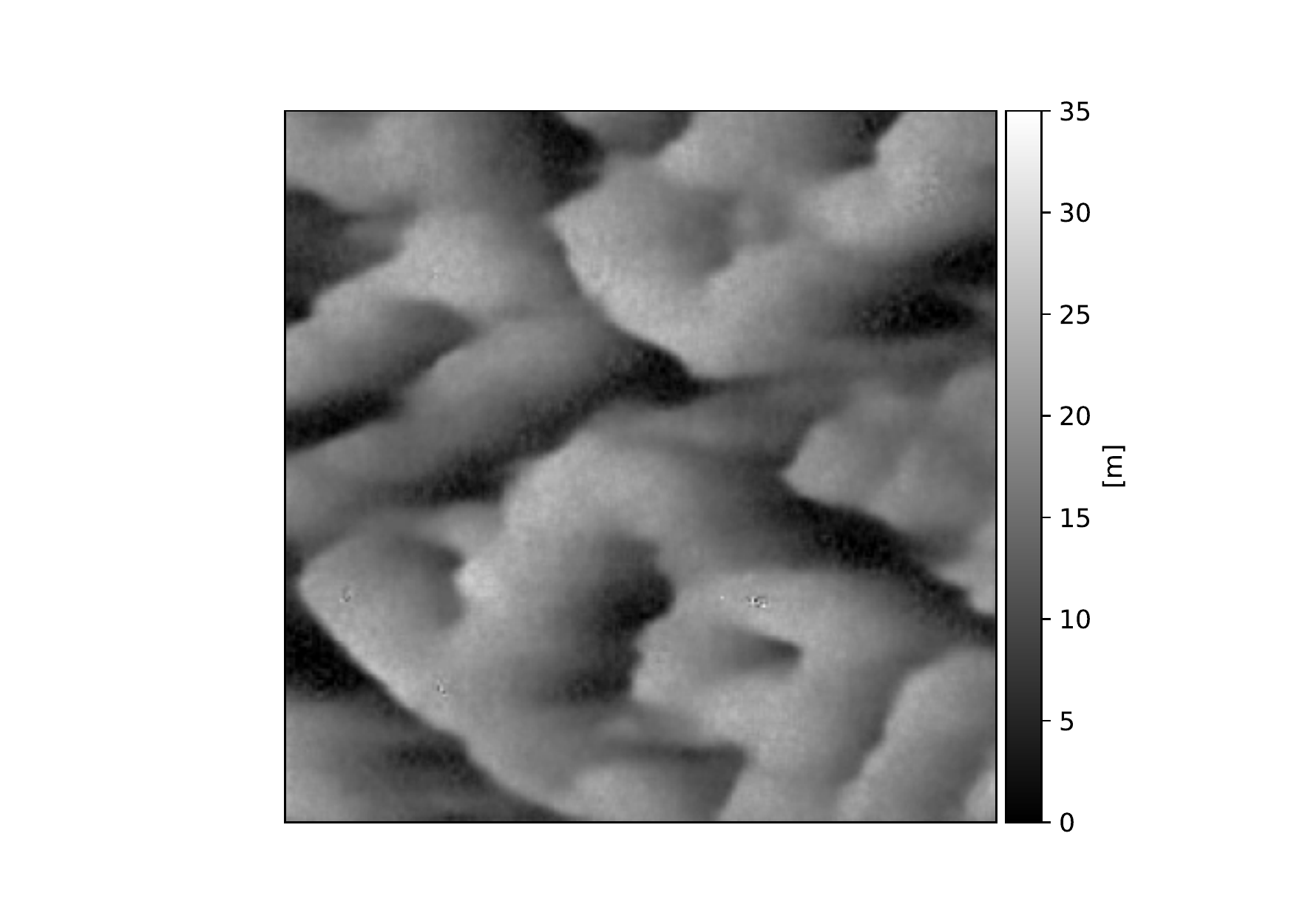}} \\
		\subfloat{\includegraphics[width=.3\textwidth, clip, trim=3cm 1cm 2cm 1cm]{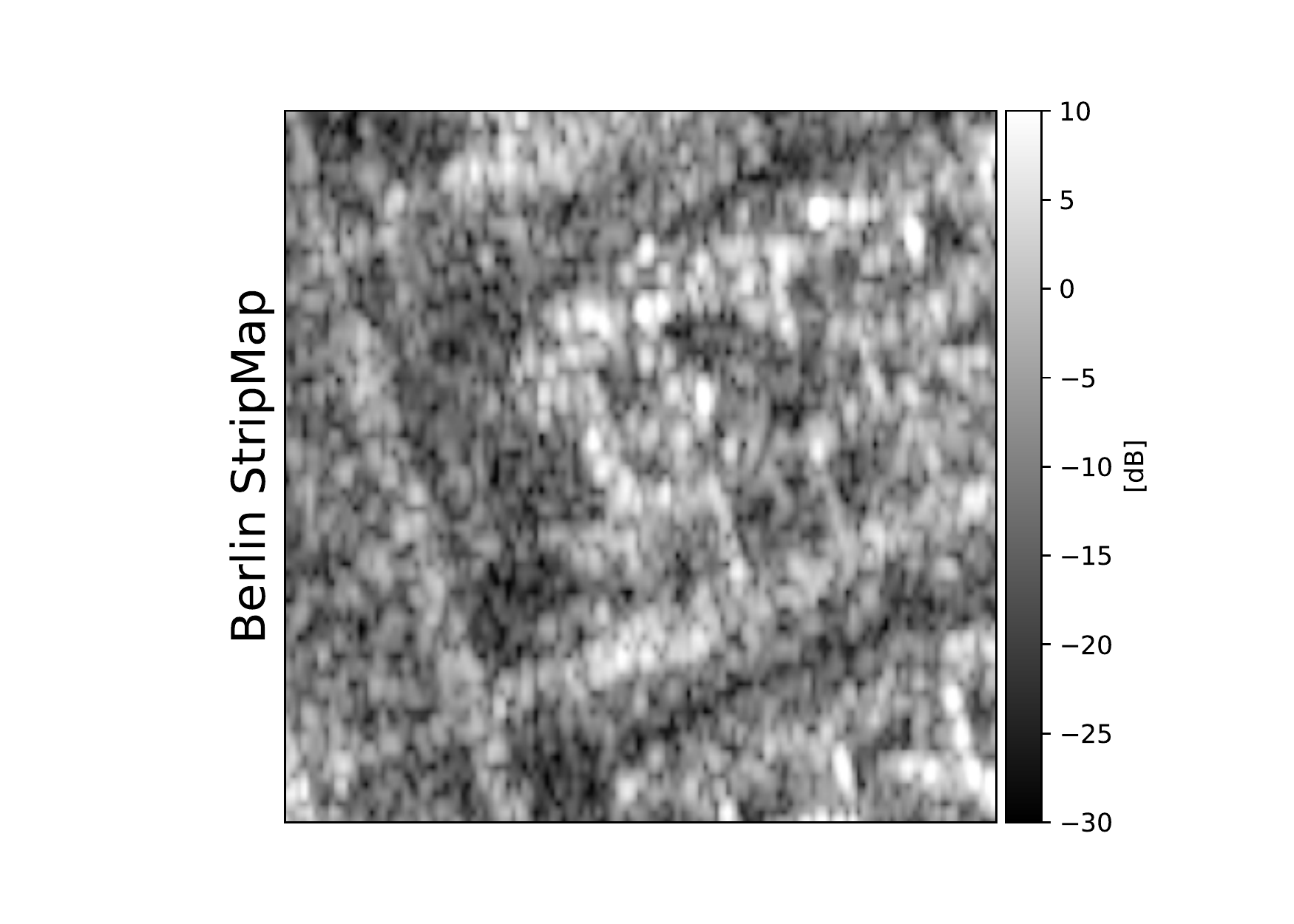}} \hspace{1mm}
		\subfloat{\includegraphics[width=.3\textwidth, clip, trim=3cm 1cm 2cm 1cm]{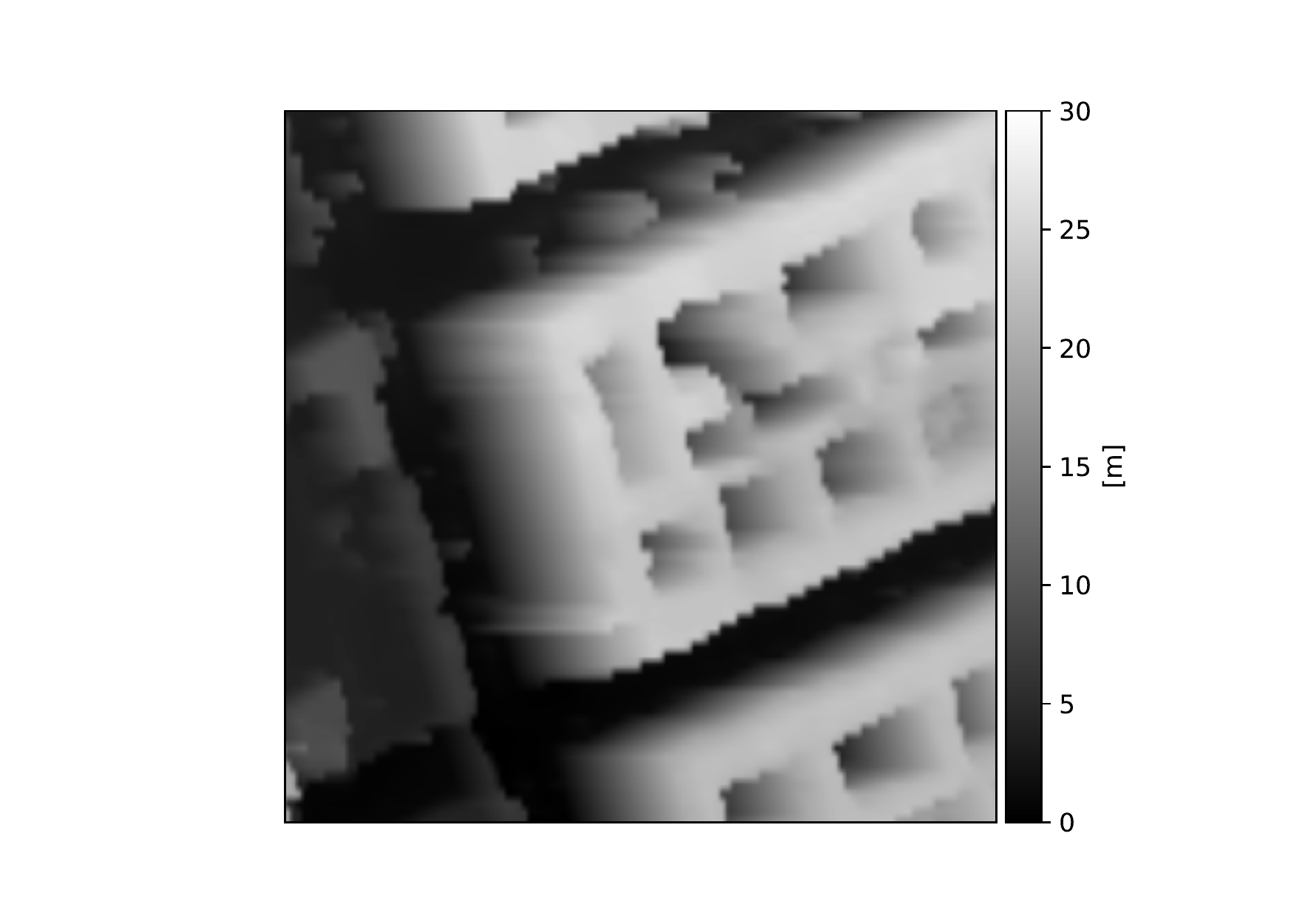}} \hspace{1mm}
		\subfloat{\includegraphics[width=.3\textwidth, clip, trim=3cm 1cm 2cm 1cm]{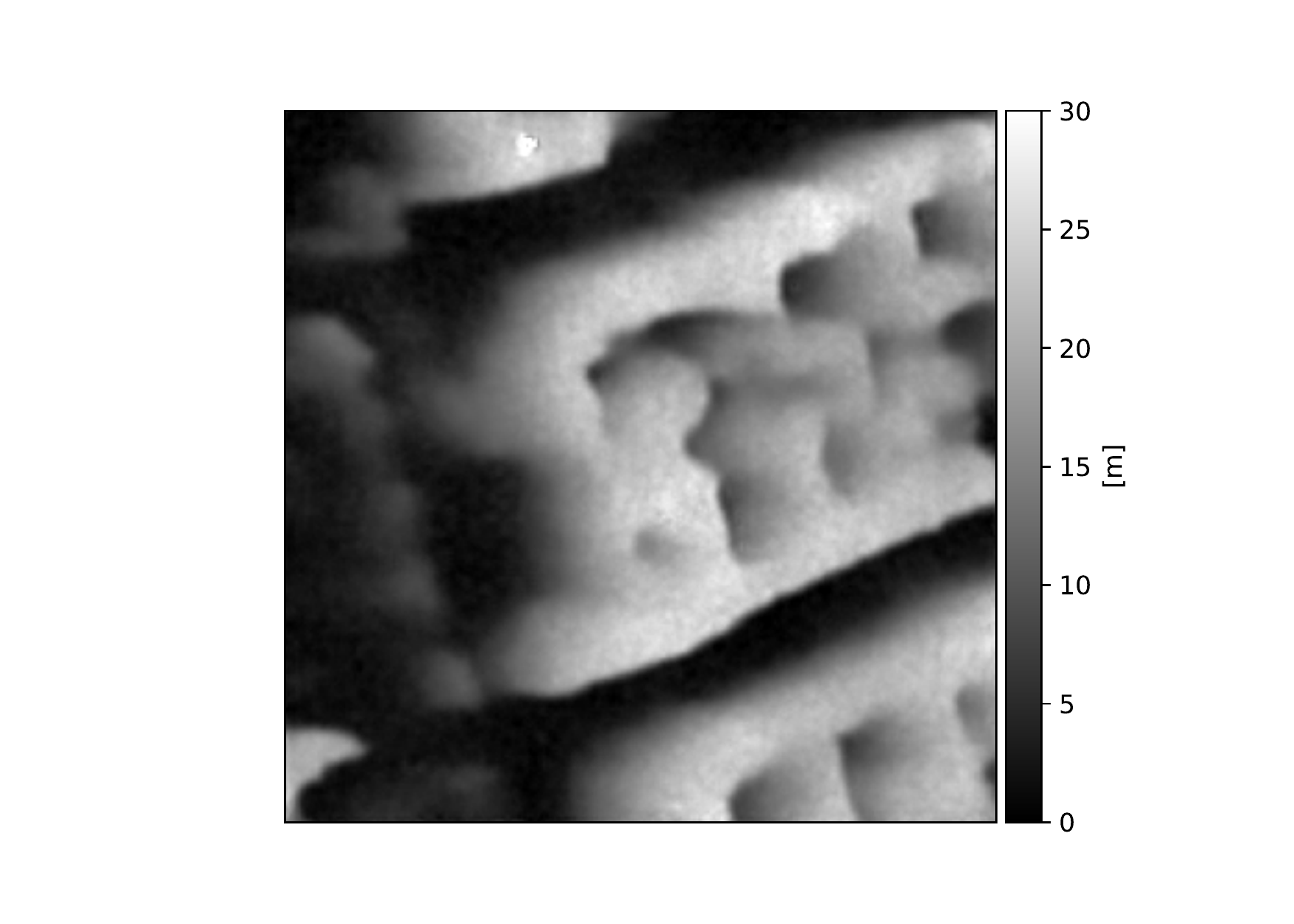}} \\
		\subfloat[SAR image]{\includegraphics[width=.3\textwidth, clip, trim=3cm 1cm 2cm 1cm]{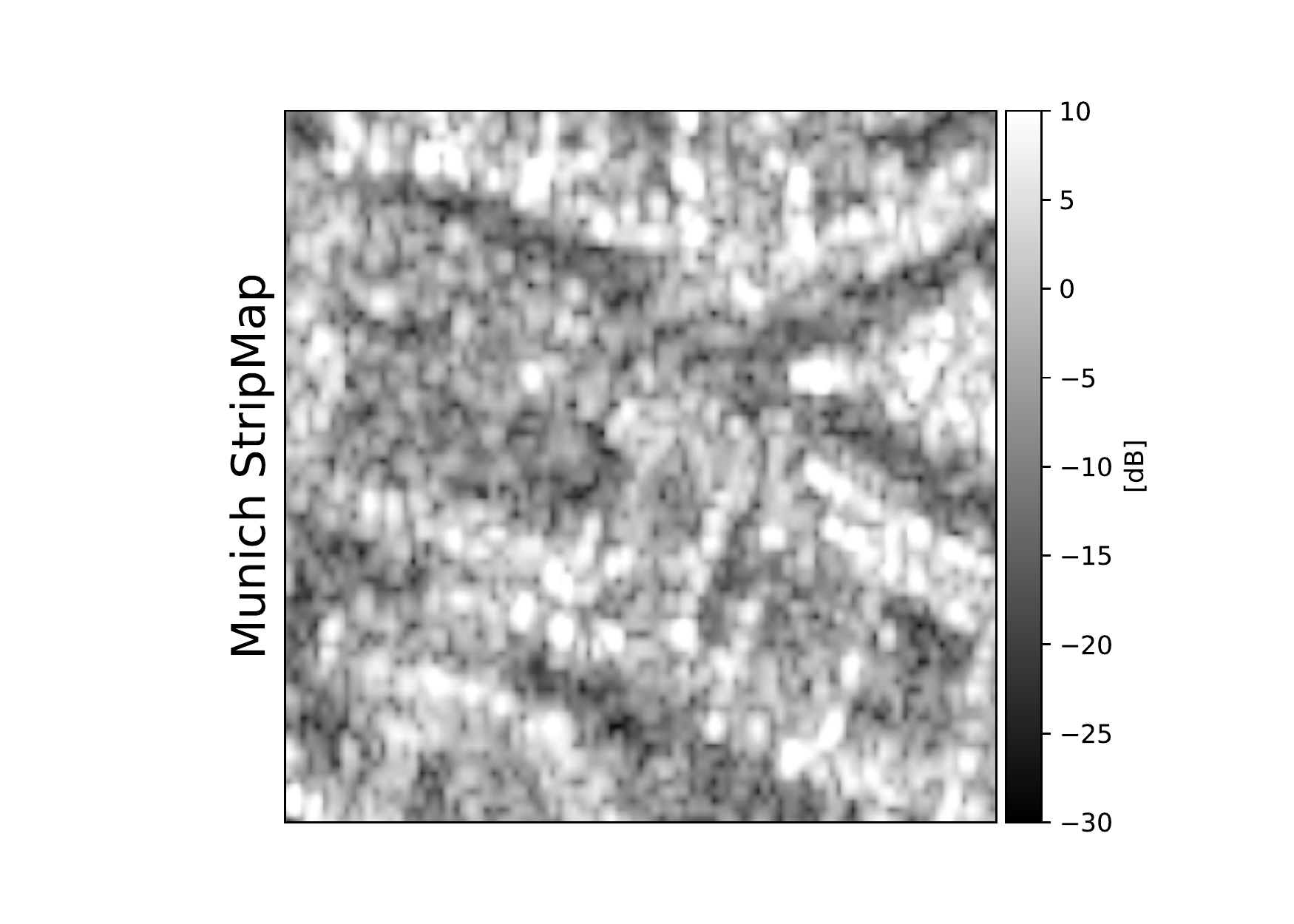}} \hspace{1mm}
		\subfloat[Ground truth]{\includegraphics[width=.3\textwidth, clip, trim=3cm 1cm 2cm 1cm]{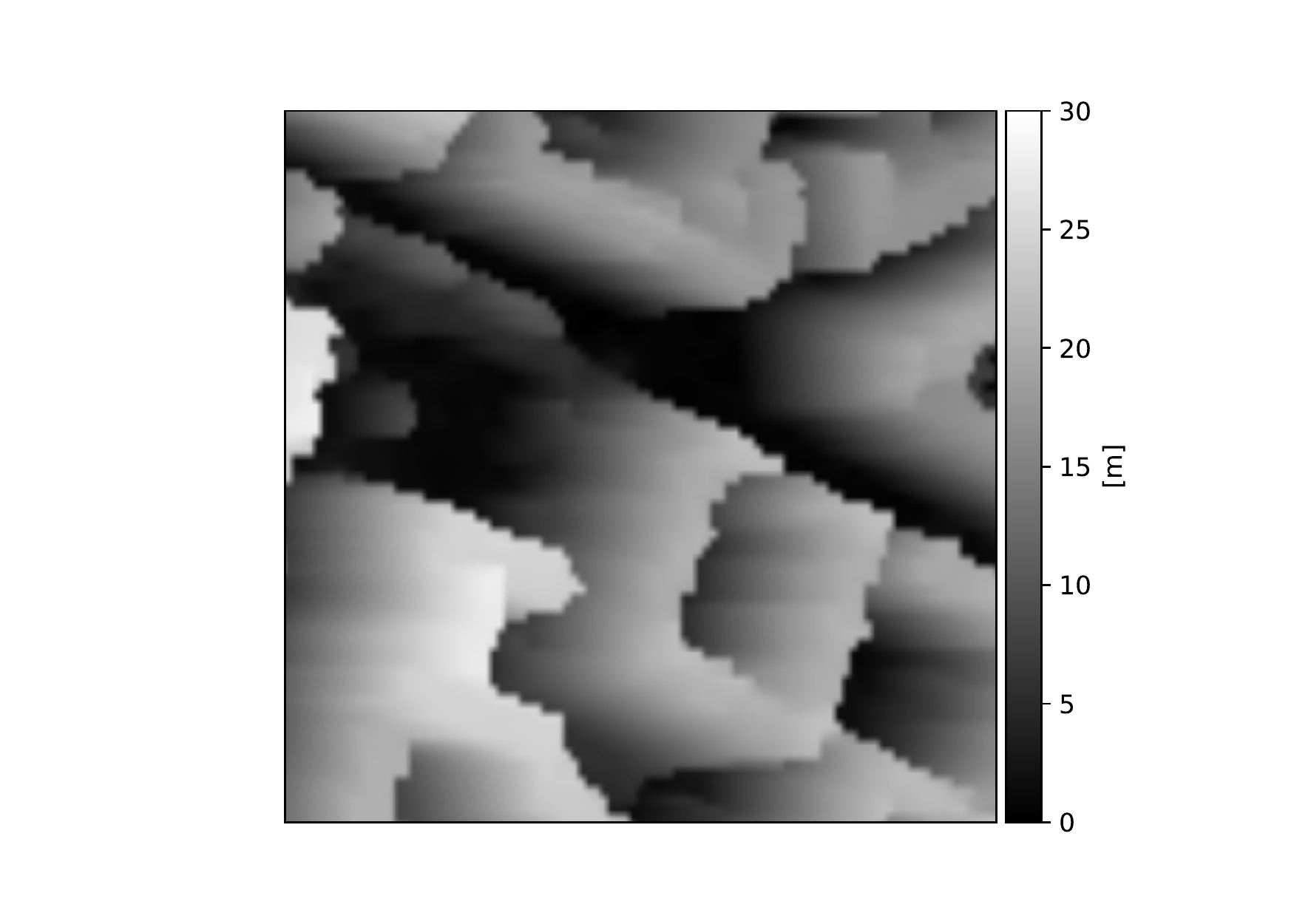}} \hspace{1mm}
		\subfloat[Estimated heights]{\includegraphics[width=.3\textwidth, clip, trim=3cm 1cm 2cm 1cm]{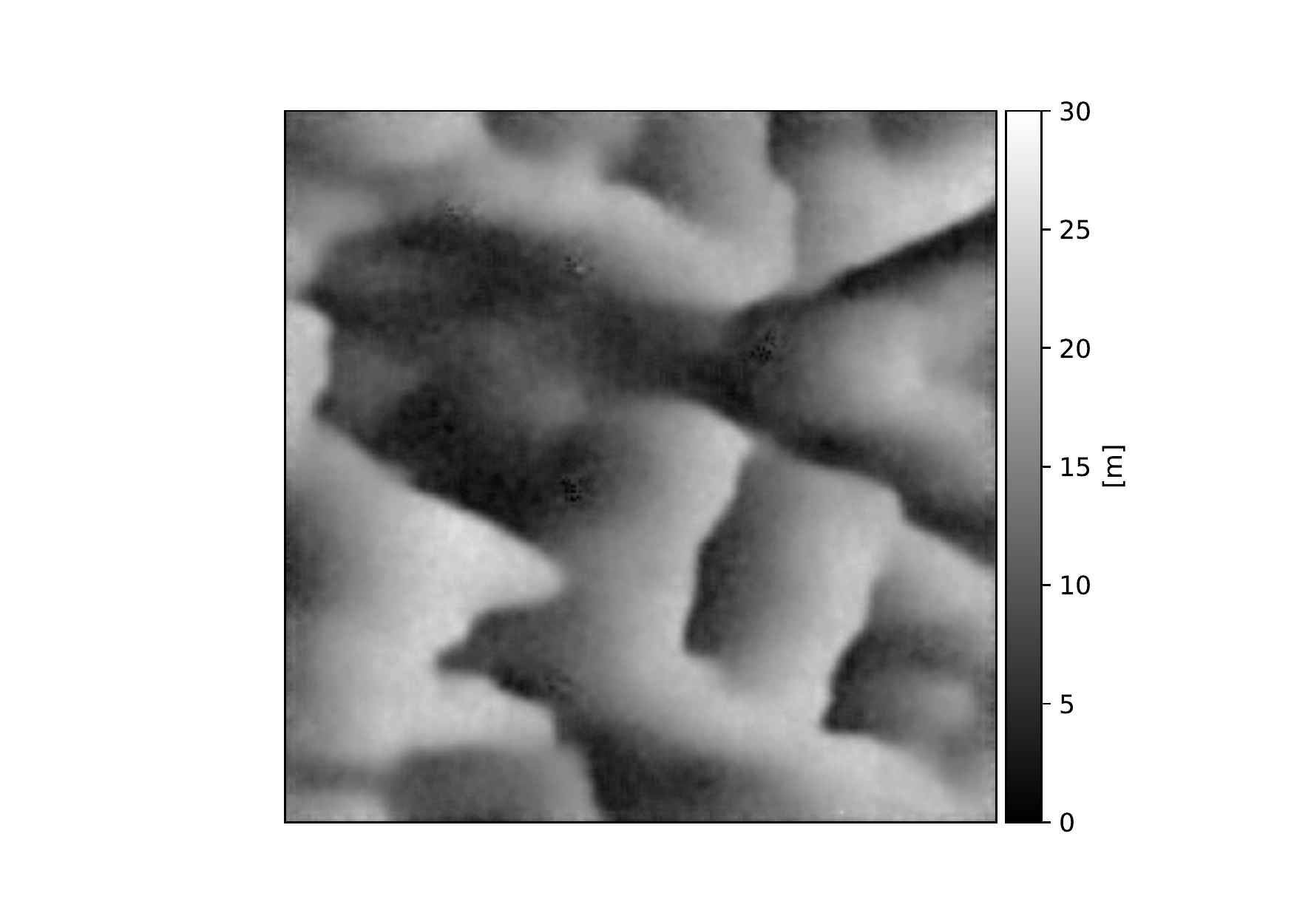}} \\
		\endgroup
		\caption{One randomly selected example output for each trained network tested on data of the same acquisition mode and scene. Each row represents another data type (see text on the left edge). The columns show the input image, the expected ground truth, and the actual output of the models, respectively. In all cases, the SAR input images were resampled to a square ground sampling distance of 1 m, regardless of the original resolutions.}
		\label{fig:internal_results}
	\end{center}
\end{figure}

Example visual results for the ``intra-scene'' experiments (i.e. training and test data come from the same scene and acquisition mode) are depicted in Figure~\ref{fig:internal_results}. As can be seen, especially the Berlin ST and Munich HS300 examples, sharp edges and fine structures are quite well reconstructed. In contrast, the SM examples are less crisp. Comparing the VHR input images with the oversampled SM images, this is not surprising, however. Still, even though the image resolution transfers to the estimated heights, also in the less highly resolved SM data the underlying structures can be discerned.

To extend those visual results, a matrix containing all numerical results for the full transferability experiment is shown in Table~\ref{tab:numericalResults}. It is important to note that for this set of experiments, all images were resampled to a square pixel spacing of 1~m, regardless of the original resolutions.

\begin{figure}[h!]
	\begin{center}
		\captionsetup[subfigure]{labelformat=empty}
		\begingroup
		\subfloat[SAR image]{\includegraphics[width=.3\textwidth, clip, trim=3cm 1cm 2cm 1cm]{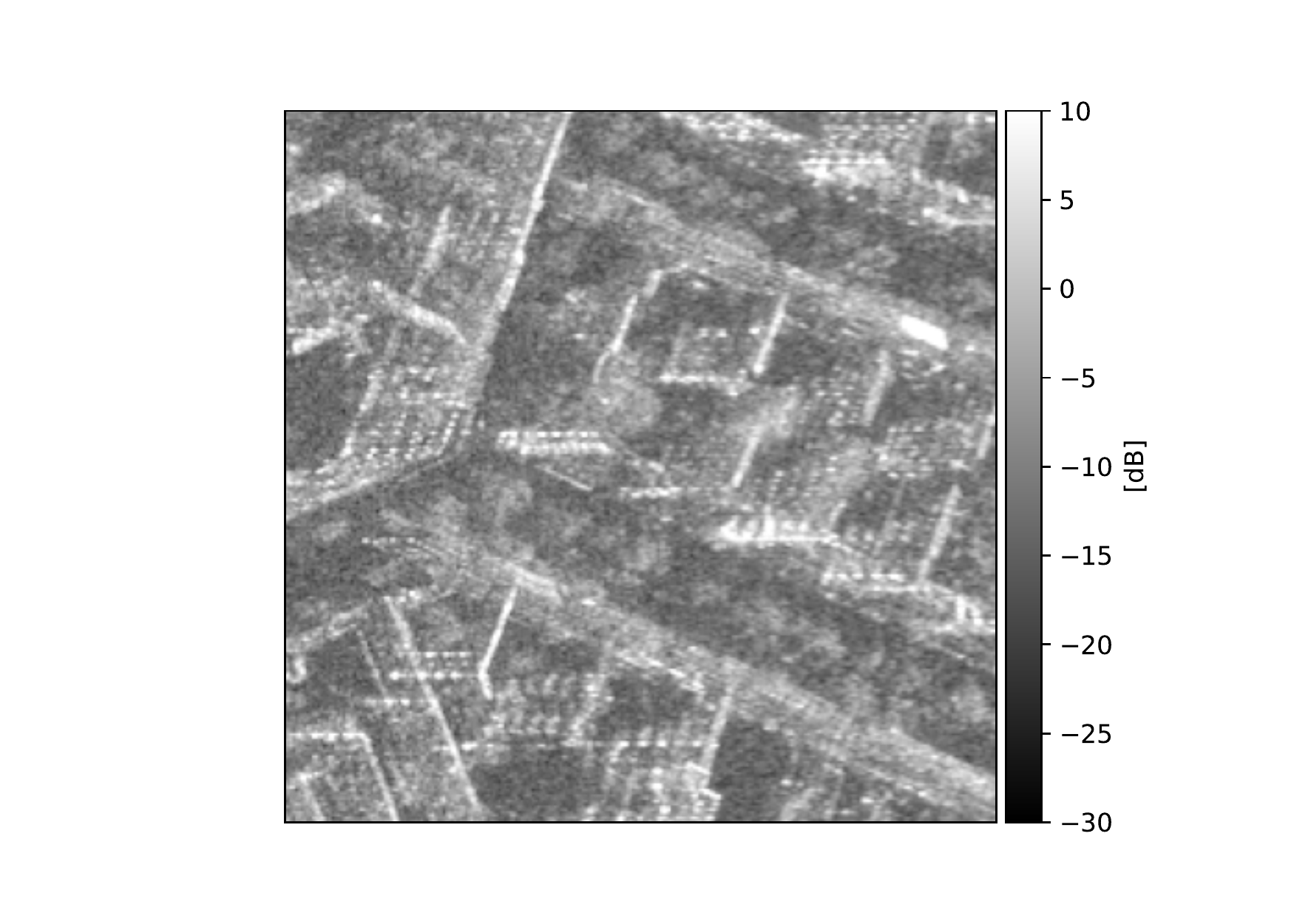}} \hspace{1mm}
		\subfloat[Ground truth]{\includegraphics[width=.3\textwidth, clip, trim=3cm 1cm 2cm 1cm]{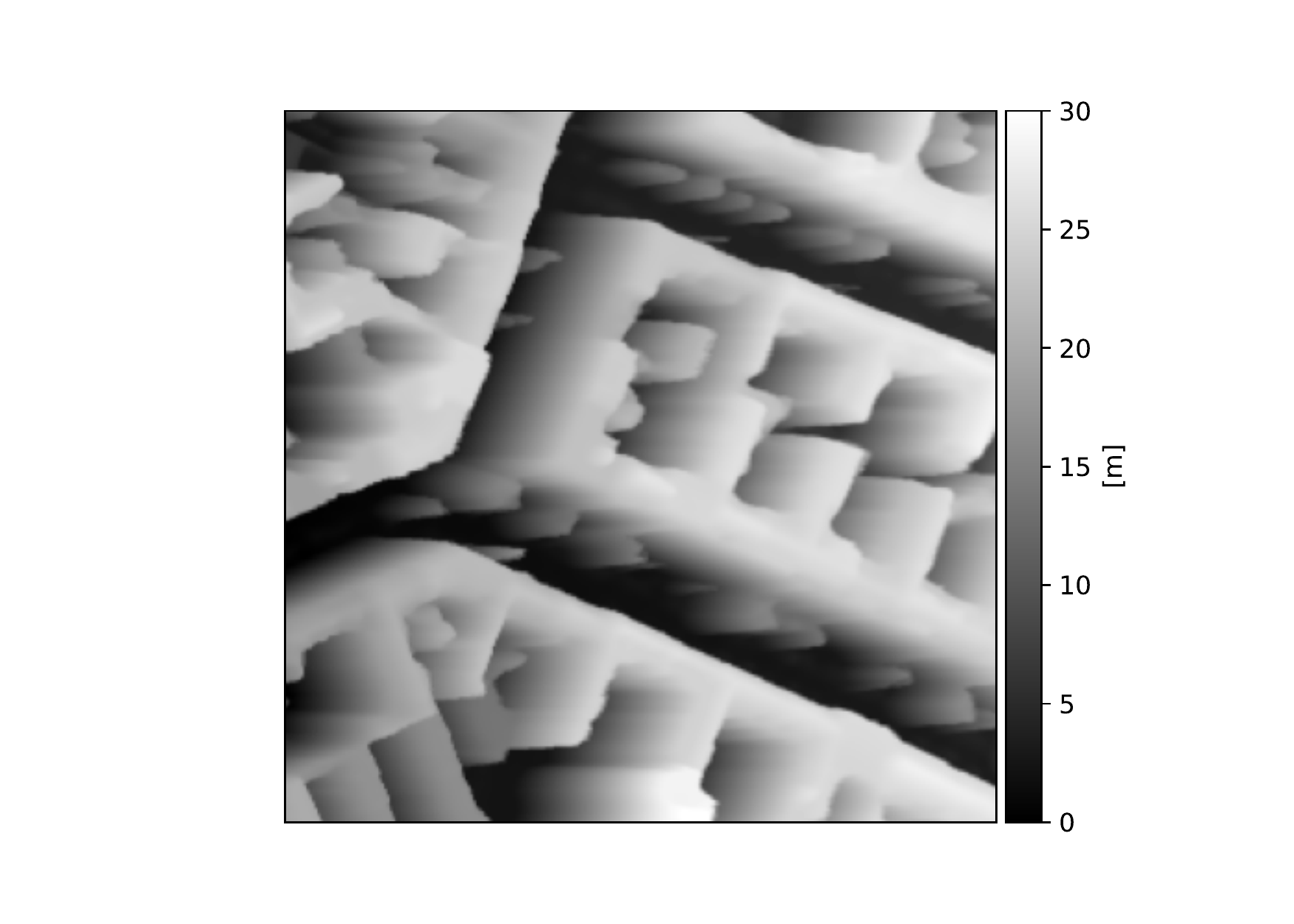}} \hspace{1mm}
		\subfloat[Estimated heights]{\includegraphics[width=.3\textwidth, clip, trim=3cm 1cm 2cm 1cm]{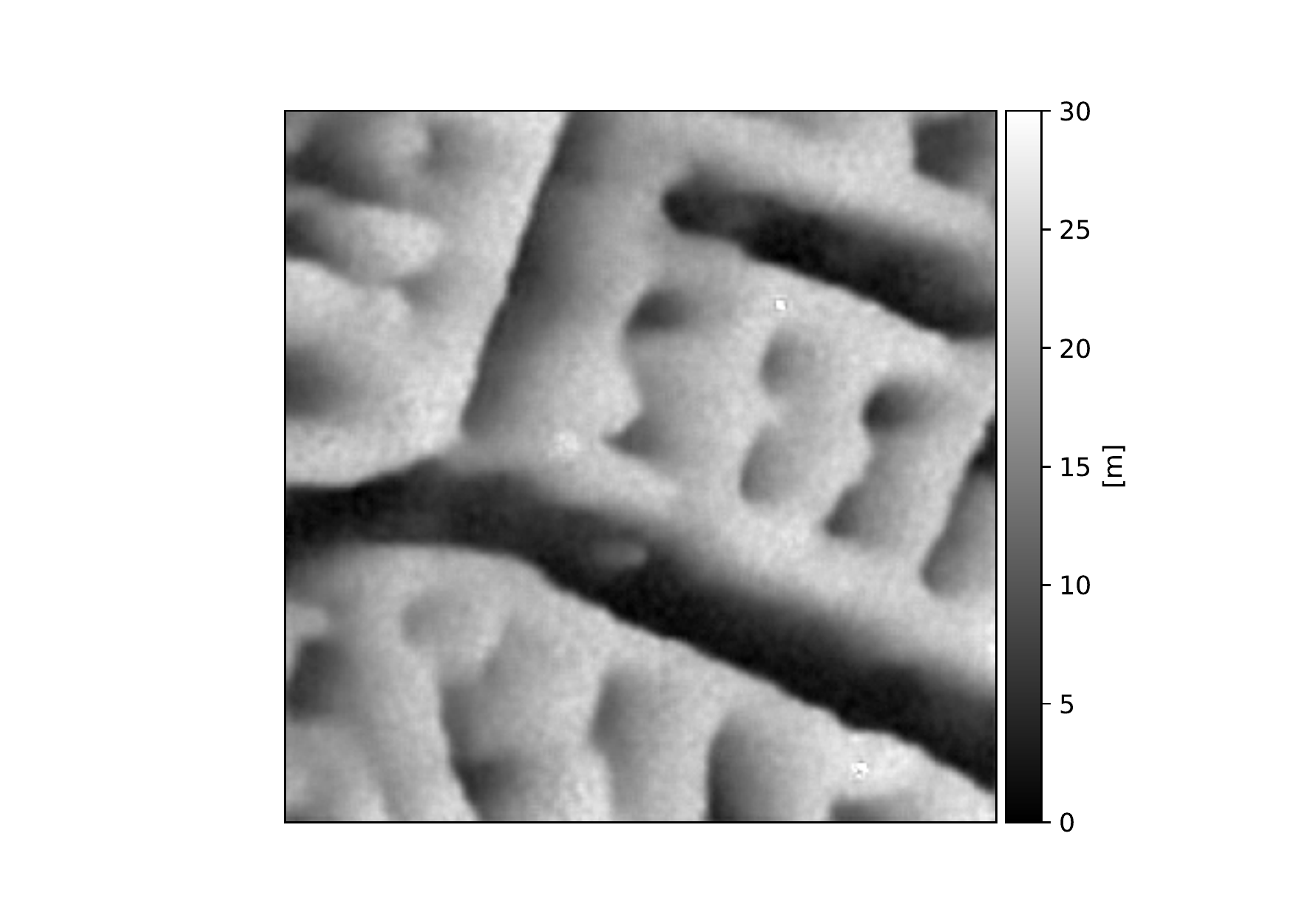}}
		\endgroup
		\caption{Example output of a model trained on Berlin SM data and tested on Berlin ST data. Both datasets were resampled to a square pixel size of 1 m. The result demonstrates the ability of the model to respond to other data types with reasonable estimates, even if the result is blurrier and not as accurate as the one trained with exactly the same data configuration.}
		\label{fig:results_transfer}
	\end{center}
\end{figure}

From Table~\ref{tab:numericalResults}, it can be seen that the VHR models trained on either the ST data of Berlin or the HS300 data of Munich perform best on the test data of the same configuration with acceptable transferability to the VHR imagery of the other scene. Both models trained on the SM data of Berlin and Munich, respectively, interestingly perform very well on the VHR ST data of the Berlin scene. Refer to Figure~\ref{fig:results_transfer} for an example transfer output from SM to VHR ST. Besides, it can be seen that as probably expected the VHR models on average perform better than the SM models.

\begin{figure}
	\begin{center}
		\captionsetup[subfigure]{labelformat=empty}
		\begingroup
		\subfloat{\includegraphics[width=.3\textwidth, clip, trim=3cm 1cm 2cm 1cm]{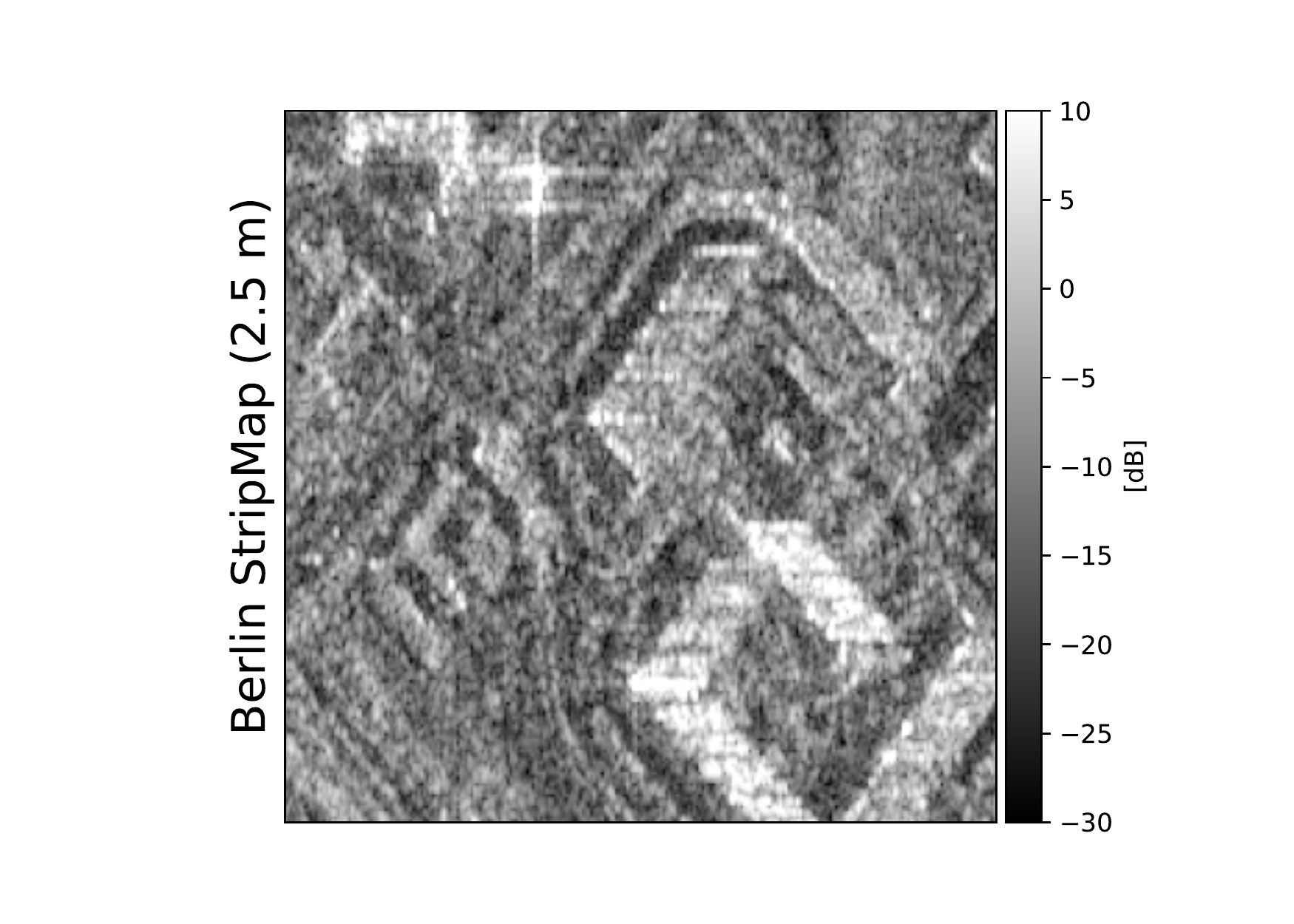}} \hspace{1mm}
		\subfloat{\includegraphics[width=.3\textwidth, clip, trim=3cm 1cm 2cm 1cm]{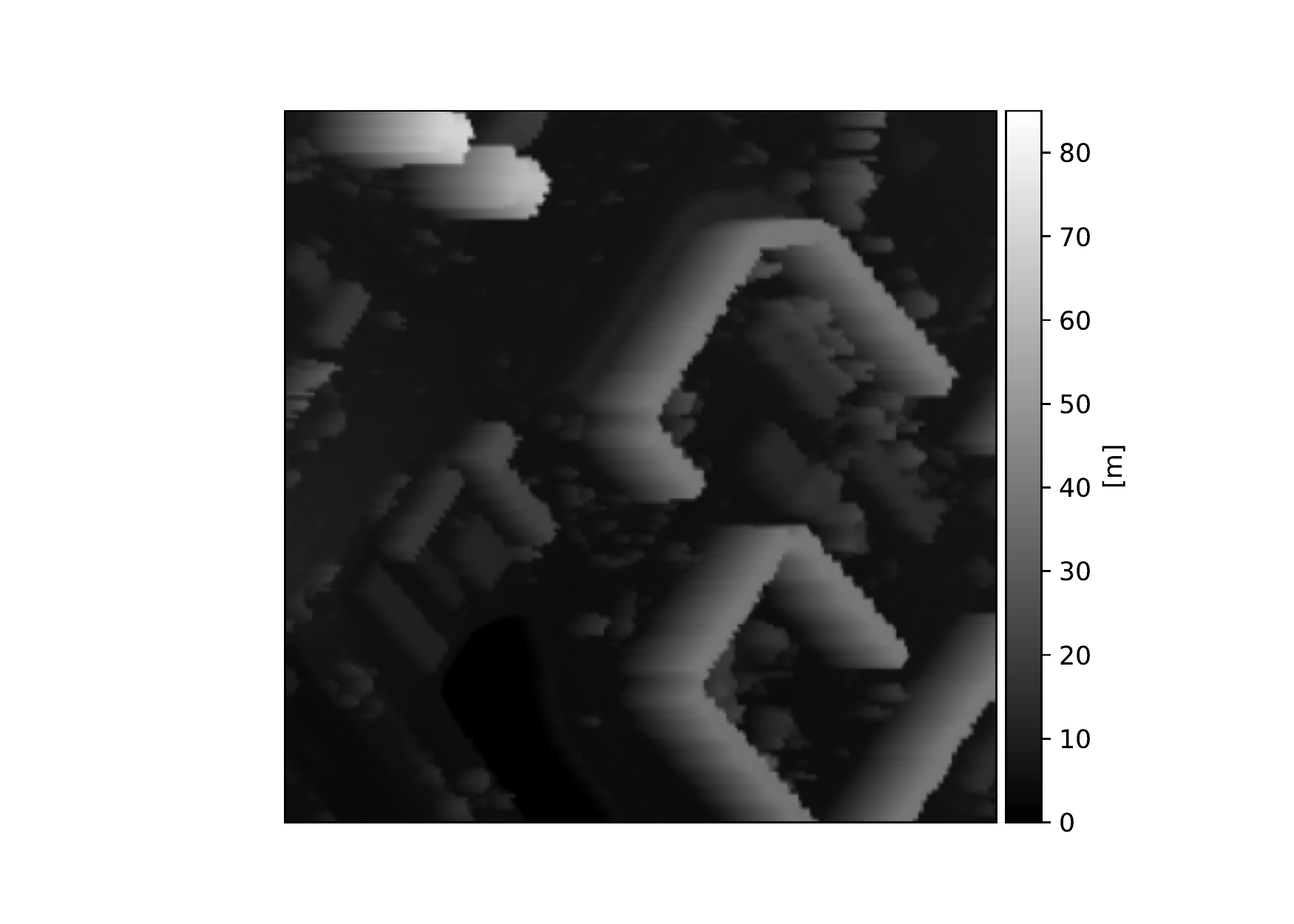}} \hspace{1mm}
		\subfloat{\includegraphics[width=.3\textwidth, clip, trim=3cm 1cm 2cm 1cm]{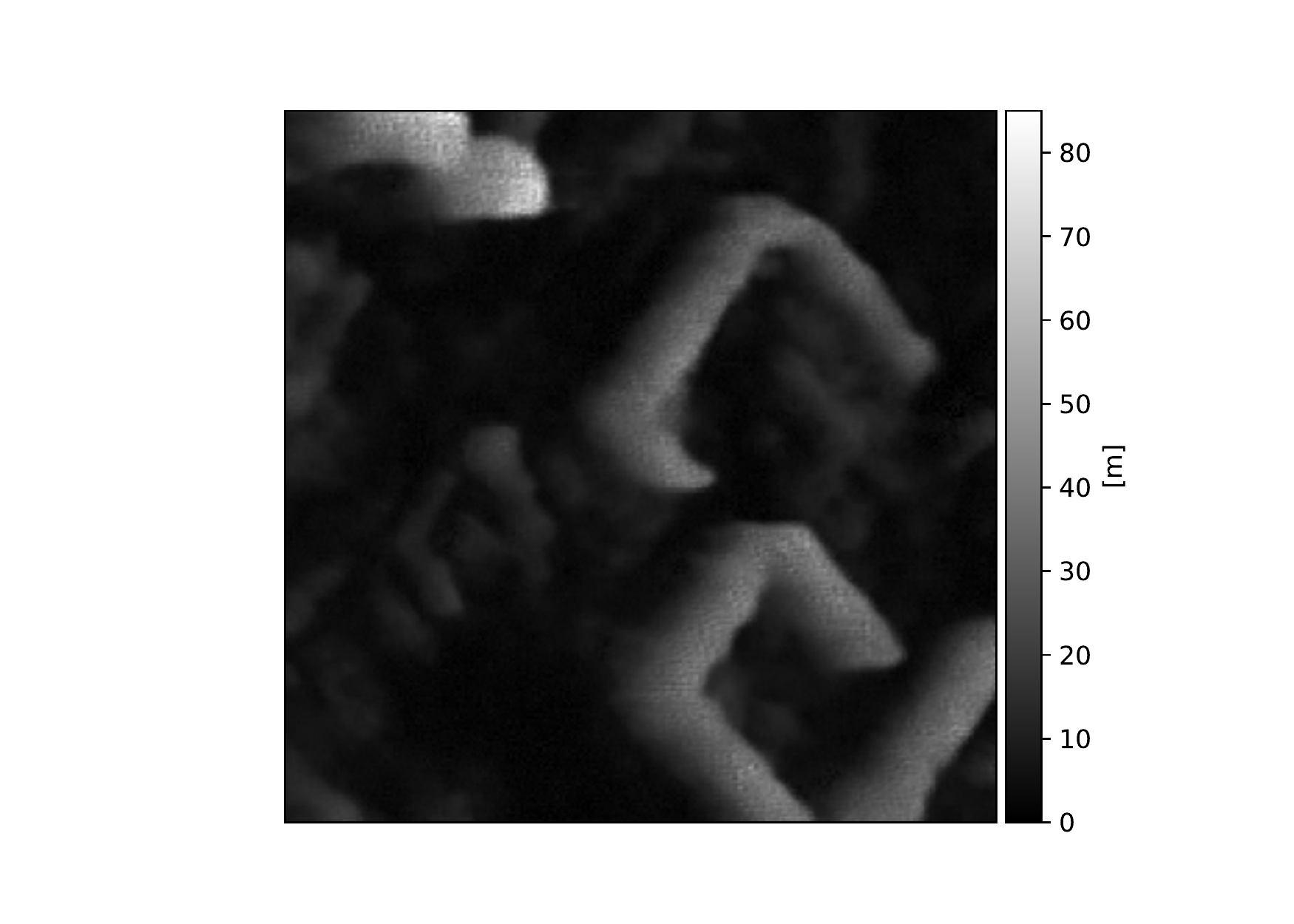}} \\
		\subfloat[SAR image]{\includegraphics[width=.3\textwidth, clip, trim=3cm 1cm 2cm 1cm]{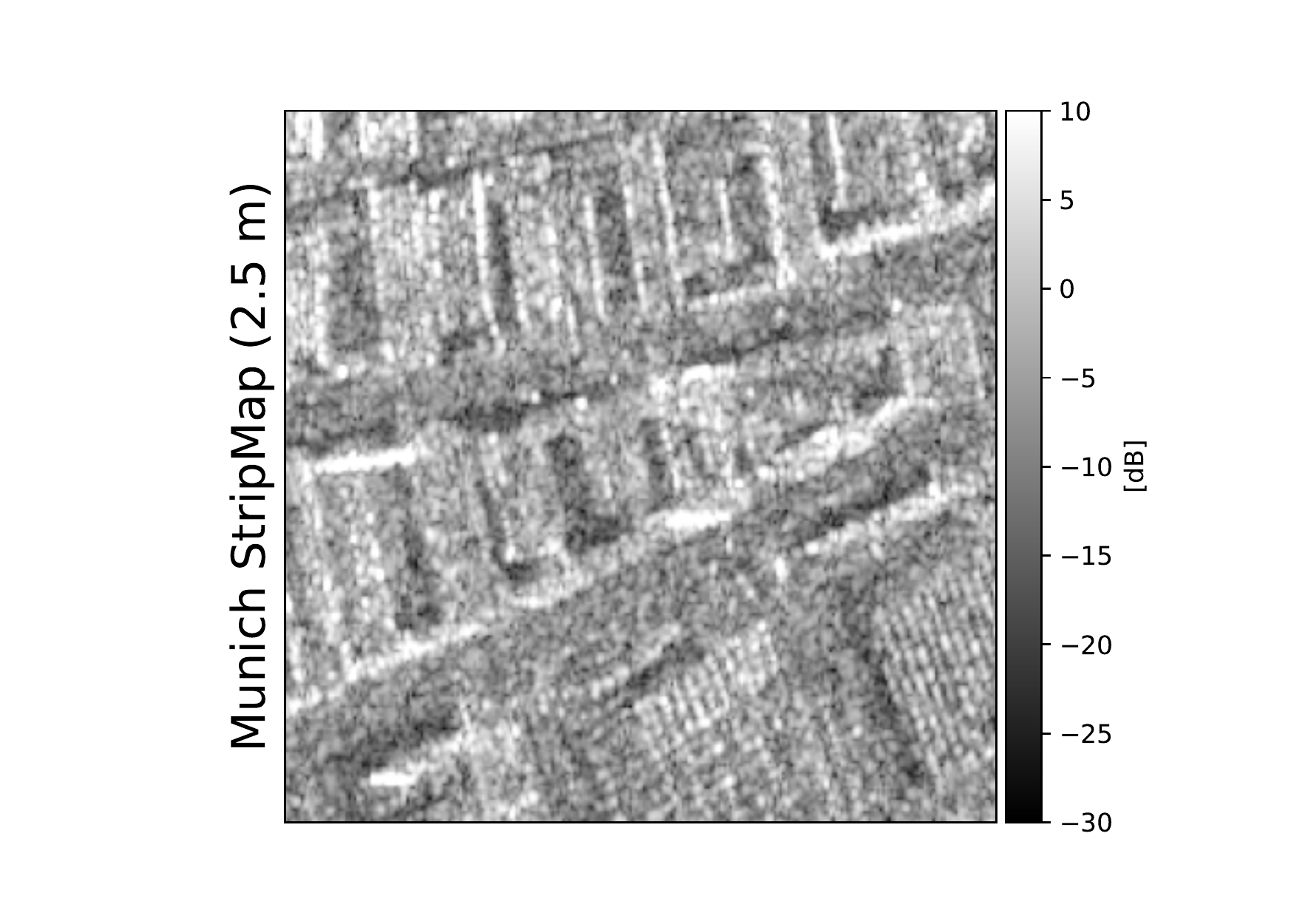}} \hspace{1mm}
		\subfloat[Ground truth]{\includegraphics[width=.3\textwidth, clip, trim=3cm 1cm 2cm 1cm]{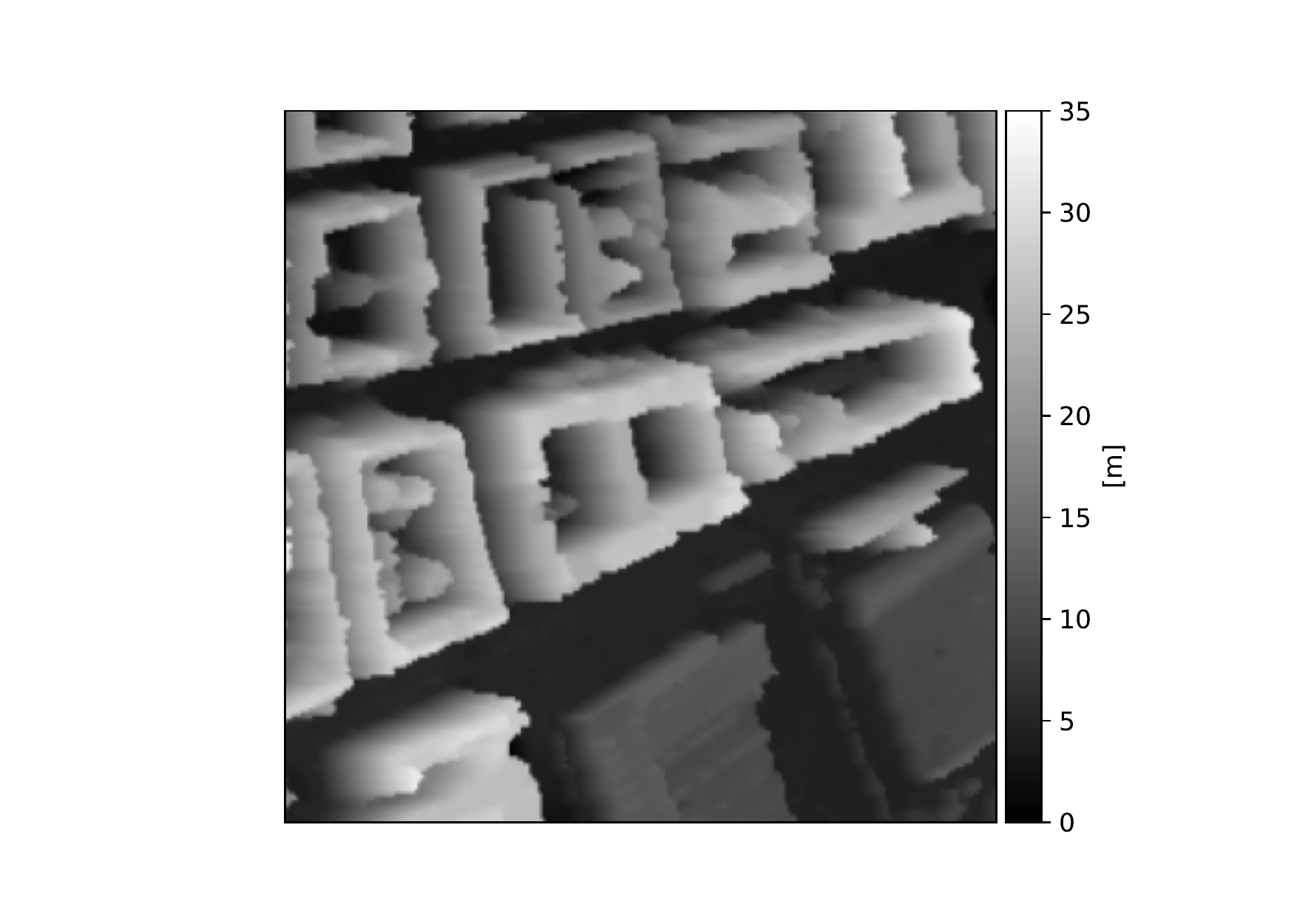}} \hspace{1mm}
		\subfloat[Estimated heights]{\includegraphics[width=.3\textwidth, clip, trim=3cm 1cm 2cm 1cm]{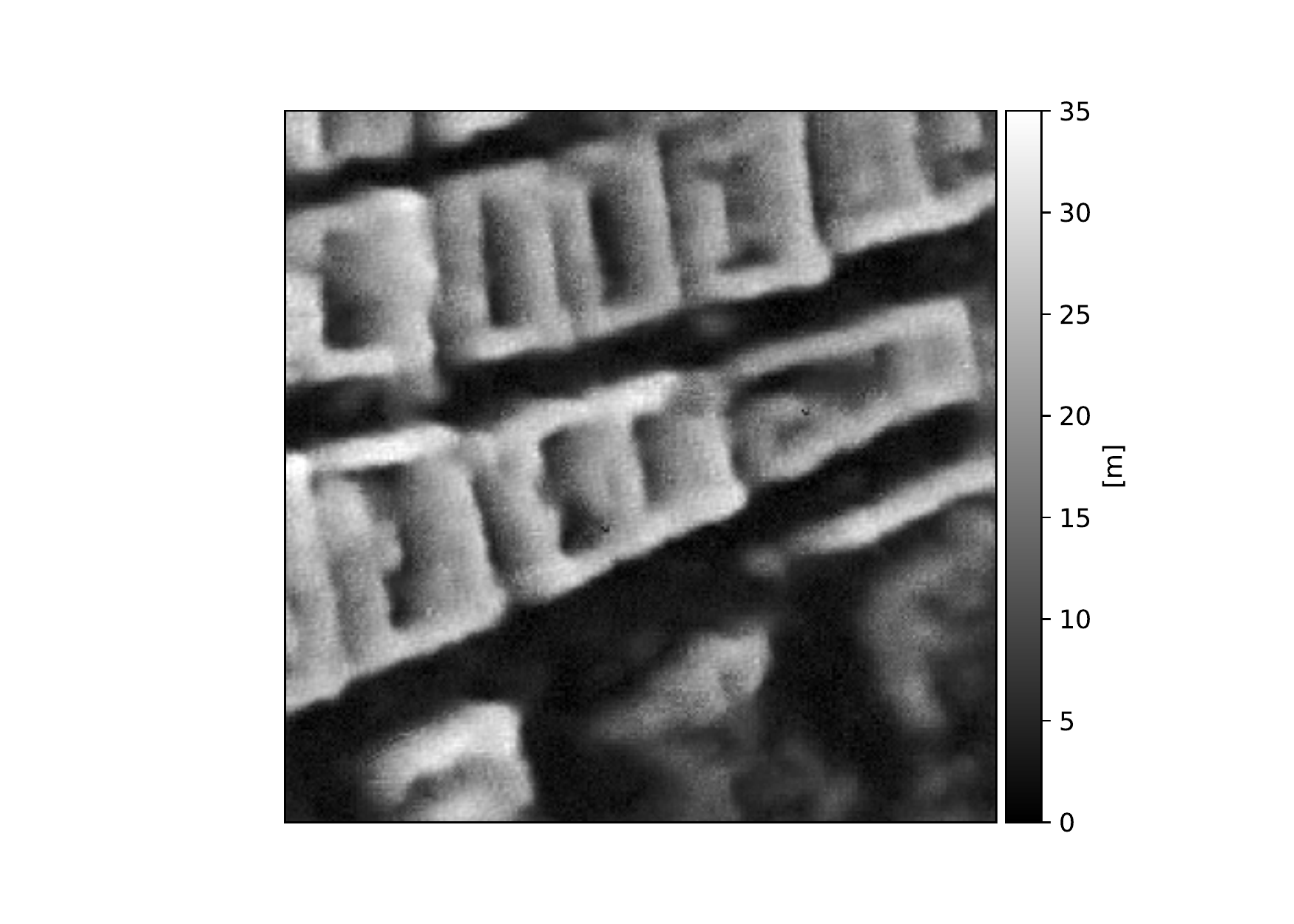}} \\
		\endgroup
		\caption{Examples of outputs from a model trained on the SM data of Berlin with a ground sampling distance of 2.5 m. The top row shows a result on the Berlin dataset (intra-scene), the bottom row shows an example of how the network reacts to SM data from Munich (cross-scene).}
		\label{fig:results_sm_2_5_m}
	\end{center}
\end{figure}

In addition to the first set of results, visual results for the SM data resampled to a square pixel spacing of 2.5~m are shown in Figure~\ref{fig:results_sm_2_5_m} with numerical results summarized in Table~\ref{tab:numericalResults_gsd_2_5_m}. From those results, it can be seen that apparently there is no benefit to be gained from oversampling the image resolution. Rather the opposite is the case -- especially the relative errors are significantly smaller compared to the previous experiments with a strongly oversampled resolution of 1~m. 

\clearpage

\clearpage

\begin{table}
    \caption{Quantitative results of four different models tested on four different test set configurations. The rows shaded in gray color indicate tests carried out on a homogeneous hold-out test set, while the remaining rows contain transferability results. Bold numbers indicate the best performance of the respective model. Note that all images have been resampled to a pixel spacing of 1~m in this experiment. The small numbers indicate the standard deviations within the test sets.}
    \vspace*{1mm}
    \centering
    \resizebox{\textwidth}{!}{%
    \begin{tabular}{llllllllll}
    \toprule
    \textbf{Train set} & \textbf{Test set} & \textbf{RMSE [m]} & $\textbf{RMSE}_\textbf{log}$ & \textbf{Rel} & $\textbf{Rel}_\textbf{log}$ & $\mathbf{\delta_1}~[\%]$ & $\mathbf{\delta_2}~[\%]$ & $\mathbf{\delta_3}~[\%]$ & \textbf{SSIM}\\
    & & \multicolumn{2}{c}{\scriptsize{lower is better}} & \multicolumn{2}{c}{\scriptsize{lower is better}} & \multicolumn{3}{c}{\scriptsize{higher is better}} &\scriptsize{high. bett.}\\
    \cmidrule(r){1-1}\cmidrule(lr){2-2}\cmidrule(lr){3-3}\cmidrule(lr){4-4}\cmidrule(lr){5-5}\cmidrule(lr){6-6}\cmidrule(lr){7-7}\cmidrule(lr){8-8}\cmidrule(lr){9-9}\cmidrule(l){10-10}
    \rowcolor{lightgray}
    \textbf{Berlin ST} & \textbf{Berlin ST} & $\mathbf{5.30 \:\scriptstyle{\pm\: 1.23}}$ & $\mathbf{0.23 \:\scriptstyle{\pm\: 0.04}}$ & $\mathbf{0.42 \:\scriptstyle{\pm\: 0.13}}$ & $\mathbf{0.16 \:\scriptstyle{\pm\: 0.03}}$ & $\mathbf{44.97 \:\scriptstyle{\pm\: 10.21}}$ & $\mathbf{66.38 \:\scriptstyle{\pm\: 8.09}}$ & $\mathbf{77.81 \:\scriptstyle{\pm\: 6.41}}$ & $\mathbf{0.78 \:\scriptstyle{\pm\: 0.07}}$ \\
    & Munich HS & $6.67 \:\scriptstyle{\pm\: 1.02}$ & $0.27 \:\scriptstyle{\pm\: 0.04}$ & $0.54 \:\scriptstyle{\pm\: 0.23}$ & $0.20 \:\scriptstyle{\pm\: 0.04}$ & $33.19 \:\scriptstyle{\pm\: 6.23}$ & $56.80 \:\scriptstyle{\pm\: 7.81}$ & $71.93 \:\scriptstyle{\pm\: 7.32}$ & $0.43 \:\scriptstyle{\pm\: 0.12}$ \\
    & Munich SM & $7.34 \:\scriptstyle{\pm\: 2.13}$ & $0.37 \:\scriptstyle{\pm\: 0.09}$ & $1.16 \:\scriptstyle{\pm\: 0.72}$ & $0.30 \:\scriptstyle{\pm\: 0.09}$ & $19.67 \:\scriptstyle{\pm\: 7.44}$ & $36.95 \:\scriptstyle{\pm\: 12.18}$ & $51.07 \:\scriptstyle{\pm\: 14.64}$ & $0.28 \:\scriptstyle{\pm\: 0.18}$ \\
    & Berlin SM & $8.28 \:\scriptstyle{\pm\: 3.19}$ & $0.37 \:\scriptstyle{\pm\: 0.10}$ & $1.10 \:\scriptstyle{\pm\: 0.79}$ & $0.29 \:\scriptstyle{\pm\: 0.09}$ & $21.42 \:\scriptstyle{\pm\: 8.70}$ & $39.43 \:\scriptstyle{\pm\: 12.82}$ & $53.37 \:\scriptstyle{\pm\: 15.21}$ & $0.52 \:\scriptstyle{\pm\: 0.18}$ \\
    \cmidrule(l{.5cm}r{.5cm}){1-10}
    & Berlin ST & $7.23 \:\scriptstyle{\pm\: 1.36}$ & $0.29 \:\scriptstyle{\pm\: 0.04}$ & $0.70 \:\scriptstyle{\pm\: 0.24}$ & $0.22 \:\scriptstyle{\pm\: 0.03}$ & $24.25 \:\scriptstyle{\pm\: 7.88}$ & $49.33 \:\scriptstyle{\pm\: 10.60}$ & $68.56 \:\scriptstyle{\pm\: 8.02}$ & $0.48 \:\scriptstyle{\pm\: 0.09}$\\
    \rowcolor{lightgray}
    \textbf{Munich HS} & \textbf{Munich HS} & $\mathbf{5.25 \:\scriptstyle{\pm\: 0.94}}$ & $\mathbf{0.22 \:\scriptstyle{\pm\: 0.05}}$ & $\mathbf{0.50 \:\scriptstyle{\pm\: 0.25}}$ & $\mathbf{0.15 \:\scriptstyle{\pm\: 0.04}}$ & $\mathbf{46.79 \:\scriptstyle{\pm\: 10.29}}$ & $\mathbf{71.50 \:\scriptstyle{\pm\: 9.45}}$ & $\mathbf{83.26 \:\scriptstyle{\pm\: 7.46}}$ & $\mathbf{0.57 \:\scriptstyle{\pm\: 0.11}}$\\
    & Munich SM & $6.67 \:\scriptstyle{\pm\: 2.36}$ & $0.35 \:\scriptstyle{\pm\: 0.08}$ & $1.03 \:\scriptstyle{\pm\: 0.56}$ & $0.28 \:\scriptstyle{\pm\: 0.08}$ & $20.24 \:\scriptstyle{\pm\: 7.40}$ & $38.44 \:\scriptstyle{\pm\: 12.28}$ & $53.33 \:\scriptstyle{\pm\: 14.73}$ & $0.21 \:\scriptstyle{\pm\: 0.14}$\\
    & Berlin SM & $9.18 \:\scriptstyle{\pm\: 3.71}$ & $0.40 \:\scriptstyle{\pm\: 0.10}$ & $1.30 \:\scriptstyle{\pm\: 0.79}$ & $0.33 \:\scriptstyle{\pm\: 0.09}$ & $14.33 \:\scriptstyle{\pm\: 6.21}$ & $31.01 \:\scriptstyle{\pm\: 11.93}$ & $47.62 \:\scriptstyle{\pm\: 15.33}$ & $0.34 \:\scriptstyle{\pm\: 0.15}$\\
   \cmidrule(l{.5cm}r{.5cm}){1-10}
    & Berlin ST & $7.74 \:\scriptstyle{\pm\: 1.51}$ & $\mathbf{0.29 \:\scriptstyle{\pm\: 0.03}}$ & $\mathbf{0.64 \:\scriptstyle{\pm\: 0.17}}$ & $\mathbf{0.24 \:\scriptstyle{\pm\: 0.03}}$ & $20.69 \:\scriptstyle{\pm\: 6.92}$ & $43.76 \:\scriptstyle{\pm\: 10.14}$ & $\mathbf{63.97 \:\scriptstyle{\pm\: 8.26}}$ & $\mathbf{0.45 \:\scriptstyle{\pm\: 0.11}}$\\
    & Munich HS & $6.97 \:\scriptstyle{\pm\: 2.24}$ & $0.35 \:\scriptstyle{\pm\: 0.09}$ & $1.10 \:\scriptstyle{\pm\: 0.62}$ & $0.27 \:\scriptstyle{\pm\: 0.08}$ & $24.49 \:\scriptstyle{\pm\: 8.10}$ & $43.90 \:\scriptstyle{\pm\: 11.76}$ & $58.07 \:\scriptstyle{\pm\: 12.87}$ & $0.27 \:\scriptstyle{\pm\: 0.18}$\\
    \rowcolor{lightgray}
    \textbf{Munich SM} & \textbf{Munich SM} & $\mathbf{5.75 \:\scriptstyle{\pm\: 1.92}}$ & $0.33 \:\scriptstyle{\pm\: 0.09}$ & $1.00 \:\scriptstyle{\pm\: 0.64}$ & $0.26 \:\scriptstyle{\pm\: 0.09}$ & $\mathbf{25.62 \:\scriptstyle{\pm\: 10.95}}$ & $\mathbf{44.23 \:\scriptstyle{\pm\: 15.15}}$ & $57.47 \:\scriptstyle{\pm\: 16.50}$ & $0.36 \:\scriptstyle{\pm\: 0.20}$\\
    & Berlin SM & $7.13 \:\scriptstyle{\pm\: 2.54}$ & $0.35 \:\scriptstyle{\pm\: 0.09}$ & $1.11 \:\scriptstyle{\pm\: 0.62}$ & $0.28 \:\scriptstyle{\pm\: 0.08}$ & $22.22 \:\scriptstyle{\pm\: 7.48}$ & $43.21 \:\scriptstyle{\pm\: 11.98}$ & $58.57 \:\scriptstyle{\pm\: 13.83}$ & $0.42 \:\scriptstyle{\pm\: 0.18}$\\
    \cmidrule(l{.5cm}r{.5cm}){1-10}
     & Berlin ST & $\mathbf{6.68 \:\scriptstyle{\pm\: 1.34}}$ & $\mathbf{0.28 \:\scriptstyle{\pm\: 0.05}}$ & $\mathbf{0.46 \:\scriptstyle{\pm\: 0.09}}$ & $\mathbf{0.21 \:\scriptstyle{\pm\: 0.04}}$ & $\mathbf{32.75 \:\scriptstyle{\pm\: 10.39}}$ & $\mathbf{54.62 \:\scriptstyle{\pm\: 10.24}}$ & $68.46 \:\scriptstyle{\pm\: 8.72}$ & $\mathbf{0.66 \:\scriptstyle{\pm\: 0.09}}$\\
    & Munich HS & $7.11 \:\scriptstyle{\pm\: 1.18}$ & $0.30 \:\scriptstyle{\pm\: 0.05}$ & $0.61 \:\scriptstyle{\pm\: 0.21}$ & $0.22 \:\scriptstyle{\pm\: 0.04}$ & $31.38 \:\scriptstyle{\pm\: 7.44}$ & $54.11 \:\scriptstyle{\pm\: 9.73}$ & $\mathbf{69.07 \:\scriptstyle{\pm\: 9.36}}$ & $0.33 \:\scriptstyle{\pm\: 0.14}$\\
    & Munich SM & $6.75 \:\scriptstyle{\pm\: 2.60}$ & $0.35 \:\scriptstyle{\pm\: 0.08}$ & $0.84 \:\scriptstyle{\pm\: 0.45}$ & $0.28 \:\scriptstyle{\pm\: 0.08}$ & $21.29 \:\scriptstyle{\pm\: 9.02}$ & $39.11 \:\scriptstyle{\pm\: 13.47}$ & $52.74 \:\scriptstyle{\pm\: 14.98}$ & $0.32 \:\scriptstyle{\pm\: 0.23}$\\
    \rowcolor{lightgray}
    \textbf{Berlin SM} & \textbf{Berlin SM} & $7.29 \:\scriptstyle{\pm\: 4.19}$ & $0.31 \:\scriptstyle{\pm\: 0.06}$ & $0.70 \:\scriptstyle{\pm\: 0.30}$ & $0.24 \:\scriptstyle{\pm\: 0.06}$ & $28.74 \:\scriptstyle{\pm\: 11.88}$ & $48.20 \:\scriptstyle{\pm\: 13.54}$ & $61.56 \:\scriptstyle{\pm\: 12.70}$ & $0.60 \:\scriptstyle{\pm\: 0.18}$\\

    \bottomrule
    \end{tabular}
    }
    \label{tab:numericalResults}
\end{table}

\begin{table}
    \caption{Numerical results for an additional test of single image height estimation from Stripmap data. The images have been resampled to a square pixel spacing of 2.5 m (in contrast to 1 m for the results in Tab.~\ref{tab:numericalResults}). The testing was performed ``intra-scene'' (rows highlighted in gray) and ``cross-scene''. Numbers in bold indicate metrics that are better to those achieved with the oversampled imagery (cf. Table~\ref{tab:numericalResults}). It can be seen that almost every metric improved.}
    \vspace*{1mm}
    \centering
    \resizebox{\textwidth}{!}{%
    \begin{tabular}{llllllllll}
    \toprule
    \textbf{Train set} & \textbf{Test set} & \textbf{RMSE [m]} & $\textbf{RMSE}_\textbf{log}$ & \textbf{Rel [m]} & $\textbf{Rel}_\textbf{log}$ & $\mathbf{\delta_1}~[\%]$ & $\mathbf{\delta_2}~[\%]$ & $\mathbf{\delta_3}~[\%]$ & \textbf{SSIM}\\
    & & \multicolumn{2}{c}{\scriptsize{lower is better}} & \multicolumn{2}{c}{\scriptsize{lower is better}} & \multicolumn{3}{c}{\scriptsize{higher is better}} &\scriptsize{high. bett.}\\
    \cmidrule(r){1-1}\cmidrule(lr){2-2}\cmidrule(lr){3-3}\cmidrule(lr){4-4}\cmidrule(lr){5-5}\cmidrule(lr){6-6}\cmidrule(lr){7-7}\cmidrule(lr){8-8}\cmidrule(lr){9-9}\cmidrule(l){10-10}
    \rowcolor{lightgray}
    \multirow{2}{18mm}{\textbf{Berlin SM \footnotesize{(GSD 2.5 m)}}} & \textbf{Berlin SM} & $\mathbf{7.01 \:\scriptstyle{\pm\: 2.31}}$ & $\mathbf{0.25 \:\scriptstyle{\pm\: 0.05}}$ & $\mathbf{0.56 \:\scriptstyle{\pm\: 0.23}}$ & $\mathbf{0.18 \:\scriptstyle{\pm\: 0.04}}$ & $\mathbf{37.11 \:\scriptstyle{\pm\: 8.49}}$ & $\mathbf{60.23 \:\scriptstyle{\pm\: 8.94}}$ & $\mathbf{74.09 \:\scriptstyle{\pm\: 8.67}}$ & $\mathbf{0.70 \:\scriptstyle{\pm\: 0.10}}$\\
    & Munich SM & $7.13 \:\scriptstyle{\pm\: 2.89}$ & $\mathbf{0.29 \:\scriptstyle{\pm\: 0.07}}$ & $\mathbf{0.63 \:\scriptstyle{\pm\: 0.30}}$ & $\mathbf{0.23 \:\scriptstyle{\pm\: 0.06}}$ & $\mathbf{27.51 \:\scriptstyle{\pm\: 9.77}}$ & $\mathbf{49.09 \:\scriptstyle{\pm\: 13.18}}$ & $\mathbf{64.67 \:\scriptstyle{\pm\: 13.31}}$ & $\mathbf{0.40 \:\scriptstyle{\pm\: 0.19}}$\\
    \cmidrule(l{.5cm}r{.5cm}){1-10}
    \textbf{Munich SM} & Berlin SM & $8.57 \:\scriptstyle{\pm\: 2.38}$ & $\mathbf{0.31 \:\scriptstyle{\pm\: 0.07}}$ & $\mathbf{0.88 \:\scriptstyle{\pm\: 0.40}}$ & $\mathbf{0.24 \:\scriptstyle{\pm\: 0.05}}$ & $\mathbf{26.51 \:\scriptstyle{\pm\: 5.12}}$ & $\mathbf{50.37 \:\scriptstyle{\pm\: 7.74}}$ & $\mathbf{67.17 \:\scriptstyle{\pm\: 9.49}}$ & $\mathbf{0.44 \:\scriptstyle{\pm\: 0.10}}$\\
    \rowcolor{lightgray}
    \textbf{\footnotesize{(GSD 2.5 m)}}& \textbf{Munich SM} & $7.17 \:\scriptstyle{\pm\: 2.08}$ & $\mathbf{0.29 \:\scriptstyle{\pm\: 0.08}}$ & $\mathbf{0.79 \:\scriptstyle{\pm\: 0.48}}$ & $\mathbf{0.23 \:\scriptstyle{\pm\: 0.07}}$ & $\mathbf{28.68 \:\scriptstyle{\pm\: 10.05}}$ & $\mathbf{50.32 \:\scriptstyle{\pm\: 14.15}}$ & $\mathbf{65.60 \:\scriptstyle{\pm\: 14.94}}$ & $\mathbf{0.38 \:\scriptstyle{\pm\: 0.16}}$\\
    \bottomrule
    \end{tabular}
    }
    \label{tab:numericalResults_gsd_2_5_m}
\end{table}

\subsection{Computational Efficiency}
To provide a feeling for the computationel efficiency of single-image height estimation, we timed the inference of a single image patch of $256 \times 256$ pixels in size. On the Deep Learning machine available to us, with an AMD\textsuperscript{\textregistered} Ryzen\texttrademark \space Threadripper\texttrademark \space 3970X 32-core processor and a NVIDIA\textsuperscript{\textregistered} Quadro RTX\texttrademark \space 8000, the evaluation of an image patch takes 4.6~ms (using only the CPU: 133~ms). To put these numbers into perspective, the test was repeated on a mid-range notebook with an Intel\textsuperscript{\textregistered} Core\texttrademark \space i5-10210U 4-core CPU and no GPU. There, a single forward pass through the network takes 750~ms.

\section{Discussion}\label{sec:Discussion}

The experimental results summarized in Section~\ref{sec:Experiments} show that single image height estimation is generally possible for both Spotlight and Stripmap SAR imagery with a standard network architecture and the data annotation approach described in Section~\ref{sec:HeightProjection}. From a detailed look at the results, more specific insights can be drawn:
\begin{itemize}\setlength{\itemsep}{0pt}
    \item As expected, the best results -- both quantitatively and qualitatively -- are achieved when both training and testing is carried out on VHR Spotlight data of the same scene. This is obviously caused by the higher granularity and lower amount of noise in such images compared to Stripmap data. In addition, the networks certainly overfit to the underlying data distribution to some extent, even though we tested on spatially separated hold-out data.
    \item The model trained on Berlin ST data transfers quite well to Munich HS300 data, while the model trained on Munich HS300 data still acceptably transfers to the Berlin ST scene. This is particularly noteworthy, as the Munich HS300 image was acquired during an ascending orbit at an approximate incidence angle of 23$^\circ$, while the Berlin ST image was acquired during a descending orbit at an approximate incidence angle of 35$^\circ$, i.e. the imaging configurations differed significantly.
    \item In contrast, the performance degradation when testing the above-mentioned VHR Spotlight models on Stripmap data -- even of the same scene -- is much more significant. Together, this indicates that the resolution is much more important than the specific acquisition mode or the study site.
    \item The models trained on the Stripmap data (oversampled to 1 m) of both study scenes interestingly provide the best test results on the Spotlight data, not generally on their own test subset. When training and testing the models with a more native pixel spacing of 2.5 m, all test metrics improve. While the error metrics are still a bit worse than those calculated for the VHR Spotlight test sets at 1 m pixel spacing, especially the structural metrics improve significantly. This indicates that oversampling the imagery is not advisable.   
\end{itemize}
In summary, the achieved results confirm the general feasibility of single image height estimation in urban areas from SAR intensity imagery, and show great potential with respect to the transferability across acquisition modes and study sites. Although there is still room left for improvement in terms of absolute accuracy, it has to be noted that the numeric results are quite comparable to those reported by \citet{Amirkolaee2019} for VHR aerial optical imagery. Given the qualitative difference between SAR and optical data from a visual point of view, this motivates further investigation of SAR-based single image height estimation.

That being said, it has to be noted that the results presented in this paper currently are limited to urban areas, which are rich with height cues \citep{Eigen2014} exploitable by the neural network. Whether single-image height estimation from SAR images also works for non-urban terrain, e.g. for croplands or forest areas, still needs to be investigated. For forest canopies promising results have already been achieved with multi-spectral satellite imagery \citep{Lang2019}, but we hypothesize that in this case the spectral profile of the individual pixels is more relevant to the tree height estimation than the spatial structure in the image.

\section{Summary and Conclusion}\label{sec:Summary}

In this paper, we have demonstrated the feasibility of deep learning-based single image height estimation from (very) high resolution SAR intensity imagery. For that purpose, we have adapted an existing network architecture that was originally developed for single image height estimation from optical imagery. In addition, we have proposed a workflow for the generation of height annotations in slant range geometry. In extensive experiments, we have put an emphasis on transferability between different SAR acquisition modes and test sites. We were able to show that single image height prediction from SAR intensity imagery is basically possible, and that the transferability between different settings is unexpectedly good, even without a specific consideration of acquisition parameters.

\section*{Acknowledgements}
This work is supported by the German Research Foundation (DFG) as grant SCHM 3322/3-1. The SAR imagery was provided by the German Aerospace Center (DLR) in the frame of the proposal MTH3753.

\bibliographystyle{unsrtnat}  
\bibliography{references}

\end{document}